\RequirePackage{iftex}\ifPDFTeX\pdfminorversion=7\fi
\documentclass[10pt,twocolumn,letterpaper]{article}

\usepackage[pagenumbers]{cvpr}

\usepackage{iftex}
\ifXeTeX
  \usepackage{fontspec}
\fi
\usepackage{cuted}
\usepackage{currfile}
\usepackage{caption}
\usepackage{booktabs}
\usepackage{multirow}
\usepackage{makecell}
\usepackage{xspace}
\usepackage{amsmath,amssymb}
\usepackage{bbding}
\usepackage{pifont}
\usepackage{tabularx}
\usepackage{colortbl}

\newcommand{\bench}{CaseScope\xspace}
\newcommand{\cmark}{\ding{51}}
\newcommand{\xmark}{\ding{55}}
\newcommand{\pmark}{(\ding{51})}

\makeatletter
\renewcommand\paragraph{\@startsection{paragraph}{4}{\z@}{2.35ex \@plus .6ex \@minus .2ex}{-1em}{\normalfont\normalsize\bfseries}}
\makeatother

\usepackage{balance}
\newcommand{\suppbalanceon}{\balance}\newcommand{\suppbalanceoff}{\nobalance}
\usepackage{longtable}

\newcommand{\nItems}{1,686}

\newcommand{\nPrompts}{6,826}
\newcommand{\nItemsAtlas}{2,993}
\newcommand{\nItemsBoth}{4,679}
\newcommand{\nPromptsAtlas}{13,764}

\newcommand{\nFieldsAtlas}{2,664}

\newcommand{\nEntitiesAtlasOnly}{453}
\newcommand{\nEntitiesBoth}{843}
\newcommand{\nLookalikeItems}{178}

\newcommand{\nMLLMs}{twenty-eight}

\newcommand{\minMLLMAtwoCirc}{0.0}
\newcommand{\maxMLLMAtwoCirc}{17.9}

\newcommand{\minMLLMAtwoVan}{9.8}
\newcommand{\maxMLLMAtwoVan}{50.6}

\newcommand{\maxMLLMCCirc}{36.4}

\newcommand{\minMLLMCqwk}{\textminus{}0.08}
\newcommand{\maxMLLMCqwk}{0.42}
\newcommand{\nCBelowRandom}{five}
\newcommand{\nCReported}{twenty-eight}
\newcommand{\randomCCirc}{11.4}

\newcommand{\minMLLMMiss}{3.3}
\newcommand{\maxMLLMMiss}{52.4}

\newcommand{\lowMissName}{Qwen2-VL-7B}
\newcommand{\lowMiss}{3.3}
\newcommand{\lowMissOvercall}{89.9}
\newcommand{\minMLLMAoneHDS}{3.8}
\newcommand{\maxMLLMAoneHDS}{34.7}

\newcommand{\minMLLMAoneExact}{2.1}
\newcommand{\maxMLLMAoneExact}{30.0}

\newcommand{\minMLLMAoneWrongBeh}{21.6}
\newcommand{\maxMLLMAoneWrongBeh}{54.1}

\newcommand{\aoneFirstName}{Patho-R1-7B}
\newcommand{\aoneFirstHDS}{34.7}

\newcommand{\aoneSecondName}{ScaleReasoner-R1}
\newcommand{\aoneSecondHDS}{32.5}

\newcommand{\aoneThirdName}{Qwen3.5-9B}
\newcommand{\aoneThirdHDS}{29.7}

\newcommand{\maxMLLMRone}{81.5}

\newcommand{\medMLLMRone}{62.0}

\newcommand{\maxMLLMRoneItem}{54.4}

\newcommand{\nRoneGain}{twenty-eight}

\newcommand{\maxRoneGain}{+16.3}
\newcommand{\medRoneGain}{+3.4}

\newcommand{\nRoneGainSigPos}{eighteen}
\newcommand{\nRoneGainSigNeg}{one}
\newcommand{\meanRoneLow}{58.8}
\newcommand{\meanRoneHigh}{57.4}
\newcommand{\nProp}{four}

\newcommand{\minPropAtwoCirc}{31.9}
\newcommand{\maxPropAtwoCirc}{61.7}

\newcommand{\minPropAtwoVan}{57.3}
\newcommand{\maxPropAtwoVan}{78.3}

\newcommand{\minPropCCirc}{54.5}
\newcommand{\maxPropCCirc}{79.5}

\newcommand{\minPropCqwk}{0.55}
\newcommand{\maxPropCqwk}{0.81}
\newcommand{\minPropMiss}{10.3}
\newcommand{\maxPropMiss}{33.6}

\newcommand{\minPropAoneHDS}{33.5}
\newcommand{\maxPropAoneHDS}{64.2}

\newcommand{\minPropAoneExact}{30.3}
\newcommand{\maxPropAoneExact}{60.6}

\newcommand{\minPropAoneWrongBeh}{14.2}
\newcommand{\maxPropAoneWrongBeh}{26.1}

\newcommand{\minPropRone}{83.2}
\newcommand{\maxPropRone}{93.2}

\newcommand{\minPropRoneItem}{55.9}
\newcommand{\maxPropRoneItem}{82.3}

\newcommand{\minPropRoneTxt}{63.7}
\newcommand{\maxPropRoneTxt}{83.0}

\newcommand{\nPropRoneGain}{four}
\newcommand{\minPropRoneGain}{+10.2}
\newcommand{\maxPropRoneGain}{+19.5}

\newcommand{\lookErrPropMin}{33.2}
\newcommand{\lookErrPropMax}{45.1}
\newcommand{\nLookProp}{four}

\newcommand{\SuffBetterText}{Twenty-seven of twenty-eight}
\newcommand{\meanSuffY}{33.0}
\newcommand{\meanSuffP}{25.8}

\newcommand{\sixWorseText}{twenty-seven of twenty-eight}
\newcommand{\meanFieldsTwo}{32.9}
\newcommand{\meanFieldsSix}{26.0}
\newcommand{\readerNCases}{120}
\newcommand{\readerAtwoClaudeSonnet}{77.5}
\newcommand{\readerAoneClaudeSonnet}{67.5}
\newcommand{\readerAtwoClaudeHaiku}{61.7}
\newcommand{\readerAoneClaudeHaiku}{46.7}
\newcommand{\readerAtwoGptSol}{75.8}
\newcommand{\readerAoneGptSol}{42.5}
\newcommand{\readerAtwoGptLuna}{62.5}
\newcommand{\readerAoneGptLuna}{39.2}
\newcommand{\readerAtwoQwenThreeFiveNine}{53.3}
\newcommand{\readerAoneQwenThreeFiveNine}{27.5}
\newcommand{\readerAtwoPathoRone}{37.5}
\newcommand{\readerAonePathoRone}{36.7}

\newcommand{\dltMinSingle}{\textminus{}7.1}
\newcommand{\dltMaxSingle}{+2.1}

\newcommand{\dltNwSingle}{twenty-eight}

\newcommand{\dltNsigwDownSingle}{six}
\newcommand{\dltNsigwUpSingle}{one}

\newcommand{\dltNwNosite}{twenty-eight}

\newcommand{\dltNsigwNosite}{twenty-five}

\newcommand{\dltMagHiNosite}{9.4}

\newcommand{\sysHardAName}{breast}

\newcommand{\sysHardBName}{lymph node \& spleen}

\newcommand{\sysEasyBName}{endocrine}

\newcommand{\sysEasyAName}{GI tract}
\newcommand{\famDistrZero}{34}

\newcommand{\famDistrFivePlus}{25}

\newcommand{\judgeN}{297}
\newcommand{\judgeKappaJudge}{0.73}
\newcommand{\judgeKappaFinal}{0.76}
\newcommand{\judgeExactFinal}{80.5}
\newcommand{\judgeMeanFinal}{0.32}
\newcommand{\judgeMeanRater}{0.49}

\newcommand{\atlRhoAtwo}{0.83}
\newcommand{\atlRhoAtwoLo}{0.59}
\newcommand{\atlRhoAtwoHi}{0.96}
\newcommand{\atlRhoAtwoOpen}{0.74}
\newcommand{\atlNAtwo}{32}

\newcommand{\atlAtlAtwoPropLo}{24.1}
\newcommand{\atlAtlAtwoPropHi}{49.1}
\newcommand{\atlAtlAtwoOpenLo}{0.0}
\newcommand{\atlAtlAtwoOpenHi}{14.2}

\newcommand{\atlRhoAone}{0.80}

\newcommand{\atlSharedPropAtwo}{\textminus{}13.0}
\newcommand{\atlSharedOpenAtwo}{+0.9}
\newcommand{\atlSharedPropAtwoMag}{13.0}

\newcommand{\atlSharedPropAone}{\textminus{}11.5}
\newcommand{\atlSharedOpenAone}{\textminus{}1.1}

\newcommand{\atlNShared}{347}
\newcommand{\atlNOnly}{453}
\newcommand{\atlDropTwo}{\textminus{}15.4}
\newcommand{\atlDropFive}{\textminus{}4.3}
\newcommand{\atlOnlyPropAtwo}{+4.0}
\newcommand{\atlOnlyOpenAtwo}{\textminus{}1.3}
\newcommand{\atlOnlyDownAtwo}{3}
\newcommand{\atlOnlyNAtwo}{32}
\newcommand{\atlOnlyPropAone}{\textminus{}11.4}
\newcommand{\atlOnlyOpenAone}{\textminus{}6.7}
\newcommand{\atlOnlyDownAone}{28}
\newcommand{\atlOnlyNAone}{32}
\newcommand{\atlNNAtwoOpen}{+5.3}
\newcommand{\atlNNAtwoProp}{+10.2}

\newcommand{\atlKeptProp}{61}
\newcommand{\atlKeptTeachProp}{68}

\newcommand{\atlKeptOpen}{26}
\newcommand{\atlKeptTeachOpen}{24}
\newcommand{\atlTeachSingleMean}{\textminus{}1.0}
\newcommand{\atlTeachSingleDown}{6}

\newcommand{\atlTeachNositeMean}{\textminus{}4.7}
\newcommand{\atlTeachNositeDown}{25}

\newcommand{\atlSingleMean}{\textminus{}2.2}

\newcommand{\atlSingleDown}{15}
\newcommand{\atlSingleN}{28}
\newcommand{\atlNositeMean}{\textminus{}2.9}

\newcommand{\atlNositeDown}{19}

\newcommand{\atlThreePlusCases}{425}
\newcommand{\atlTeachThreePlusCases}{379}

\newcommand{\atlGainPropMean}{+5.0}

\newcommand{\atlGainPropN}{4}
\newcommand{\atlGainPropSigPos}{3}

\newcommand{\atlGainOpenMean}{+3.8}

\newcommand{\atlGainOpenN}{28}
\newcommand{\atlGainOpenSigPos}{14}
\newcommand{\atlGainOpenSigNeg}{one}

\newcommand{\atlTextOnlyHi}{74.7}
\newcommand{\atlGonePropLo}{39.9}
\newcommand{\atlGonePropHi}{61.5}

\newcommand{\atlGonePropN}{4}
\newcommand{\atlGonePropAboveFixed}{4}

\newcommand{\atlGoneOpenLo}{0.7}
\newcommand{\atlGoneOpenHi}{42.7}

\newcommand{\atlGoneOpenN}{28}
\newcommand{\atlGoneOpenAboveFixed}{6}

\newcommand{\atlGoneChance}{8.5}
\newcommand{\atlGoneChanceVan}{31.5}

\newcommand{\atlGoneFixedBest}{31.7}
\newcommand{\atlGoneFixedBestImage}{2}

\newcommand{\atlMissPropMean}{32.2}
\newcommand{\atlMissTeachPropMean}{18.6}

\newcommand{\atlMissOpenMean}{30.2}
\newcommand{\atlMissTeachOpenMean}{21.2}

\newcommand{\detailCases}{436}
\newcommand{\detailEntities}{390}
\newcommand{\detailMemberships}{413}
\newcommand{\detailRetainedFields}{2,448}
\newcommand{\detailInputFields}{2,277}
\newcommand{\detailCapCases}{64}
\newcommand{\detailSingletons}{357}
\newcommand{\detailSingletonPct}{91.5}
\newcommand{\detailMultiCaseEntities}{33}
\newcommand{\detailHeY}{194}
\newcommand{\detailHeP}{221}
\newcommand{\detailHeN}{21}

\newcommand{\detailAtlCases}{800}
\newcommand{\detailAtlFields}{2,664}
\newcommand{\detailAtlTone}{637}
\newcommand{\detailAtlTtwo}{128}
\newcommand{\detailAtlTthree}{35}
\newcommand{\detailAtlWithB}{762}
\newcommand{\detailAtlWithoutB}{38}
\newcommand{\detailAtlSeries}{413}
\newcommand{\detailAtlMultiPanel}{387}
\newcommand{\detailAtlOntology}{361}
\newcommand{\detailAtlWho}{309}
\newcommand{\detailAtlNew}{130}
\newcommand{\detailAtlShared}{347}

\newcommand{\detailAtlAdded}{439}
\newcommand{\detailAtlHeFallback}{289}

\newcommand{\forcedN}{14,623}
\newcommand{\forcedModels}{eight}
\newcommand{\forcedLeft}{one is}
\newcommand{\forcedFirstName}{Qwen3.5-9B}
\newcommand{\forcedFirst}{34.7}
\newcommand{\forcedSecondName}{Qwen3-VL-8B-Thinking}
\newcommand{\forcedSecond}{23.0}
\newcommand{\forcedRestMax}{5.7}

\newcommand{\budgetOtherHits}{36}
\newcommand{\budgetOtherN}{288,260}
\newcommand{\budgetMedgemmaPct}{17.6}
\newcommand{\budgetMedgemmaAtlasAtwoPct}{32.5}
\newcommand{\budgetLlavaAoneHits}{148}

\definecolor{cvprblue}{rgb}{0.21,0.49,0.74}
\usepackage[pagebackref,breaklinks,colorlinks,allcolors=cvprblue]{hyperref}
\newcommand{\sref}[1]{\ref{#1}}
\newcommand{\ofmain}{\xspace}

\title{CaseScope: A Multi-Field Benchmark for\\
Case-Level Pathology Diagnosis with Multimodal LLMs}

\makeatletter
\patchcmd{\@maketitle}{\vspace*{24pt}}{\vspace*{12pt}}{}{\PackageWarning{arxiv}{title gap patch failed}}
\patchcmd{\@maketitle}{\vskip .375in}{\vskip .1in}{}{\PackageWarning{arxiv}{top gap patch failed}}
\patchcmd{\abstract}{\vspace*{12pt}}{\vspace*{6pt}}{}{\PackageWarning{arxiv}{abstract gap patch failed}}
\makeatother

\newcommand{\authsep}{\hspace{0.6em}}
\author{%
Wenchuan Zhang$^{1,2,*}$\authsep
Chengyang Zhang$^{1,2,3,*}$\authsep
Fengling Li$^{1,2,*}$\authsep
Ting Yin$^{1,2}$\authsep
Chen Shu$^{3}$\authsep
Jie Chen$^{2}$\\
Shuwan Zhang$^{4}$\authsep
Xiaoyan Wu$^{5}$\authsep
Zhang Zhang$^{1,2}$\authsep
Yuhao Yi$^{1,2,3,\dagger}$\authsep
Jiancheng Lv$^{3}$\authsep
Hong Bu$^{1,2}$\\[0.2em]
\small\begin{tabular}[t]{@{}c@{}}
$^1$Department of Pathology, West China Hospital, Sichuan University\\
$^2$Institute of Clinical Pathology, West China Hospital, Sichuan University\\
$^3$College of Computer Science, Sichuan University\\
$^4$Department of Pathology, Shengjing Hospital of China Medical University\\
$^5$Department of Pathology, Renmin Hospital of Wuhan University\\[0.2em]
\texttt{zhangwenchuan@stu.scu.edu.cn}\quad \texttt{yuhaoyi@scu.edu.cn}\\
$^*$Equal contribution\quad $^\dagger$Corresponding author
\end{tabular}%
}

\begin{document}
\maketitle
\newsavebox{\caseoverviewbox}
\begin{figure*}[t]
\begin{lrbox}{\caseoverviewbox}
\begin{minipage}{\textwidth}
\centering
\includegraphics[width=\textwidth]{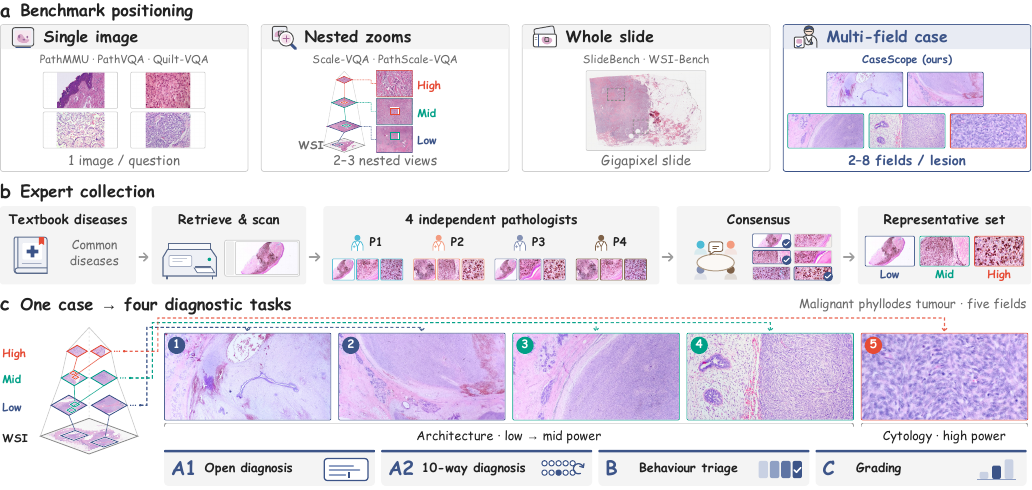}
\caption{\textbf{\bench overview.} (a) Input settings of existing benchmarks and \bench.
(b) Textbook-guided slide retrieval and scanning, independent sampling of one WSI by four pathologists, and consensus selection.
(c) Complementary fields from one lesion support four diagnostic tasks.}
\label{fig:teaser}
\end{minipage}
\end{lrbox}
\typeout{FIGURE1-HEIGHT: \the\dimexpr\ht\caseoverviewbox+\dp\caseoverviewbox\relax}
\typeout{FIGURE1-WITH-GAP: \the\dimexpr\ht\caseoverviewbox+\dp\caseoverviewbox+\dbltextfloatsep\relax}
\typeout{FIGURE1-HALF-PAGE-LIMIT: \the\dimexpr\textheight/2\relax}
\ifdim\dimexpr\ht\caseoverviewbox+\dp\caseoverviewbox+\dbltextfloatsep\relax>\dimexpr\textheight/2\relax
  \PackageError{CaseScope figure}{Figure 1 exceeds the half-page budget}{Reduce the figure height or caption.}
\fi
\centering\usebox{\caseoverviewbox}
\end{figure*}

\newcommand{\abstractcodelink}{ The evaluation code is available at \href{https://github.com/Wenchuan-Zhang/CaseScope}{https://github.com/\allowbreak\mbox{Wenchuan-Zhang/CaseScope}}.}
\providecommand{\abstractcodelink}{}
\begin{abstract}
Pathologists rarely diagnose from a single image: they scan a section at low power for architecture, zoom into selected fields for cytology, and integrate the evidence across views and scales.
Pathology benchmarks for multimodal large language models (MLLMs), however, mostly ask about one image or require whole-slide pipelines, and the few that show several fields draw nested regions from cancer slides of one public cohort.
We introduce \bench, a benchmark of diagnosis from several curated fields of one lesion: \bench-Teaching, 436 cases from a residency-training image library, reconstructed, de-identified, harmonised against a WHO-based ontology and re-examined by a pathologist wherever an automated pre-review raised a doubt, and \bench-Atlas, 800 cases from 88 published atlases (387 of them figure series of one disease), labelled from the captions by two LLMs that must agree.
They cover \nEntitiesBoth\ entities across 18 organ systems and pose \nItemsBoth\ items in six tasks, from open-ended and ten-way diagnosis with ontology-guided hard negatives under circular evaluation to the verification and location of morphological findings.
Across \nMLLMs\ open and \nProp\ proprietary MLLMs, diagnoses are unstable---the best open and proprietary models are correct under all ten option rotations on \maxMLLMAtwoCirc\% and \maxPropAtwoCirc\% of the teaching cases---and wrong answers favour the classic look-alike; open models make little use of the additional fields on teaching cases, where the one-line specimen site matters more, although on atlas cases with three or more fields the fields count about as much as the site; and most open models verify findings barely above chance (median \medMLLMRone\% vs.\ 50\%, best \maxMLLMRone\%), gaining a median \medRoneGain\ points from images; proprietary models reach \minPropRone--\maxPropRone\%.
Rankings broadly replicate between the sources.\abstractcodelink
\end{abstract}

\section{Introduction}
\label{sec:intro}

A surgical pathologist does not read a case from one picture: at low magnification they judge architecture (circumscription, infiltration, growth pattern), at high magnification cytology (nuclear size, chromatin, mitoses), and they move between the two until the fields agree.
No single field carries the answer: a circumscribed spindle-cell proliferation means little until a high-power field shows (or fails to show) atypia, and a bland gland is reassuring only if the low-power field shows no infiltration.
Residents are trained and examined on exactly this: several fields of one lesion at increasing magnification.

Multimodal large language models (MLLMs) are increasingly proposed as assistants for pathology~\cite{lu2024pathchat,seyfioglu2024quiltllava,sun2025pathgen}, yet their evaluation rarely matches this task.
Most pathology visual-question-answering benchmarks~\cite{he2020pathvqa,sun2024pathmmu,seyfioglu2024quiltllava,chen2026omnipathovqa} ask about a single image, whole-slide benchmarks~\cite{chen2024wsivqa,chen2025slidechat,liang2025wsillava,liu2026pathoargus} mostly test slide encoders or navigation agents, and pathology MLLMs are tuned on single image--text pairs~\cite{li2023llavamed,seyfioglu2024quiltllava,sun2025pathgen,zhang2025pathor1}; whether they can read several fields of one case is largely untested.
The first cross-scale benchmarks~\cite{phan2026scalevqa,phan2026pathscale} show nested regions of TCGA cancer slides with LLM-generated questions, and general multi-image benchmarks~\cite{fu2024blink,wang2024muirbench,meng2024mmiu} probe generic cross-image relations, not expert evidence integration.
Missing is the setting in which pathologists are trained: \emph{several curated fields of one lesion from which a diagnosis must be synthesised}, across diagnostic pathology rather than the cancers of one cohort, with a test of whether models integrate the fields at all.
Curated fields isolate this integration step: the regions are already chosen, so failure cannot come from navigation or slide size, and a case can be shown with fewer fields or without its site to see what an answer rests on.

Teaching collections are a natural source, since an expert chose each case's fields, but are not benchmark-ready: in our residency library the same case is re-filed under several organ systems and names, with labels from several WHO editions; we document each conversion step so that others can reuse their own material.
Published atlases, whose captioned figures were chosen to show an entity, are a second source and test whether findings carry over.

We introduce \bench, a benchmark for case-level, multi-field pathology diagnosis (\cref{fig:teaser}).
\bench-Teaching contains 436 cases---2,448 H\&E fields, median five per case, ordered from low to high magnification---covering 390 entities across 18 organ systems of the residency curriculum; \bench-Atlas adds 800 cases (\nFieldsAtlas\ fields, median three) from 88 published atlases, one per entity, \nEntitiesAtlasOnly\ of them without a teaching case.
Six tasks probe the skill at different levels: open-ended and ten-way diagnosis with ontology-guided distractors that include each entity's classic look-alikes, under circular evaluation~\cite{liu2024mmbench}; behaviour triage; ordinal grading (the four diagnostic tasks of \cref{fig:teaser}c); verification of morphological findings at low and at high power; and, on the atlas cases, locating a described finding among the fields.

Experiments with \nMLLMs\ open and \nProp\ proprietary MLLMs show three things.
First, MLLM diagnoses are fragile: on the teaching cases single-pass ten-way accuracy reaches at most \maxMLLMAtwoVan\% for open and \maxPropAtwoVan\% for proprietary models, the correct answer under all ten option rotations at most \maxMLLMAtwoCirc\% and \maxPropAtwoCirc\%, and wrong answers favour the classic look-alike.
Second, open models make little use of the additional fields on teaching cases: keeping only the highest-power field changes circular accuracy by \dltMinSingle\ to \dltMaxSingle\ points, whereas removing the one-line specimen site lowers it by up to \dltMagHiNosite; on atlas cases with three or more fields the further fields count about as much as the site.
Third, most open models barely verify the findings behind a diagnosis: they judge a median \medMLLMRone\% of the statements on architecture and cytology correctly (R1, chance 50\%; best \maxMLLMRone\%), and the images add a median \medRoneGain\ points; proprietary models judge \minPropRone--\maxPropRone\% correctly and gain \minPropRoneGain\ to \maxPropRoneGain\ points from the images.
Rankings broadly replicate between the sources ($\rho=\atlRhoAtwo$ on A2), although on the shared entities the proprietary models lose \atlSharedPropAtwoMag\ points on atlas cases.

\paragraph{Contributions.}
(1)~A curated multi-field benchmark from two sources: teaching cases with a reproducible construction pipeline (de-duplication, de-identification, ontology harmonisation, pathologist confirmation of flagged cases) and atlas cases labelled from their captions.
(2)~A six-task suite of \nItemsBoth\ items separating diagnosis, triage, grading, finding verification and finding location, with circular evaluation and case-level bootstrap intervals.
(3)~A study of \nMLLMs\ open and \nProp\ proprietary MLLMs quantifying whether and how they use multiple fields, and whether findings carry over between sources.

\section{Related Work}
\label{sec:related}

\paragraph{Pathology foundation models and MLLMs.}
Vision-language pre-training on pathology image--text pairs produced CLIP-style models such as PLIP~\cite{huang2023plip}, QuiltNet~\cite{ikezogwo2023quilt}, CONCH~\cite{lu2024conch} and MUSK~\cite{xiang2025musk}, and self-supervised encoders such as UNI~\cite{chen2024uni}.
Building on instruction tuning~\cite{liu2023llava}, pathology MLLMs adapt general models with pathology dialogue or caption data~\cite{seyfioglu2024quiltllava,lu2024pathchat,sun2025pathgen}, lately with reinforcement-learned reasoning~\cite{zhang2025pathor1,xu2025pathcopilot,liu2025teampath,jiang2026pathreasoner}, knowledge retrieval or memory~\cite{zhang2026pathoagenticrag,li2026pathmem} and evidence-grounded agents~\cite{xu2026pathpocket,cao2026sql}; medical generalists~\cite{li2023llavamed,chen2024huatuogptvision} and frontier general MLLMs~\cite{bai2025qwen25vl,zhu2025internvl3} are applied to pathology as well, also as autonomous slide-viewing agents~\cite{aubreville2026agents}; almost all are evaluated one image at a time.

\paragraph{Pathology benchmarks for MLLMs.}
\Cref{fig:teaser}a and \cref{tab:positioning} contrast \bench with existing pathology benchmarks for MLLMs.
Most ask about a single image, with questions from textbook captions~\cite{he2020pathvqa}, educational videos~\cite{seyfioglu2024quiltllava} or expert-validated multiple-choice generation~\cite{sun2024pathmmu,sun2025pathbench}; PathQABench~\cite{lu2024pathchat} adds clinical context to one pathologist-selected region, OmniPathoVQA~\cite{chen2026omnipathovqa} covers thousands of entities from educational material, and VIPER~\cite{weishaupt2026viper} veterinary toxicologic pathology.
Recent ones probe grounding, counting and spatial relations~\cite{chen2026pathview,zhang2026pathovernier,zhong2025pathvg}, whether answers depend on the image at all~\cite{zhang2026pathbind,zhang2026cleanslide}, robustness to diagnostic wording~\cite{cheng2026pathlang} or hematopathology~\cite{wang2026pbsbench}.
Slide-level benchmarks derive questions from TCGA reports~\cite{chen2024wsivqa,chen2025slidechat,liang2025wsillava}, add pathologist-authored questions or diagnostic paths~\cite{buckley2025giant,liao2026evipathbench} and multi-slide or patient-level cases~\cite{liu2026pathoargus,ge2026planpath}; they require slide encoders~\cite{li2026beyondpath} or agentic navigation~\cite{chen2025pathagent,chen2026adaptivepath,yang2026pathnavigate,zhao2026slidebank}, also across magnifications~\cite{xu2026mmnavagent,xu2026pathpilot}; TCGA-derived splits can overlap with training folds~\cite{zhang2026leakage}, and LLM baselines hinge on patch size, count and magnification~\cite{weihrauch2026designchoices}.
Pathology's Last Exam~\cite{reitsam2025ple} reveals text-only cases block by block.
Closest to our setting, cross-scale benchmarks~\cite{phan2026scalevqa,phan2026pathscale} show two or three nested regions of one TCGA slide with LLM-generated, pathologist-validated questions; PathMMU-HR\textsuperscript{2}~\cite{sun2025cpathagent} asks about large regions; and DDxBench~\cite{weishaupt2025slideseek} gives ten expert-selected regions of each of 150 neoplastic slides for open-ended differential diagnosis.
Medical suites~\cite{hu2024omnimedvqa,chen2024gmaimmbench,zuo2025medxpertqa} include pathology as one modality; encoder benchmarks~\cite{majzoub2025histovl,ma2025pathbenchfm} score embeddings, not answers.
\bench differs in unit, breadth and probes: each item is a case of two to eight curated fields of one lesion (in atlas figure series, of one disease), across diagnostic pathology including non-neoplastic, benign and rare entities; options come from an ontology, not an LLM; and field-set ablations and finding verification test whether models use the fields when a case must simply be diagnosed (Sec.~\sref{supp:crossscale}).
Its distractors include each entity's classic look-alikes, so a correct choice requires separating entities that are confused in practice.

\paragraph{Multi-image evaluation.}
BLINK~\cite{fu2024blink}, MuirBench~\cite{wang2024muirbench} and MMIU~\cite{meng2024mmiu} show that general MLLMs struggle with relations across images, and MMBench~\cite{liu2024mmbench} introduced circular evaluation to expose answer-position bias.
Medical multi-image benchmarks find that MLLMs treat a sequence of images as separate instances~\cite{yu2025medframeqa} and degrade as images are added~\cite{xu2026medmultibench}, also on rare-disease case reports~\cite{ning2026mmrarebench}; microscopy benchmarks~\cite{lozano2024microbench,burgess2025microvqa} cover research imaging.
We bring these concerns to diagnostic pathology, where the images are related (one lesion, several scales) and integrating them is necessary.

\begin{table}[t]
\centering
\footnotesize
\renewcommand{\arraystretch}{0.90}
\setlength{\tabcolsep}{2.2pt}
\caption{\textbf{Pathology benchmarks for MLLMs} (selection; full comparison and size scopes in Table~\sref{tab:positioning-full}).
\emph{Size}: number of evaluation items. \emph{Exp.}: pathologist review; \emph{MF}: separate fields of one lesion; \emph{Case}: patient diagnosis; \emph{HN}: close distractors; \emph{Rob.}: circular evaluation or consistency under controlled input changes.
\pmark: partial; \xmark: not reported; --: not applicable.}
\label{tab:positioning}
\begin{tabular}{@{}lrccccc@{}}
\toprule
Benchmark & Size & Exp. & MF & Case & HN & Rob. \\
\midrule
\multicolumn{7}{@{}l}{\textit{Single image or region}} \\
PathVQA~\cite{he2020pathvqa} & 32.8k & \xmark & \xmark & \xmark & \xmark & \xmark \\
Quilt-VQA~\cite{seyfioglu2024quiltllava} & 1.3k & \pmark & \xmark & \xmark & \xmark & \xmark \\
PathQABench~\cite{lu2024pathchat} & 105 & \cmark & \xmark & \cmark & \pmark & \xmark \\
PathMMU~\cite{sun2024pathmmu} & 33.4k & \cmark & \xmark & \xmark & \pmark & \xmark \\
PathBench~\cite{sun2025pathbench} & 6.3k & \cmark & \xmark & -- & \pmark & \xmark \\
PathoVernier~\cite{zhang2026pathovernier} & 759 & \cmark & \xmark & \xmark & \xmark & \xmark \\
VIPER~\cite{weishaupt2026viper} & 1.3k & \cmark & \xmark & \xmark & \pmark & \pmark \\
\multicolumn{7}{@{}l}{\textit{Nested zoom path: 2--3 regions of one slide}} \\
Scale-VQA~\cite{phan2026scalevqa} & 4.7k & \cmark & \cmark & \pmark & \pmark & \xmark \\
PathScale-VQA~\cite{phan2026pathscale} & 10.4k & \cmark & \cmark & \pmark & \pmark & \xmark \\
\multicolumn{7}{@{}l}{\textit{Whole slide}} \\
SlideBench~\cite{chen2025slidechat} & 15.1k & \pmark & \xmark & \cmark & \pmark & \xmark \\
WSI-Bench~\cite{liang2025wsillava} & 180k & \pmark & \xmark & \cmark & \xmark & \xmark \\
MultiPathQA~\cite{buckley2025giant} & 934 & \pmark & \xmark & \cmark & \xmark & \xmark \\
DDxBench~\cite{weishaupt2025slideseek} & 150 & \cmark & \pmark & \cmark & -- & \xmark \\
PathoArgus-Bench~\cite{liu2026pathoargus} & 22.1k & \xmark & \xmark & \cmark & \xmark & \cmark \\
\midrule
\bench-Teaching (ours) & \nItems & \cmark & \cmark & \cmark & \cmark & \cmark \\
\bench-Atlas (ours) & \nItemsAtlas & \cmark & \pmark & \pmark & \cmark & \cmark \\
\bottomrule
\end{tabular}
\end{table}

\section{The \bench Benchmark}
\label{sec:benchmark}

\subsection{Source collection}
\label{sec:source}
\bench-Teaching is built from an in-house image library for pathology residency training; \bench-Atlas (\cref{sec:atlas}) adds cases from published atlases.
The library follows the residency curriculum and covers 18 organ systems.
Common diseases were selected from pathology textbooks, and the corresponding histological slides were retrieved and scanned into whole-slide images (WSIs).
Four pathologists independently collected fields at different magnifications from the same WSI, then jointly discussed the candidate fields to select the most representative images (\cref{fig:teaser}b).
Each case thus shows one entity in several fields of the same lesion, typically from low to high magnification.

\subsection{From a teaching library to benchmark cases}
\label{sec:construction}
The raw library contains 3,291 image files but only 2,586 distinct files, because the same case is filed under several organ systems, often under different names.
We therefore treat construction as a reconstruction problem (Sec.~\sref{supp:pipeline}).

\paragraph{Duplicates and cases.}
We first find every copy of a picture---exact, re-encoded, resized or colour-edited---by file and perceptual hashes and by pathology-CLIP~\cite{ikezogwo2023quilt} similarity confirmed at the pixel level, since two \emph{different} fields of one lesion routinely exceed a cosine of 0.985 (Sec.~\sref{supp:pipeline}); 2,512 distinct pictures remain.
Groups of fields that share a picture are copies of the same case and are merged, unless the library marks them as distinct cases; this gives 453 cases.
We exclude non-H\&E images (17) from the model inputs, cut three composites into their H\&E fields, and keep cases with at least two H\&E fields: \textbf{436 cases with 2,448 fields} form \bench-Teaching.

\paragraph{Field order.}
Every question states that the fields run from low to high power, but no single cue recovers this order reliably: nuclear scale fails on overviews, and the original photograph order does not carry over once copies of a case are merged.
We rank fields by a zero-shot pathology-CLIP magnification score adjusted by the original photograph order (Sec.~\sref{supp:pipeline}) and keep at most eight fields spanning the range; a pathologist then checked every case and corrected 76 orders (17\%), 45 of them by swapping two neighbouring fields of similar power.

\subsection{Label ontology}
\label{sec:ontology}
Library labels mix WHO editions and naming conventions; we map each of them to one of 407 canonical entities aligned with the WHO Classification of Tumours (5th edition) and, for non-neoplastic disease, standard textbook terminology.
Each entity records English and Chinese names, synonyms including outdated terms found in the library (\eg \emph{haemangiopericytoma} for solitary fibrous tumour), a behaviour class (non-neoplastic, benign, intermediate---borderline, \emph{in situ} or precursor---or malignant), a lineage, a family and an \emph{H\&E-sufficiency} prior: whether the exact entity is diagnosable from H\&E alone (Y), suggested but usually confirmed by ancillary tests (P), or not determinable without them (N; \eg TFE3-rearranged renal cell carcinoma).
Pictures filed under several organ-specific names (\eg one squamous cell carcinoma under cervix, vulva, vagina and skin) receive a site-agnostic label and no site hint; the other 400 of 436 cases give the specimen site in the prompt, as in practice.

\begin{figure}[t]
\centering
\includegraphics[width=\linewidth]{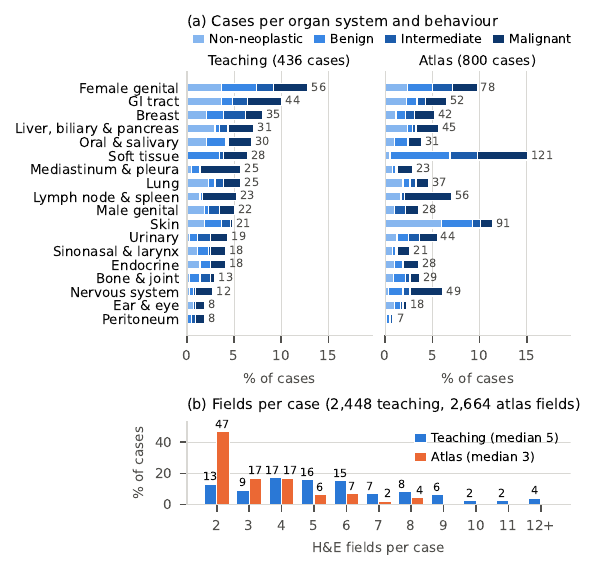}
\caption{\textbf{Composition of the two sources.} (a) Share of each source's cases per primary organ system, coloured by behaviour; numbers: cases. (b) Share of cases by number of H\&E fields; a teaching case keeps all its fields, of which at most eight are shown to the models.}
\label{fig:stats}
\end{figure}

\subsection{Label quality control}
\label{sec:verification}
Each label is the diagnosis under which the teaching pathologists filed the case, harmonised to the ontology.
An automated pre-review by a multimodal LLM agent (DeepSeek-V4.1-Flash, release 1017) inspected every case---its fields, label, site and the names under which the library filed it---and every A2 item, and flagged 64 cases (14.7\%) and 7 items; it changed nothing (the flags by type and the confirmation protocol: Sec.~\sref{supp:review}).
A pathologist re-examined every flagged case and item and decided whether to keep, correct or remove it; the 372 unflagged cases keep the library label.
Human review thus covers the labels of 64 cases and the field order of all 436 (\cref{sec:construction}).

\begin{figure*}[t]
\centering
\includegraphics[width=\textwidth]{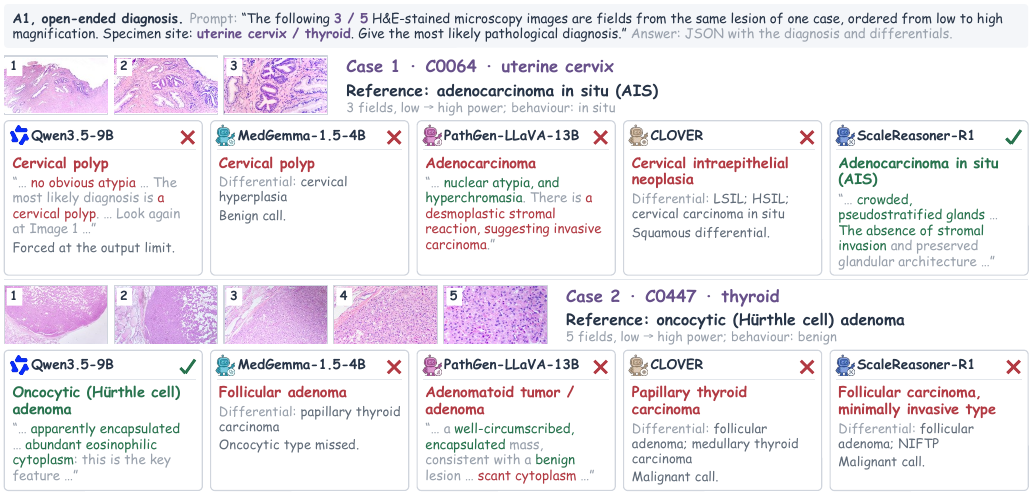}
\caption{\textbf{A1 diagnosis on two \bench cases.} Top: the prompt, identical for both cases except for the number of fields and the site. Each card gives the diagnosis, an excerpt of the response (green/red: findings consistent/inconsistent with the reference; grey: the rest) or the differential the model returned, and an editorial note. Ticks/crosses follow the A1 judge.}
\label{fig:qualitative}
\end{figure*}

\subsection{Statistics}
\cref{fig:stats} compares the two sources. \bench-Teaching has 165 malignant, 61 intermediate, 96 benign and 114 non-neoplastic cases.
The female genital tract, digestive tract and breast are the largest systems, mirroring the curriculum.
\bench-Atlas (\cref{sec:atlas}) has 293 malignant, 107 intermediate, 190 benign and 210 non-neoplastic cases and gives soft tissue, skin and the nervous system larger shares.
By the H\&E-sufficiency prior, 194 cases are expected to be fully diagnosable from H\&E, 221 suggestively so, and 21 require ancillary tests for the exact entity; we report results on all cases and on the H\&E-sufficient subset.
Further statistics are in Sec.~\sref{supp:dataset-detail}, and all entities are listed in Sec.~\sref{supp:ontology}.

\subsection{\bench-Atlas: a published second source}
\label{sec:atlas}
One teaching collection fixes one set of entities and one group's choice of fields.
To test whether our findings carry over, \bench-Atlas draws multi-field cases from 88 published pathology atlases, whose figures carry captions.
Panels that share a caption form a case of one lesion (387 cases); consecutive figures of one entity within a section form a case of one disease (413, of which 26 are stated to be one case), and the prompt says which.
Only H\&E panels are used, and each case is mapped from its caption to the ontology by two open LLMs (Qwen2.5-32B and Qwen3.5-9B) that must agree.
The ontology grows by 439 entities---WHO entities of each organ system and non-neoplastic diagnoses from the atlases' section titles---without changing any Teaching entity or item.
Taking one case per entity, common diagnoses first, gives \textbf{800 cases with 2,664 fields} (median 3) and 800 entities, 453 of which have no Teaching case (843 entities in all).
Fields are ordered by the magnification stated in the caption where there is one and as in \cref{sec:construction} otherwise.
An LLM read every case's caption, label and site, replaced 126 cases whose label was wrong, duplicated another entity or too generic, and corrected the site or name of 128 more; a pathologist then verified the labels.

\section{Tasks and Evaluation Protocol}
\label{sec:tasks}

Every item presents the H\&E fields of one case or, for the 387 atlas figure series, of one disease (at most eight, long side 1,024\,px) ordered from low to high power, numbered in the prompt, together with the specimen site when known.
\cref{tab:tasks} lists the tasks, five on each source.

\begin{table}[t]
\centering
\footnotesize
\setlength{\tabcolsep}{2.5pt}
\caption{\textbf{Task suite.} Items per source (Teach.: \bench-Teaching), counted before circular rotation. Chance is the vanilla (single-pass) expectation.}
\label{tab:tasks}
\begin{tabular}{llrrrl}
\toprule
 & Task & Teach. & Atlas & Chance & Metric \\
\midrule
A1 & Open-ended diagnosis & 436 & 800 & -- & HDS, Top-$k$ \\
A2 & Diagnosis, 10-way MCQ & 436 & 800 & 10\% & CircularEval \\
B & Behaviour triage, 4-way & 380 & 762 & 25\% & CircularEval \\
C & Ordinal grading, 2--4-way & 44 & -- & 1/$n$ & Circ., QWK \\
R1 & Finding verification, T/F\textsuperscript{$\dagger$} & 390 & 195 & 50\% & Stmt./item acc. \\
G1 & Finding location, 2--8-way & -- & 436 & 1/$n$ & CircularEval \\
\midrule
 & Total & \nItems & \nItemsAtlas & & \\
\bottomrule
\end{tabular}
\end{table}

\paragraph{A1: open-ended diagnosis.}
The model returns a diagnosis, two differentials and a confidence in JSON.
We score with a hierarchical diagnostic score (HDS): 1 for the reference entity, any of its synonyms or an accepted alternate; 0.5 for another entity of the same family; 0 otherwise.
The item score is the higher of an open-weights LLM judge (Qwen2.5-32B-Instruct, AWQ), which resolves free-text equivalence given the reference, its synonyms and family, and a deterministic ontology match that gives family credit only within the same behaviour class. Against blinded 2/1/0 grades of \judgeN\ responses, the final A1 grade agrees exactly on \judgeExactFinal\% (quadratic-weighted $\kappa$ \judgeKappaFinal; LLM judge alone \judgeKappaJudge) and is stricter on average (mean grade \judgeMeanFinal\ vs.\ \judgeMeanRater), so A1 scores err on the low side.
We also report Top-1 exact and Top-3 (diagnosis or differentials) accuracy.

\paragraph{A2: diagnosis as multiple choice.}
Nine distractors are drawn from the ontology in priority order: curated differential partner, same family, same organ system with the same lineage and behaviour, same system with the same lineage, same system, same lineage.
Every classic look-alike of the entity---\nLookalikeItems\ cases have one among their options, from 63 curated textbook differentials and 40 pairs the library itself sets side by side---is always among the distractors unless the exclusions below forbid it, so that each case's classic confusion is tested.
Distractors may never be a synonym, parent or subtype, organ-site variant or a partner that H\&E cannot separate (Sec.~\sref{supp:distractors}), and grading siblings are left to C; no model is used to mine them, so no evaluated model is adversarially disadvantaged.

\paragraph{B: behaviour triage.}
Non-neoplastic, benign, intermediate or malignant, on the cases whose behaviour code is uncontroversial (we exclude entities such as GIST or thymoma whose coding is debated).
We also derive the \emph{critical miss rate}: malignant or \emph{in situ} cases called benign or non-neoplastic.

\paragraph{C: ordinal grading.}
Options are all levels of the case's ordinal group, two to four per group: 17 groups such as DCIS nuclear grade, phyllodes tumour grade, Gleason pattern, squamous intraepithelial lesion grade, astrocytoma grade and endometrial dating.
Besides circular accuracy we report quadratic weighted $\kappa$ (QWK) on the levels.

\paragraph{R1: finding verification.}
To ask whether models see the morphology behind a diagnosis, 195 cases (all 194 whose entity is diagnosable from H\&E alone, plus the case of \cref{fig:teaser}c) carry a reference route from low-power architecture to high-power cytology, the differential and the conclusion.
Each case gives two items, one per level, of four statements to judge, two true and two false: true ones are essential criteria of the labelled entity, false ones defining features of its look-alikes or distractors that the label excludes, so every truth value follows from the expert label, not from reading the image.
\textsuperscript{$\dagger$}Routes and statements were generated by an LLM (DeepSeek-V4.1-Flash, release 1017) from the label and WHO criteria and subsequently reviewed by a pathologist; they never name the entity or the site, and wording heuristics score at chance (two shortest statements true: 50\%; negated statements true: 51\%).
We report statement accuracy and the share of items with all four right, by level, and a text-only baseline.

\paragraph{Atlas tasks.}
\bench-Atlas has A1 and A2 as above, B on the cases with a certain behaviour code (the ICD-O code of a WHO entity, a confident ontology code, or the agreement of both LLMs for a new entity), and two tasks taken from the captions.
In R1-Atlas a true statement restates a finding the caption states and a false one contradicts it minimally; a statement is kept only if its source span occurs in the caption and a second LLM reaches the same truth value from the caption alone (LLM-written, verified by a pathologist).
G1 asks which field shows a finding that the caption ties to one panel or figure, with the entity's name removed (the options are the case's fields, with CircularEval over them).

\paragraph{Circular evaluation and statistics.}
Each multiple-choice item is asked once per rotation of its options (ten for A2, four for B, two to four for C, one per field for G1); an item counts as correct only if every rotation is answered correctly~\cite{liu2024mmbench}.
We also report vanilla (mean over rotations) accuracy.
95\% confidence intervals and paired comparisons use bootstrap resamples of cases (2,000; 10,000 for the field-set ablations).

\section{Experiments}
\label{sec:experiments}
\newcommand{\diagNModels}{28}
\newcommand{\diagNCases}{436}
\newcommand{\diagNLookCases}{178}
\newcommand{\diagLookObserved}{26.6}
\newcommand{\diagLookExpected}{14.7}
\newcommand{\diagLookMin}{18.5}
\newcommand{\diagLookMax}{37.7}
\newcommand{\diagContentMin}{0.2}
\newcommand{\diagContentMax}{21.8}
\newcommand{\diagAtlNModels}{28}
\newcommand{\diagAtlNCases}{800}
\newcommand{\diagAtlNLookCases}{99}
\newcommand{\diagAtlLookObserved}{24.4}
\newcommand{\diagAtlLookExpected}{13.0}
\newcommand{\diagAtlLookMin}{11.1}
\newcommand{\diagAtlLookMax}{40.4}
\newcommand{\diagAtlContentMin}{0.1}
\newcommand{\diagAtlContentMax}{21.4}

\subsection{Setup}
\label{sec:setup}
\paragraph{Models.}
We evaluate \nMLLMs\ open MLLMs, with selected models listed in \cref{tab:main} and the complete set with references in Table~\sref{tab:main-detail}: general-purpose models that accept multiple images (Qwen, InternVL, MiniCPM, MiMo and GLM families, including reasoning variants), medical and pathology models; the single-image LLaVA-Med, Quilt-LLaVA and PathGen-LLaVA receive the fields as one numbered mosaic.
We also evaluate four proprietary models: Claude Sonnet 5.5, Claude Haiku 5.5, GPT-5.6-sol and GPT-5.6-luna.
\paragraph{Inference.}
Open models run with vLLM, greedy decoding and up to 1,024 new tokens (4,096 for reasoning models, made to answer if still reasoning at the limit~\cite{yang2025qwen3}); Qwen2.5-VL-32B runs in Transformers with 4-bit NF4 weights to fit a 24\,GB GPU, HuatuoGPT-Vision and Hulu-Med with their official code (Sec.~\sref{supp:parsing}).
As in MMBench~\cite{liu2024mmbench}, the instruction follows the options, a deterministic parser maps responses to options and the few it cannot map go to the judge LLM; unmapped answers count as wrong.
R1 answers are read as four booleans; every model also answers R1 from the text alone; the field-set ablations of \cref{sec:multiview} were run for the open models only.

\begin{table*}[t]
\centering
\scriptsize
\renewcommand{\arraystretch}{0.94}
\setlength{\tabcolsep}{2.5pt}
\caption{\textbf{Results on \bench-Teaching and \bench-Atlas} (\%; tasks and metrics as in \cref{sec:tasks}). A2, B, C, G1: CircularEval; A2\textsubscript{v}: single-pass accuracy. Miss: critical miss rate. C\textsubscript{$\kappa$}: QWK. $\Delta_{\text{img}}$: R1 with minus without the images ($^\ast$: 95\% paired bootstrap interval excludes 0). $^\dagger$Single-image model given a mosaic of the fields. --: not applicable. \textbf{Bold}/\underline{underlined}: best/second best per column except $\Delta_{\text{img}}$, separately for proprietary and open models. 95\% half-widths reach 4.8 points on \bench-Teaching and 3.4 on \bench-Atlas; per-model intervals: Tables~\sref{tab:main-detail} and~\sref{tab:atlas-detail}.}
\label{tab:main}
\begin{tabular}{lccccccccc@{\hspace{8pt}}cccccccc}
\toprule
 & \multicolumn{9}{c}{\bench-Teaching} & \multicolumn{8}{c}{\bench-Atlas} \\
\cmidrule(lr){2-10}\cmidrule(lr){11-18}
 & A1 & A2 & A2\textsubscript{v} & B & Miss$\downarrow$ & C & C\textsubscript{$\kappa$} & R1 & $\Delta_{\text{img}}$ & A1 & A2 & A2\textsubscript{v} & B & Miss$\downarrow$ & R1 & $\Delta_{\text{img}}$ & G1 \\
\midrule
Random choice & -- & 0.0 & 10.1 & 0.5 & 49.7 & 11.4 & 0.08 & -- & -- & -- & 0.0 & 9.9 & 0.9 & 50.9 & -- & -- & 9.2 \\
\midrule
\multicolumn{18}{l}{\textit{Proprietary MLLMs}} \\
Claude Sonnet 5.5 & \textbf{64.2} & \textbf{61.7} & \textbf{78.3} & \textbf{75.5} & \underline{11.4} & \textbf{79.5} & \textbf{0.81} & \textbf{93.2} & +10.2$^\ast$ & \textbf{41.6} & \textbf{49.1} & \textbf{68.5} & \textbf{61.5} & \textbf{21.6} & \textbf{80.8} & +6.0$^\ast$ & \textbf{61.5} \\
Claude Haiku 5.5 & \underline{43.9} & 39.0 & 61.0 & \underline{57.4} & \textbf{10.3} & 54.5 & 0.55 & 83.2 & +19.5$^\ast$ & \underline{28.8} & 29.9 & 52.9 & \underline{47.0} & \underline{25.4} & 73.6 & +0.0 & \underline{48.2} \\
GPT-5.6-sol & 43.5 & \underline{49.3} & \underline{67.4} & 55.5 & 19.4 & \underline{61.4} & \underline{0.64} & \underline{92.9} & +12.6$^\ast$ & 27.6 & \underline{37.1} & \underline{57.1} & 41.9 & 35.4 & \underline{78.6} & +5.6$^\ast$ & 45.6 \\
GPT-5.6-luna & 33.5 & 31.9 & 57.3 & 51.1 & 33.6 & 56.8 & 0.61 & 87.1 & +15.4$^\ast$ & 19.8 & 24.1 & 46.6 & 32.7 & 46.4 & 73.1 & +8.5$^\ast$ & 39.9 \\
\midrule
\multicolumn{18}{l}{\textit{General MLLMs (open)}} \\
Qwen2.5-VL-7B~\cite{bai2025qwen25vl} & 11.7 & 6.0 & 28.1 & 18.7 & 15.8 & 9.1 & 0.06 & 63.0 & +6.3$^\ast$ & 10.9 & 9.8 & 28.7 & 17.2 & 19.8 & 63.6 & +7.9$^\ast$ & \underline{39.4} \\
InternVL3.5-8B~\cite{wang2025internvl35} & 18.3 & 11.0 & 34.7 & 30.5 & 15.0 & 27.3 & 0.16 & 66.5 & +3.4$^\ast$ & 12.3 & 9.4 & 31.4 & 19.9 & 22.9 & 67.6 & +4.9$^\ast$ & 36.7 \\
Qwen3-VL-8B~\cite{bai2025qwen3vl} & 11.9 & 10.6 & 30.2 & 30.5 & 6.7 & 27.3 & 0.02 & 63.3 & \textminus{}0.6 & 12.3 & 12.1 & 30.3 & 17.1 & 21.7 & 64.6 & +3.7$^\ast$ & 32.8 \\
Qwen3.5-9B~\cite{qwen2026qwen35} & 29.7 & \textbf{17.9} & \textbf{50.6} & 31.8 & 45.1 & \textbf{36.4} & \textbf{0.42} & \textbf{81.5} & +16.3$^\ast$ & 18.6 & \textbf{14.2} & \textbf{41.7} & 28.1 & 44.0 & \textbf{69.5} & +5.1$^\ast$ & \textbf{42.7} \\
MiMo-VL-7B~\cite{xiaomi2025mimovl} & 15.9 & 7.8 & 29.9 & 26.8 & 22.5 & 29.5 & 0.13 & 66.1 & +9.7$^\ast$ & 12.2 & 8.9 & 28.3 & 23.6 & 47.9 & 67.3 & +6.7$^\ast$ & 33.5 \\
GLM-4.1V-9B-Thinking~\cite{glmv2025glm41v} & 19.4 & 12.6 & 36.3 & 31.1 & 24.7 & 29.5 & 0.15 & \underline{74.7} & +9.0$^\ast$ & 15.4 & \underline{12.6} & \underline{33.6} & \underline{31.6} & 42.6 & \underline{68.3} & +1.5 & 36.2 \\
\midrule
\multicolumn{18}{l}{\textit{Medical MLLMs}} \\
HuatuoGPT-Vision-7B (Q2.5-VL)~\cite{chen2024huatuogptvision} & 12.0 & 8.0 & 27.3 & 30.8 & 6.1 & 13.6 & 0.05 & 57.8 & +3.9$^\ast$ & 11.8 & 9.4 & 29.2 & 24.7 & 13.9 & 64.9 & +14.4$^\ast$ & 31.7 \\
Lingshu-7B~\cite{lasa2025lingshu} & 21.0 & 5.7 & 33.6 & \underline{34.2} & 35.4 & \textbf{36.4} & 0.32 & 57.8 & +2.4$^\ast$ & 14.8 & 7.0 & 29.1 & \textbf{32.3} & 30.8 & 61.3 & +0.0 & \underline{39.4} \\
OctoMed-7B~\cite{ossowski2026octomed} & 14.1 & 4.8 & 32.1 & 12.9 & 21.8 & 20.5 & 0.02 & 62.2 & +5.2$^\ast$ & 8.4 & 4.8 & 26.4 & 15.1 & 40.6 & 60.4 & \textminus{}1.0 & 22.5 \\
MedGemma-1.5-4B~\cite{sellergren2025medgemma} & 11.5 & 2.5 & 21.5 & 18.9 & 15.2 & 20.5 & 0.15 & 51.7 & +0.6 & 5.9 & 1.8 & 16.4 & 17.7 & 19.6 & 57.8 & +2.7 & 11.0 \\
Hulu-Med-7B~\cite{jiang2025hulumed} & 17.3 & \underline{13.3} & 41.2 & 33.4 & 19.7 & 13.6 & 0.28 & 63.3 & +4.9$^\ast$ & 9.6 & 4.8 & 29.5 & 20.7 & 29.3 & 63.2 & \textminus{}1.2 & 21.6 \\
\midrule
\multicolumn{18}{l}{\textit{Pathology MLLMs}} \\
Quilt-LLaVA-7B$^\dagger$~\cite{seyfioglu2024quiltllava} & 3.8 & 1.6 & 17.7 & 1.1 & 40.0 & 2.3 & \textminus{}0.08 & 49.4 & +3.4$^\ast$ & 2.0 & 0.6 & 14.5 & 10.6 & 62.9 & 46.4 & +12.3$^\ast$ & 15.1 \\
PathGen-LLaVA-13B$^\dagger$~\cite{sun2025pathgen} & 17.4 & 4.4 & 20.7 & 13.9 & 52.4 & 4.5 & 0.06 & 50.1 & \textminus{}0.4 & 8.5 & 1.9 & 18.7 & 22.8 & 72.5 & 49.7 & \textminus{}1.4 & 16.1 \\
CLOVER~\cite{chen2024clover} & 12.5 & 4.1 & 25.1 & 20.5 & 22.9 & \textbf{36.4} & 0.18 & 59.9 & +3.7$^\ast$ & 11.6 & 7.2 & 25.7 & 14.2 & 42.4 & 63.6 & +6.4$^\ast$ & 16.1 \\
Patho-R1-3B~\cite{zhang2025pathor1} & 13.4 & 0.2 & 9.8 & 2.4 & 36.6 & 6.8 & \textbf{0.42} & 12.1 & +8.5$^\ast$ & 14.3 & 0.1 & 11.5 & 3.1 & 40.8 & 14.4 & +11.4$^\ast$ & 2.1 \\
Patho-R1-7B~\cite{zhang2025pathor1} & \textbf{34.7} & 7.1 & 36.2 & 33.2 & 11.1 & 29.5 & 0.28 & 36.1 & +8.5$^\ast$ & \textbf{21.5} & 7.6 & 28.9 & 19.4 & 18.7 & 32.2 & +6.7$^\ast$ & 20.6 \\
ScaleReasoner-R1~\cite{phan2026scalevqa} & \underline{32.5} & 10.8 & \underline{41.8} & \textbf{34.5} & 6.1 & 25.0 & \underline{0.33} & 29.2 & \textminus{}2.0 & \underline{20.1} & 8.6 & 32.5 & 28.7 & \underline{9.3} & 26.3 & +6.2$^\ast$ & 18.8 \\
\bottomrule
\end{tabular}
\end{table*}

\begin{figure}[t]
\centering
\includegraphics[width=\linewidth]{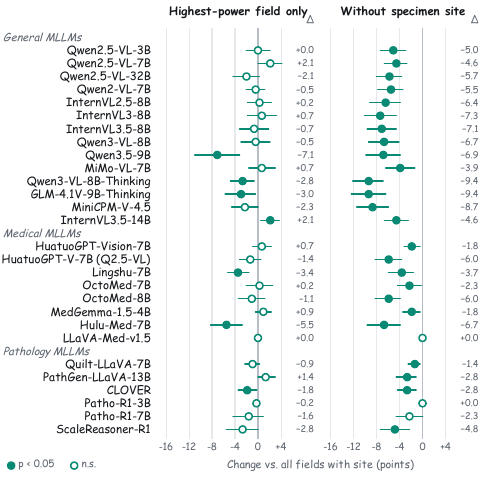}
\caption{\textbf{Do models use the fields?} Change of the open models' A2 circular accuracy when only the highest-power field is given (left) or the specimen site is removed (right), against all fields with the site; paired case bootstrap, 95\% CI; filled: $p{<}0.05$; numbers: change in points.}
\label{fig:fielduse}
\end{figure}

\subsection{Main results}
\label{sec:main}
\cref{tab:main} reports all tasks on both sources (\bench-Atlas: \cref{sec:atlas-results}). Ranges for \bench-Teaching below cover all \nMLLMs\ open models in Table~\sref{tab:main-detail}, and the \nProp\ proprietary models separately.
\textbf{Diagnosis is unstable.}
Single-pass ten-way accuracy ranges from \minMLLMAtwoVan\% to \maxMLLMAtwoVan\% for open and from \minPropAtwoVan\% to \maxPropAtwoVan\% for proprietary models (chance 10\%), but requiring the correct answer under all ten option rotations leaves \minMLLMAtwoCirc--\maxMLLMAtwoCirc\% and \minPropAtwoCirc--\maxPropAtwoCirc\% (Table~\sref{tab:bias}).
Across the \diagNModels\ complete open runs (Fig.~\sref{fig:diagnostic-reliability}a), the same diagnosis is returned in all ten rotations for only \diagContentMin--\diagContentMax\% of cases.
Open-ended diagnosis (A1) shows the same gap without options: hierarchical scores range from \minMLLMAoneHDS\ to \maxMLLMAoneHDS\ for open and from \minPropAoneHDS\ to \maxPropAoneHDS\ for proprietary models, the exact entity is named in \minMLLMAoneExact--\maxMLLMAoneExact\% and \minPropAoneExact--\maxPropAoneExact\% of cases, and \minMLLMAoneWrongBeh--\maxMLLMAoneWrongBeh\% of the open models' answers get the behaviour wrong (proprietary: \minPropAoneWrongBeh--\maxPropAoneWrongBeh\%).
Among open models the pathology reasoning models lead A1 (\aoneFirstName\ \aoneFirstHDS, \aoneSecondName\ \aoneSecondHDS; next \aoneThirdName\ \aoneThirdHDS) but are no more consistent on A2 than the general models: open-ended recall and a stable choice among close alternatives are different skills. No open model leads every task, a complementarity also seen among frontier models in scientific reasoning~\cite{guo2026sciorch}.
\textbf{Triage errs in both directions.}
Between \minMLLMMiss\% and \maxMLLMMiss\% of malignant or \emph{in situ} cases are triaged as benign or non-neoplastic by open models (\minPropMiss--\maxPropMiss\% by proprietary).
The lowest open-model miss rate does not mean better triage: \lowMissName\ misses only \lowMiss\% because it labels \lowMissOvercall\% of the benign and non-neoplastic cases malignant or borderline.
\cref{fig:qualitative} shows one example of each error in open-ended diagnosis: an adenocarcinoma \emph{in situ} called a benign polyp (Case 1, MedGemma-1.5-4B) and a benign oncocytic adenoma called a carcinoma (Case 2, CLOVER and ScaleReasoner-R1).
\textbf{Grading is weak for open models.}
On ordinal grading (C) their circular accuracy is at most \maxMLLMCCirc\% (random choice: \randomCCirc\%; \nCBelowRandom\ of \nCReported\ fall below it) and quadratic-weighted $\kappa$ ranges from \minMLLMCqwk\ to \maxMLLMCqwk; proprietary models reach \minPropCCirc--\maxPropCCirc\% and $\kappa$ \minPropCqwk--\maxPropCqwk.
\textbf{Open models barely verify findings.}
On R1 open models judge a median \medMLLMRone\% of the statements correctly (best \maxMLLMRone\%; chance 50\%) and none gets all four of an item right in more than \maxMLLMRoneItem\% of items, at low and high power alike (\meanRoneLow\% vs.\ \meanRoneHigh\%); the proprietary models judge \minPropRone--\maxPropRone\% correctly and all four in \minPropRoneItem--\maxPropRoneItem\%.

\subsection{Do models use the fields?}
\label{sec:multiview}
\cref{fig:fielduse} re-runs A2 for the open models with the same cases and options but less evidence (all values: Table~\sref{tab:multiview}).
Answering from the highest-power field alone changes circular accuracy by \dltMinSingle\ to \dltMaxSingle\ points, significantly lowering it for \dltNsigwDownSingle\ of \dltNwSingle\ models and raising it for \dltNsigwUpSingle.
Removing the one-line specimen site, in contrast, lowers it for \dltNsigwNosite\ of \dltNwNosite\ models, by up to \dltMagHiNosite\ points---more than half of what most of them reach with it.
The images are not ignored altogether: on R1 (Fig.~\sref{fig:r1}) they raise statement accuracy over the text alone for \nRoneGainSigPos\ of \nRoneGain\ open models and lower it for \nRoneGainSigNeg, but by a median \medRoneGain\ points (max \maxRoneGain); much of what they get right is in the text.
The proprietary models, not re-run with fewer fields, gain more from the images on R1: \minPropRoneGain\ to \maxPropRoneGain\ points, significantly for all \nPropRoneGain, over \minPropRoneTxt--\maxPropRoneTxt\% from the text alone.

\subsection{Where do models fail?}
\textbf{Look-alikes attract the errors.}
For \diagNLookCases\ cases a classic look-alike of the entity is among the nine distractors; valid wrong answers on these items pick a look-alike \diagLookObserved\% of the time (\diagLookMin--\diagLookMax\% across models), against \diagLookExpected\% if they spread evenly over the distractors (Fig.~\sref{fig:diagnostic-reliability}c; means across the \diagNModels\ open models); the \nLookProp\ proprietary models pick a look-alike \lookErrPropMin--\lookErrPropMax\% of the time.
Single-pass accuracy falls from \famDistrZero\% for items without a distractor from the reference's own family to \famDistrFivePlus\% for items with five or more.
\textbf{Difficulty follows the annotation.}
\SuffBetterText\ open models answer cases diagnosable from H\&E alone more accurately than cases where H\&E is only suggestive (\meanSuffY\% vs.\ \meanSuffP\%), and \sixWorseText\ answer cases with six or more fields less accurately than two-field cases (\meanFieldsSix\% vs.\ \meanFieldsTwo\%; Table~\sref{tab:subsets}); harder teaching cases get more fields, so the fields show no benefit rather than harm.
The hardest organ systems are \sysHardAName\ and \sysHardBName, the easiest \sysEasyAName\ and \sysEasyBName, and A1 errors are mostly of the wrong behaviour class or another family of the same lineage, not near misses (Fig.~\sref{fig:per_system}, Table~\sref{tab:a1}).

We conducted a reader study comparing model performance with that of residents and attendings; details are provided in Sec.~\sref{supp:review}.

\subsection{Does the picture hold on published atlases?}
\label{sec:atlas-results}
The right half of \cref{tab:main} repeats the evaluation on \bench-Atlas, and Fig.~\sref{fig:replication} sets each model's two scores side by side; ranges cover all models in Table~\sref{tab:atlas-detail}.
\textbf{The ranking broadly replicates, the level does not.}
Rankings agree between the sources ($\rho=\atlRhoAtwo$, 95\% CI \atlRhoAtwoLo--\atlRhoAtwoHi, for A2 over \atlNAtwo\ models; \atlRhoAone\ for A1), though less so within the open models alone (\atlRhoAtwoOpen\ on A2), and proprietary models stay ahead on A2 (\atlAtlAtwoPropLo--\atlAtlAtwoPropHi\% vs.\ \atlAtlAtwoOpenLo--\atlAtlAtwoOpenHi\%).
We expected atlas figures, chosen to be typical, to be easier.
Instead, on the \atlNShared\ entities in both sources proprietary models change by \atlSharedPropAtwo\ points on A2 and \atlSharedPropAone\ on A1 (Atlas minus Teaching), and open models by \atlSharedOpenAtwo\ and \atlSharedOpenAone.
Fewer fields do not explain this alone: at equal numbers of fields the proprietary models change by \atlDropTwo\ points on A2 with two fields and by \atlDropFive\ with five or more.
The sources also differ in image reproduction and in the distractor pool, which this comparison cannot separate.
\textbf{New entities are harder to name, not to choose.}
Against the shared entities, the \atlNOnly\ Atlas-only entities score \atlOnlyPropAone\ points on A1 for proprietary and \atlOnlyOpenAone\ for open models, significantly lower for \atlOnlyDownAone\ of \atlOnlyNAone\ models, but A2 barely differs (\atlOnlyPropAtwo\ and \atlOnlyOpenAtwo; significantly lower for \atlOnlyDownAtwo\ of \atlOnlyNAtwo).
\textbf{Instability and missed malignancies recur.}
Requiring all ten rotations keeps \atlKeptOpen\% of the single-pass accuracy of open models (Teaching: \atlKeptTeachOpen\%) and \atlKeptProp\% of that of proprietary models (\atlKeptTeachProp\%).
Critical misses are more frequent than on \bench-Teaching (means of \atlMissPropMean\% vs.\ \atlMissTeachPropMean\% for proprietary and \atlMissOpenMean\% vs.\ \atlMissTeachOpenMean\% for open models).
Non-neoplastic cases are not harder than neoplastic ones, contrary to our expectation: the difference on A2 is \atlNNAtwoProp\ points for proprietary and \atlNNAtwoOpen\ for open models.
\textbf{Fields weigh as much as the site here.}
For the \atlSingleN\ open models with both ablations on both sources, on cases with three or more fields (\atlThreePlusCases\ on Atlas, \atlTeachThreePlusCases\ on Teaching), the highest-power field alone changes A2 by \atlSingleMean\ points on Atlas against \atlTeachSingleMean\ on Teaching (significantly lower for \atlSingleDown\ vs.\ \atlTeachSingleDown\ models), and removing the site by \atlNositeMean\ against \atlTeachNositeMean\ (\atlNositeDown\ vs.\ \atlTeachNositeDown).
The finding of \cref{sec:multiview} therefore holds on the teaching cases but does not carry over in full: on atlas cases the further fields count about as much as the site.
On G1 the options rotate while the images keep their order, so naming one image throughout is right on \atlGoneChanceVan\% of the items (independent guesses: \atlGoneChance\%); the 95\% interval lies above this for \atlGonePropAboveFixed\ of \atlGonePropN\ proprietary (\atlGonePropLo--\atlGonePropHi\%) but only \atlGoneOpenAboveFixed\ of \atlGoneOpenN\ open models (\atlGoneOpenLo--\atlGoneOpenHi\%; Table~\sref{tab:atlas-detail}).
R1-Atlas stays largely answerable from the text (up to \atlTextOnlyHi\%); the images add a mean \atlGainPropMean\ points for proprietary and \atlGainOpenMean\ for open models, significantly raising accuracy for \atlGainPropSigPos\ of \atlGainPropN\ and \atlGainOpenSigPos\ of \atlGainOpenN\ and lowering it for \atlGainOpenSigNeg\ open model.

\section{Discussion and Limitations}
\label{sec:discussion}

\paragraph{What \bench measures.}
Teaching cases and atlas figures show fields chosen to illustrate an entity, easier than those a pathologist must find on an unselected slide; \bench thus measures \emph{textbook-level} reasoning across fields, necessary but not sufficient for diagnostic use.
Most entities have a single case, so we report results by organ system, behaviour and task rather than per entity.

\paragraph{What the results imply.}
Accuracy alone overstates what a model reads: rotation consistency and field and site ablations belong beside it.

\paragraph{Limitations of the protocol.}
Fields come in a fixed, stated order, so field selection is untested; only H\&E and site are given; R1 truth follows from labels (Teaching) or captions (Atlas), not annotated regions.

\paragraph{Contamination.}
No teaching or atlas image has a near-duplicate among 228,899 public pathology images (Sec.~\sref{supp:results}), though a whole-image search misses crops.

\section{Conclusion}
\bench turns a residency teaching library and published atlases into a benchmark of diagnosis from several fields of one lesion or, in atlas figure series, of one disease.
Across \nMLLMs\ open and \nProp\ proprietary MLLMs, diagnoses change with the order of the options, extra fields matter far less than a one-line clinical hint on teaching cases and about as much on atlas cases with three or more fields, and findings are verified barely above chance by most open models and far better by proprietary ones, whose diagnoses are nonetheless unstable: current MLLMs do not yet read cases reliably.
Rotated options and reduced field sets show what the answers rest on.

{
    \small
    \linespread{0.98}\selectfont
    \setlength{\bibsep}{0pt plus 1pt}
    \bibliographystyle{ieeenat_fullname}
    \bibliography{main}
}

\maketitlesupplementary
\setcounter{section}{0}\setcounter{table}{0}\setcounter{figure}{0}\setcounter{equation}{0}
\renewcommand{\thesection}{S\arabic{section}}
\renewcommand{\thetable}{S\arabic{table}}
\renewcommand{\thefigure}{S\arabic{figure}}
\renewcommand{\theequation}{S\arabic{equation}}
\renewcommand{\theHsection}{S\arabic{section}}
\renewcommand{\theHtable}{S\arabic{table}}
\renewcommand{\theHfigure}{S\arabic{figure}}
\renewcommand{\theHequation}{S\arabic{equation}}

\section{Comparison with pathology benchmarks}
\label{supp:positioning}
\Cref{tab:positioning-full} extends Table~1\ofmain to all 20 benchmarks we surveyed, with venue, visual input, reported size and QA construction.

\begin{table*}[t]
\centering
\footnotesize
\setlength{\tabcolsep}{2.5pt}
\renewcommand{\arraystretch}{1.12}
\caption{\textbf{Pathology benchmarks for MLLMs: full version of Table~1\ofmain.}
Visual input, reported dataset size, QA construction and evaluation properties are compared; subset scopes and special input settings are specified below.}
\label{tab:positioning-full}
\begin{tabularx}{\textwidth}{@{}>{\raggedright\arraybackslash}p{3.35cm}>{\raggedright\arraybackslash}p{1.45cm}>{\raggedright\arraybackslash}X>{\raggedleft\arraybackslash}p{1.75cm}>{\raggedright\arraybackslash}p{2.05cm}ccccc@{}}
\toprule
Benchmark & Venue & Visual input & Size & \makecell[l]{QA\\construction} & Expert & \makecell{Multi-\\field} & \makecell{Case\\dx} & \makecell{Hard\\neg.} & Robust \\
\midrule
\multicolumn{10}{@{}l}{\textit{Single-image and region-based benchmarks}} \\
PathVQA~\cite{he2020pathvqa} & arXiv'20 & 1 image & 32.8k QA & Rule-based & \xmark & \xmark & \xmark & \xmark & \xmark \\
Quilt-VQA~\cite{seyfioglu2024quiltllava} & CVPR'24 & 1 image & 1.3k QA & LLM-generated & \pmark & \xmark & \xmark & \xmark & \xmark \\
PathQABench~\cite{lu2024pathchat} & Nature'24 & 1 ROI & 105 QA & Expert-written & \cmark & \xmark & \cmark & \pmark & \xmark \\
PathMMU~\cite{sun2024pathmmu} & ECCV'24 & 1 image & 33.4k QA & LLM-generated & \cmark & \xmark & \xmark & \pmark & \xmark \\
PathBench~\cite{sun2025pathbench} & TMI'25 & 1 image / 1 WSI & 6.3k QA & NR & \cmark & \xmark & -- & \pmark & \xmark \\
PathMMU-HR\textsuperscript{2}~\cite{sun2025cpathagent} & NeurIPS'25 & 1 ROI & 1.7k QA & LLM-generated & \cmark & \xmark & \xmark & \pmark & \xmark \\
PathBind~\cite{zhang2026pathbind} & arXiv'26 & 1 image & 2.6k items & Mixed & \cmark & \xmark & \xmark & \pmark & \xmark \\
PathoVernier~\cite{zhang2026pathovernier} & arXiv'26 & 1 patch & 759 QA & Mixed & \cmark & \xmark & \xmark & \xmark & \xmark \\
PathView-Bench~\cite{chen2026pathview} & NeurIPS'26 & 1 ROI / 1 WSI & 308k items & Rule-based & \pmark & \xmark & \xmark & \xmark & \xmark \\
PBSBench~\cite{wang2026pbsbench} & CVPR-F'26 & 1 cell ROI / 1 smear WSI & 8.3k QA & LLM-generated & \pmark & \xmark & \pmark & \xmark & \xmark \\
VIPER~\cite{weishaupt2026viper} & NeurIPS'26 & 1 ROI & 1.3k QA & Mixed & \cmark & \xmark & \xmark & \pmark & \pmark \\
\midrule
\multicolumn{10}{@{}l}{\textit{Nested fields at multiple magnifications}} \\
Scale-VQA~\cite{phan2026scalevqa} & MICCAI'26 & 3 nested ROIs & 4.7k QA & LLM-generated & \cmark & \cmark & \pmark & \pmark & \xmark \\
PathScale-VQA~\cite{phan2026pathscale} & arXiv'26 & 2--3 nested ROIs & 10.4k QA & LLM-generated & \cmark & \cmark & \pmark & \pmark & \xmark \\
\midrule
\multicolumn{10}{@{}l}{\textit{Whole-slide and case-based benchmarks}} \\
WSI-VQA~\cite{chen2024wsivqa} & ECCV'24 & 1 WSI & 8.7k QA & Mixed & \pmark & \xmark & \cmark & \xmark & \xmark \\
SlideBench~\cite{chen2025slidechat} & CVPR'25 & 1 WSI & 15.1k QA & LLM-generated & \pmark & \xmark & \cmark & \pmark & \xmark \\
WSI-Bench~\cite{liang2025wsillava} & ICCV'25 & 1 WSI & 180k QA & LLM-generated & \pmark & \xmark & \cmark & \xmark & \xmark \\
MultiPathQA~\cite{buckley2025giant} & arXiv'25 & 1 WSI & 934 QA & Mixed & \pmark & \xmark & \cmark & \xmark & \xmark \\
DDxBench~\cite{weishaupt2025slideseek} & arXiv'25 & 1 WSI / 10 ROIs & 150 cases & Rule-based & \cmark & \pmark & \cmark & -- & \xmark \\
EviPathBench~\cite{liao2026evipathbench} & arXiv'26 & 1 ROI / 1 WSI & 20.6k QA & Expert-written & \cmark & \xmark & \pmark & \pmark & \xmark \\
PathoArgus-Bench~\cite{liu2026pathoargus} & arXiv'26 & ${\geq}1$ WSI per case & 22.1k QA & Rule-based & \xmark & \xmark & \cmark & \xmark & \cmark \\
\midrule
\rowcolor{blue!5}
\bench-Teaching (ours) & -- & 2--8 fields per case & \nItems\ items & Mixed & \cmark & \cmark & \cmark & \cmark & \cmark \\
\rowcolor{blue!5}
\bench-Atlas (ours) & -- & 2--8 fields per case & \nItemsAtlas\ items & Mixed & \cmark & \pmark & \pmark & \cmark & \cmark \\
\bottomrule
\end{tabularx}
\par\vspace{4pt}
\begin{minipage}{\textwidth}
\footnotesize
Visual inputs are abbreviated as ROI (region of interest) and WSI (whole-slide image); a slash separates the inputs of different tasks or baselines, not inputs given together, and clinical text is not counted as an image. Sizes count question--answer pairs (QA) or, where a benchmark mixes formats, task instances (items), with k for thousands. They cover the reported splits, not necessarily test data only, without answer-position rotations, and for some benchmarks a subset: PathQABench counts its 105 MCQs, which come with optional clinical context; PathBench only PatchVQA, without its WSI tasks; SlideBench only its VQA subsets, without captioning; and EviPathBench its two visual MCQ tasks, without text-only QA and WSI localisation. DDxBench counts 150 cases; its agent reads the WSI, whereas the MLLM baseline receives 10 selected ROIs, with one fixed diagnostic prompt and labels taken from the reports. PathMMU-HR\textsuperscript{2} uses 16k$\times$16k ROIs, and VIPER rat tissue.\par
\smallskip
For QA construction, rule-based means templates or rules applied to captions, labels or annotations, mixed means several construction methods or reused items combined with new tasks, and NR that the method could not be established from the sources. \bench is mixed: \bench-Teaching has 1,296 diagnostic items (A1--C) built by rules from expert labels and an ontology and 390 LLM-generated finding-verification items (R1), and \bench-Atlas \nItemsAtlas\ items derived from the atlas captions by LLMs; the R1 statements and the atlas labels were verified by a pathologist. Of the properties, Expert means that pathologists reviewed the evaluation items, Multi-field that the input shows separate fields of one lesion, Case dx that a patient's diagnosis is the target (both partial for \bench-Atlas, whose 387 figure series show one disease, possibly from several patients), Hard neg.\ that distractors are deliberately close, and Robust that the evaluation is circular or tests consistency under controlled input changes. \cmark\ marks a property as present, \pmark\ as partial (a subset, some tasks, a baseline setting, text-only shortcut filtering, or accuracy averaged over option rotations), \xmark\ as not reported and -- as not applicable or not reported.
\end{minipage}
\end{table*}

\section{Relation to cross-scale benchmarks}
\label{supp:crossscale}
Scale-VQA~\cite{phan2026scalevqa} and its extension PathScale-VQA~\cite{phan2026pathscale} are the benchmarks closest to \bench: both show a model several images of one lesion at different magnifications.
\Cref{tab:crossscale} contrasts them.
The two designs answer different questions.
Cross-scale benchmarks ask whether a model can relate views when a question is constructed to need them (\eg locate a high-power field within the low-power view), and dropping views there lowers accuracy (from 64.8\% with all views to 27.8\% text-only, averaged over five models~\cite{phan2026pathscale}).
\bench asks the question a pathologist faces---what is the diagnosis of this case?---over the fields an expert chose to show it, and tests with field-set ablations and dedicated probes whether models use those fields; on this question the open models largely do not (Sec.~5.3\ofmain).

\begin{table*}[t]
\centering
\footnotesize
\setlength{\tabcolsep}{4pt}
\caption{\textbf{\bench vs.\ cross-scale pathology benchmarks.} Facts as reported in~\cite{phan2026scalevqa,phan2026pathscale}.}
\label{tab:crossscale}
\begin{tabular}{@{}p{2.1cm}p{4.5cm}p{4.5cm}p{5.2cm}@{}}
\toprule
 & Scale-VQA~\cite{phan2026scalevqa} & PathScale-VQA~\cite{phan2026pathscale} & \bench (ours) \\
\midrule
Status & MICCAI 2026 & arXiv 2026, extends Scale-VQA & this work \\
Image source & 177 TCGA WSIs, 15 organs & 373 TCGA WSIs, 14 organ sites & residency teaching library (436 cases) and 88 published atlases (800 cases), 18 organ systems \\
One item & zoom path: a region selected at 10$\times$ and corresponding regions at 40$\times$ and 200$\times$ & 1,368 expert-verified zoom paths (10$\times$, 40$\times$, 200$\times$) & a case: 2--8 fields chosen to show the diagnosis, by an expert (teaching, median 5) or in a published atlas (median 3), not necessarily nested \\
Diagnostic scope & TCGA cancer projects & TCGA cancer projects & \nEntitiesBoth\ entities (390 teaching, 800 atlas); teaching: 165 malignant, 61 intermediate, 96 benign, 114 non-neoplastic cases \\
Questions & GPT-5.2-generated MCQs; text-only adversarial screening; pathologist validation & generated from decomposed cross-scale feature sets; text-only screening; structure-controlled distractors; pathologist verification & teaching: rule-based questions and distractors from expert labels harmonised to a WHO-based ontology, LLM pre-review followed by pathologist adjudication, R1 statements LLM-generated; atlas: labels, statements and finding locations from the captions by LLMs; R1 statements and atlas labels verified by a pathologist \\
Tasks & correspondence, confirmation, localisation, explanation, diagnosis & the same five, plus five visual-grounding tasks & open diagnosis, ten-way diagnosis, behaviour triage, ordinal grading, finding verification, finding location \\
Size & 4,685 MCQs (3,230 / 450 / 1,005 train / val / test) & 10,373 MCQs (semantic: 3,680 train+val, 2,350 test; grounding: 4,343 test) & \nItemsBoth\ items (\nItems\ teaching, \nItemsAtlas\ atlas; \nPrompts\ and \nPromptsAtlas\ prompts with rotations); test only \\
Protocol & accuracy; answer positions balanced & accuracy; view-dropping ablation & CircularEval and single-pass accuracy, case-level bootstrap intervals, paired tests; field-set ablations; colour-matched and position-balanced probes; judge- and ontology-graded open answers \\
Purpose & training (GRPO) and evaluation & training and evaluation & evaluation only \\
Contamination & public TCGA slides & public TCGA slides & teaching images private, atlas figures published; no near-duplicate of either in 228,899 public images \\
\bottomrule
\end{tabular}
\end{table*}

\section{Construction pipeline}
\label{supp:pipeline}
\begin{figure*}[t]
\centering
\includegraphics[width=\textwidth]{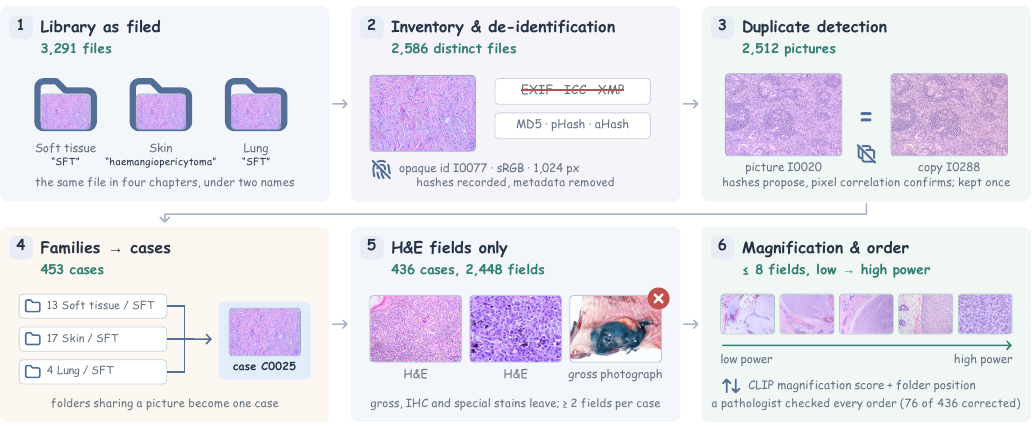}
\caption{\textbf{From the library as filed to benchmark cases.} The steps of this section, each with the count after it, shown on real items: a solitary fibrous tumour picture filed under four chapters and two names (C0025; three shown; SFT: solitary fibrous tumour), the re-encoded copy I0288 of picture I0020, the gross photograph I0019 leaving case C0065, and the five fields of the phyllodes case of Fig.~1\ofmain ordered from low to high power. The input is the representative set of Fig.~1b\ofmain, as filed in the teaching library.}
\label{fig:pipeline}
\end{figure*}
\Cref{fig:pipeline} summarises the steps. Fig.~1b\ofmain shows how the library was built: common diseases were listed from textbooks, their slides retrieved and scanned, four pathologists independently photographed fields of each lesion at several magnifications, and a consensus discussion kept the representative set. What we received is the output of that process as filed for teaching: the representative fields of each case, stored by organ system, with the same case often re-filed under other chapters and under the names of several WHO editions. The pipeline below starts from these files and recovers the cases; each step undoes one consequence of the filing, and the count after each step lets the reconstruction be audited. The steps are implemented as deterministic scripts, with a fixed seed for random operations.
\paragraph{File inventory.} The library files the same picture under several chapters, often under different names, so we first record exactly what is there: the hashes find copies, and the metadata fields are kept until de-identification can be checked against them. For each of the 3,291 files we record folder hierarchy, MD5, decode status, size, colour mode, embedded ICC profile, EXIF fields and three perceptual hashes (aHash $16{\times}16$, pHash and dHash $8{\times}8$).
One JPEG is truncated; two intact copies exist under other chapters.
\paragraph{De-identification.} Camera, date and author information sit in EXIF, ICC and XMP fields, and the folder names carry diagnosis and site; both must go before release, and colour profiles are unified so that every model sees the same sRGB image. One output per distinct MD5 with an opaque, randomly permuted identifier. Metadata are removed from sRGB JPEGs without recompression; all other files are decoded, converted to sRGB (23 Adobe RGB or generic RGB profiles), flattened onto white if they had an alpha channel (679 PNGs), and re-encoded (JPEG $q{=}95$, 4:4:4, or PNG). Model inputs are resized to a long side of 1,024\,px. A full scan of all 5,170 output files finds no EXIF, ICC or XMP payload.
\paragraph{Duplicate detection.} Copies are re-encoded, resized or colour-edited, so MD5 misses them, while embeddings alone over-merge: two different fields of one lesion routinely exceed a cosine of 0.985. Hashes therefore propose candidates and pixel correlation confirms them. Candidate pairs: pHash $\le 14$, aHash $\le 20$ or QuiltNet cosine $\ge 0.965$ (1,613 pairs). Accepted as the same picture if (pHash $\le 8$ and (cosine $\ge 0.95$ or 64$\times$64 pixel NCC $\ge 0.5$)) or aHash $\le 12$ or (cosine $\ge 0.99$ and NCC $\ge 0.4$). We inspected every accepted pair accepted by a single criterion and every rejected pair with NCC $> 0.6$: embedding-only matches at cosine $\ge 0.985$ were, in 20 of 22 cases, different fields of the same lesion (NCC $\approx 0$); one pHash-only match was a false positive (a low- and a high-power field); colour-edited copies of the same picture (pHash 0) are common. 76 pairs were accepted, giving 2,512 pictures.
\paragraph{Families and cases.} A case filed under several chapters would otherwise count several times and leak its label across items, so folders sharing a picture are merged; they are split only on explicit markers so that one case is never cut into two by guesswork. Leaf folders sharing a picture form a family (union--find). A family is split only on explicit filename markers present in every part with at least two pictures (``case~1/case~2'' and trailing numerals ``one/two/three'', written in Chinese in the filenames; \emph{papillary}/\emph{adamantinomatous}; \emph{high}/\emph{low grade}); 14 families were split into 29 cases. When the leaves of a family map to different entities (one family: extrahepatic cholangiocarcinoma vs.\ ampullary carcinoma) the majority entity is kept and the others become accepted alternates.
\paragraph{Magnification estimation.} Every prompt states that the fields run from low to high power, but the photograph order is lost once copies are merged, so magnification has to be estimated from the image; nuclear scale is one cue. Haematoxylin channel by colour deconvolution, normalised to the 1--99.5 percentile range; scale-normalised Laplacian-of-Gaussian responses at 22 geometric scales $\sigma \in [0.8, 24]$\,px; $\sigma^\ast$ is the scale with the largest 99th-percentile response, refined by a parabola in $\log\sigma$. $\sigma^\ast$ at the upper bound indicates large dark blobs (tissue fragments at very low power) and is treated as unreliable.
\paragraph{Field ordering.} No single cue is reliable (nuclear scale fails on overviews, and file numbers order photographs only within one folder), so the cues are combined and a pathologist checks every case; eight fields is the common input budget for all models. Zero-shot magnification score: the mean over four pathology CLIP models of the image's similarity to ``high magnification view of cell nuclei'' minus ``low magnification overview of tissue architecture''. Fields are ranked by this score adjusted by the photograph's position in its folder (file numbers order photographs only within one folder), with $\sigma^\ast$ as a tie-breaker; $\sigma^\ast$ alone fails on overviews, whose nuclei fall below the pixel scale. Cases with more than eight fields keep eight spanning the range. A pathologist checked every order (76 of 436 corrected).

\raggedbottom
\section{Detailed dataset composition}
\label{supp:dataset-detail}
\Cref{fig:dataset-detail} details the composition of \bench-Teaching and \cref{fig:dataset-detail-atlas} that of \bench-Atlas; both complement Fig.~2\ofmain, which covers the two sources together.
Panel (a) shows that coverage follows the residency curriculum, with the female genital tract, gastrointestinal tract and breast contributing the most entities; (b) that most intermediate and malignant cases are only suggestive from H\&E (P), which is why results are also reported on the H\&E-sufficient subset; (c) that most cases present three to seven fields and that the eight-field cap touches \detailCapCases\ cases (\detailInputFields\ of \detailRetainedFields\ fields are shown to the models); and (d) that \detailSingletonPct\% of entities have a single case, so coverage is not a large sample of each disease and results are reported by organ system, behaviour and task rather than per entity.
\bench-Atlas has one case per entity, so its coverage is a list of diseases rather than a sample of each: \detailAtlTone\ of its entities are common diagnoses (T1), soft tissue and skin contribute the most, and \detailAtlShared\ entities also have a Teaching case (\cref{fig:dataset-detail-atlas}; all entities in \cref{tab:atlas-entity-catalogue}).
Its cases are shorter: \detailAtlFields\ fields in all, two to four for most cases, and none above the eight-field cap.

\section{Ontology}
\label{supp:ontology}
\suppbalanceon
The ontology file lists, for every entity, Chinese and English names, synonyms, behaviour, lineage and family; 46 behaviour codes are marked low-confidence (\eg GIST, thymoma, paraganglioma, Kaposi sarcoma), and those entities are excluded from task B.
Site hints come from per-chapter keyword rules applied to the folder name first and its ancestors second; bonus folders that illustrate another organ (\eg endometrial tuberculosis under the male genital chapter) carry an explicit site.
\Cref{tab:entity-catalogue} lists the canonical English diagnoses of both sources by organ system: the \detailCases\ \bench-Teaching cases under the primary-system assignments of \cref{fig:dataset-detail}(a), with case counts within each system (entities shared across systems appear in each), and the one \bench-Atlas case of each of its \detailAtlCases\ entities.
The \detailEntities\ Teaching entities are a subset of the 407-entity ontology; for \bench-Atlas the ontology was extended by \detailAtlAdded\ entities (\detailAtlWho\ WHO entities and \detailAtlNew\ other diagnoses; \cref{tab:ontology-counts}).

\par\noindent\begin{minipage}{\linewidth}
\centering\small
\captionof{table}{Entities by behaviour and H\&E-sufficiency prior: the 407 entities of the ontology of \bench-Teaching and the \detailAtlAdded\ added for \bench-Atlas. The H\&E prior of the added entities was assigned by two LLMs and is P where they disagreed (\detailAtlHeFallback\ entities).}
\label{tab:ontology-counts}
\begin{tabular}{@{}lrrrr@{}}
\toprule
Behaviour & Y & P & N & Total \\
\midrule
\multicolumn{5}{@{}l}{\textit{Ontology of \bench-Teaching (407)}} \\
Non-neoplastic & 66 & 44 & 1 & 111 \\
Benign & 52 & 41 & 1 & 94 \\
Intermediate & 18 & 37 & 0 & 55 \\
Malignant & 42 & 85 & 20 & 147 \\
\textit{Total} & 178 & 207 & 22 & 407 \\
\midrule
\multicolumn{5}{@{}l}{\textit{Added for \bench-Atlas (439)}} \\
Non-neoplastic & 10 & 105 & 5 & 120 \\
Benign & 7 & 100 & 0 & 107 \\
Intermediate & 0 & 58 & 0 & 58 \\
Malignant & 1 & 153 & 0 & 154 \\
\textit{Total} & 18 & 416 & 5 & 439 \\
\midrule
\textbf{All entities} & 196 & 623 & 27 & \textbf{846} \\
\bottomrule
\end{tabular}

\end{minipage}\par\medskip
\onecolumn
\suppbalanceoff\flushbottom
\begin{center}
\includegraphics[width=0.9\textwidth]{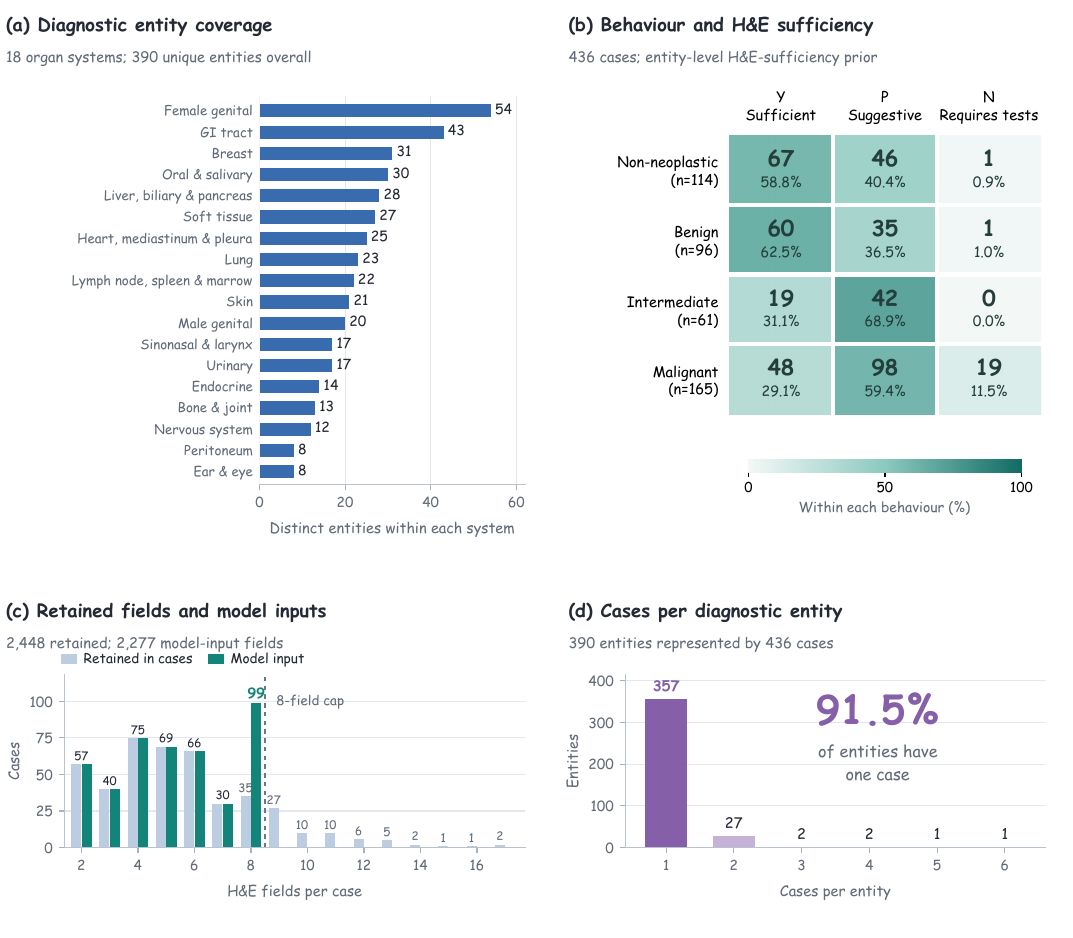}
\captionof{figure}{\textbf{Detailed composition of \bench-Teaching}.
(a) Distinct diagnostic entities within each primary organ system. An entity represented in multiple systems contributes to multiple bars (\detailMemberships\ system--entity memberships; \detailEntities\ unique entities overall).
(b) Case counts and within-behaviour percentages by H\&E-sufficiency prior: Y, expected to be diagnosable from H\&E; P, suggestive but usually confirmed with ancillary tests; N, ancillary tests required for the exact entity. Colour encodes the within-behaviour percentage; totals are \detailHeY\ (Y), \detailHeP\ (P) and \detailHeN\ (N) cases.
(c) Number of retained versus model-input fields per case (numbers: cases), with model inputs taken directly from the A1 items; both distributions contain the same \detailCases\ cases, of which \detailCapCases\ have more than eight retained fields before input selection.
(d) Number of entities represented by each case count, on a linear scale: \detailSingletons\ entities have one case each, and \detailMultiCaseEntities\ have two or more.}
\label{fig:dataset-detail}
\end{center}
\begingroup
\footnotesize
\setlength{\tabcolsep}{4pt}
\setlength{\LTcapwidth}{\textwidth}
\setlength{\LTpre}{0pt}
\renewcommand{\arraystretch}{1.02}
\begin{longtable}{@{}>{\raggedright\arraybackslash}p{\dimexpr.385\textwidth-\tabcolsep\relax}>{\raggedleft\arraybackslash}p{\dimexpr.04\textwidth-\tabcolsep\relax}>{\raggedleft\arraybackslash}p{\dimexpr.04\textwidth-\tabcolsep\relax}@{\hspace{.07\textwidth}}>{\raggedright\arraybackslash}p{\dimexpr.385\textwidth-\tabcolsep\relax}>{\raggedleft\arraybackslash}p{\dimexpr.04\textwidth-\tabcolsep\relax}>{\raggedleft\arraybackslash}p{\dimexpr.04\textwidth-\tabcolsep\relax}@{}}
\caption{\textbf{Diagnostic entities of \bench.} T: \bench-Teaching cases of the entity filed under the organ system (an entity filed under several systems is listed in each); A: its \bench-Atlas case (totals in the last row). Grouped by organ system and alphabetised by the canonical English name, reading down the left list and then the right list within each group.}\label{tab:entity-catalogue}\label{tab:atlas-entity-catalogue}\\
\toprule
Diagnostic entity & T & A & Diagnostic entity & T & A \\
\midrule
\endfirsthead
\multicolumn{6}{@{}l}{\textbf{Table \thetable.} Diagnostic entities of \bench\ (continued).}\\[3pt]
\toprule
Diagnostic entity & T & A & Diagnostic entity & T & A \\
\midrule
\endhead
\midrule
\multicolumn{6}{r@{}}{\emph{Continued on the next page.}}\\
\endfoot
\bottomrule
\endlastfoot
\multicolumn{6}{@{}l}{\textbf{Oral \& salivary} (Teaching 30 cases; Atlas 31)}\\*[2pt]
Acinic cell carcinoma & 1 & 1 & Granulomatous cheilitis & -- & 1 \\*
Adenoid cystic carcinoma & 1 & 1 & Kimura disease & 1 & -- \\
Adenomatoid odontogenic tumour & -- & 1 & Lichen planus & 1 & -- \\
Ameloblastic carcinoma & 1 & 1 & Lymphangioma (lymphatic malformation) & 1 & -- \\
Ameloblastic fibrosarcoma & -- & 1 & Lymphoepithelial sialadenitis (Sjögren syndrome) & 1 & 1 \\
Ameloblastoma & 1 & -- & Mucocele / ranula & 1 & 1 \\
Ameloblastoma, extraosseous/peripheral & -- & 1 & Mucoepidermoid carcinoma & 1 & 1 \\
Ameloblastoma, unicystic & -- & 1 & Myoepithelial carcinoma & 1 & 1 \\
Amyloidosis & 1 & -- & Myoepithelioma & 1 & -- \\
Apical granuloma & -- & 1 & Odontogenic fibroma, NOS & -- & 1 \\
Arteriovenous malformation & -- & 1 & Odontogenic keratocyst & -- & 1 \\
Basal cell adenocarcinoma & -- & 1 & Odontogenic myxoma & -- & 1 \\
Basal cell adenoma & 1 & 1 & Pleomorphic adenoma & 1 & 1 \\
Branchial cleft cyst & 1 & -- & Salivary duct carcinoma & 1 & 1 \\
Carcinoma ex pleomorphic adenoma & 1 & 1 & Secretory carcinoma (mammary analogue) & 1 & -- \\
Cemento-ossifying fibroma & -- & 1 & Sialoblastoma & -- & 1 \\
Cementoblastoma & -- & 1 & Squamous cell carcinoma & 1 & -- \\
Chalazion & -- & 1 & Squamous cell carcinoma in situ & 1 & -- \\
Dentigerous cyst & -- & 1 & Squamous papilloma & 1 & -- \\
Epithelial-myoepithelial carcinoma & 1 & 1 & Thyroglossal duct cyst & 1 & -- \\
Epulis (reactive gingival lesion) & 1 & 1 & Vascular malformation & 1 & -- \\
Fibrous dysplasia & 1 & -- & Verrucous carcinoma & 1 & -- \\
Ghost cell odontogenic carcinoma & -- & 1 & Warthin tumor & 1 & 1 \\
Granular cell tumor & 1 & -- &  &  &  \\[4pt]
\multicolumn{6}{@{}l}{\textbf{GI tract} (Teaching 44 cases; Atlas 52)}\\*[2pt]
Acute gangrenous appendicitis & 1 & -- & Gastric intestinal metaplasia & 1 & 1 \\*
Acute phlegmonous (suppurative) appendicitis & 1 & 1 & Gastric mucosal erosion & 1 & -- \\
Acute simple appendicitis & 1 & -- & Gastric ulcer & -- & 1 \\
Adenocarcinoma, intestinal type & -- & 1 & Gastritis cystica profunda & 1 & 1 \\
Adenosquamous carcinoma & -- & 1 & Gastroblastoma & -- & 1 \\
Bacillary dysentery & -- & 1 & Gastrointestinal stromal tumour & 1 & 1 \\
Barrett's esophagus & -- & 1 & Goblet cell adenocarcinoma (appendix) & 1 & -- \\
Basaloid carcinoma & -- & 1 & Hirschsprung disease & 1 & 1 \\
Basaloid squamous cell carcinoma & 1 & -- & Inflammatory fibroid polyp & 1 & 1 \\
Carcinosarcoma & 1 & -- & Melanoma & 1 & -- \\
Chronic atrophic gastritis & 1 & 1 & Monomorphic epitheliotropic intestinal T-cell lymphoma & -- & 1 \\
Chronic non-atrophic gastritis, mild & 1 & 1 & Mucinous carcinoma, gastric type & -- & 1 \\
Chronic non-atrophic gastritis, moderate, active & 1 & -- & Mucinous cystadenoma & -- & 1 \\
Colloid adenocarcinoma & -- & 1 & Neuroendocrine tumour & 1 & -- \\
Colorectal adenocarcinoma & 1 & 1 & Oesophageal adenocarcinoma & -- & 1 \\
Colorectal hyperplastic polyp & 1 & -- & Oesophageal squamous dysplasia, high grade & 1 & 1 \\
Columnar cell change / hyperplasia & -- & 1 & Oesophageal squamous dysplasia, low grade & 1 & -- \\
Crawling-type (hand-in-hand) gastric adenocarcinoma & 1 & -- & Paget disease, extramammary & -- & 1 \\
Crohn disease & 1 & 1 & Peutz-Jeghers polyp & 1 & 1 \\
Cryptococcosis & -- & 1 & Plexiform fibromyxoma & -- & 1 \\
Cysticercosis & -- & 1 & Pseudomyxoma peritonei & 1 & -- \\
Diffuse large B-cell lymphoma & 1 & -- & Reflux esophagitis & -- & 1 \\
Duodenal adenocarcinoma & -- & 1 & Sarcomatoid squamous cell carcinoma & -- & 1 \\
Duodenal adenoma & 1 & 1 & Schwannoma & 1 & -- \\
Enteropathy-associated T-cell lymphoma & -- & 1 & Signet-ring cell carcinoma (poorly cohesive carcinoma) & 1 & 1 \\
Eosinophilic appendicitis & 1 & -- & Squamous cell carcinoma & 1 & 1 \\
Eosinophilic Gastritis & -- & 1 & Squamous cell carcinoma in situ & -- & 1 \\
Extranodal marginal zone lymphoma of mucosa-associated lymphoid tissue & 2 & -- & Tuberculosis & 1 & -- \\
Fetal adenocarcinoma & -- & 1 & Tubular adenoma & -- & 1 \\
Fundic gland polyp & 1 & 1 & Tubular adenoma with high-grade dysplasia & 1 & -- \\
Ganglioneuromatosis & -- & 1 & Tubular adenoma with low-grade dysplasia & 1 & 1 \\
Gastric adenocarcinoma & 1 & 1 & Tubulovillous adenoma & -- & 1 \\
Gastric carcinoma with lymphoid stroma (lymphoepithelioma-like) & 1 & 1 & Ulcerative colitis & 1 & 1 \\
Gastric dysplasia, high grade & 1 & 1 & Vascular malformation & -- & 1 \\
Gastric dysplasia, low grade & 1 & 1 & Villous adenoma & -- & 1 \\
Gastric hyperplastic polyp & 1 & 1 & Xanthogranulomatous appendicitis & 1 & -- \\[4pt]
\multicolumn{6}{@{}l}{\textbf{Sinonasal \& larynx} (Teaching 18 cases; Atlas 21)}\\*[2pt]
Biphenotypic sinonasal sarcoma & -- & 1 & NUT carcinoma & -- & 1 \\*
Chronic dacryocystitis & -- & 1 & Olfactory neuroblastoma & 1 & 1 \\
Chronic rhinosinusitis with nasal polyps & 1 & -- & Papillary squamous cell carcinoma & -- & 1 \\
Chronic tonsillitis & 1 & 1 & Pituitary adenoma / PitNET & 1 & -- \\
Craniopharyngioma & 2 & -- & Plasmacytoma, extramedullary & -- & 1 \\
Ewing sarcoma & 1 & -- & Rhinoscleroma & -- & 1 \\
Fungal rhinosinusitis & 1 & 1 & Sinonasal papilloma, exophytic type & -- & 1 \\
Laryngeal keratosis & -- & 1 & Sinonasal undifferentiated carcinoma & 1 & -- \\
Low-grade nasopharyngeal papillary adenocarcinoma & -- & 1 & Squamous cell carcinoma, HPV-associated & -- & 1 \\
Lymphoepithelial carcinoma & 1 & -- & Squamous cell carcinoma, non-keratinizing, NOS & -- & 1 \\
Malignant triton tumour & -- & 1 & Squamous cell papilloma, inverted & -- & 1 \\
Melanoma & 1 & -- & Squamous dysplasia (high-grade) & 1 & 1 \\
Meningioma & 1 & -- & Tuberculosis & 1 & -- \\
Nasopharyngeal angiofibroma & 1 & 1 & Verrucous carcinoma & -- & 1 \\
Nasopharyngeal carcinoma (non-keratinizing) & 1 & 1 & Vocal cord polyp / nodule & 1 & 1 \\
Neuroendocrine tumour & 1 & -- &  &  &  \\[4pt]
\multicolumn{6}{@{}l}{\textbf{Lung} (Teaching 25 cases; Atlas 37)}\\*[2pt]
Acute Lung Injury & -- & 1 & Lung adenocarcinoma & 1 & 1 \\*
Adenocarcinoma in situ, mucinous & -- & 1 & Lymphangioleiomyomatosis & 1 & 1 \\
Adenocarcinoma in situ, non-mucinous & -- & 1 & Lymphomatoid granulomatosis & -- & 1 \\
Alveolar adenoma & -- & 1 & Microcystic stromal tumour & -- & 1 \\
Atypical adenomatous hyperplasia & 1 & 1 & Minimally invasive adenocarcinoma, non-mucinous & -- & 1 \\
Bronchiectasis & -- & 1 & Mixed squamous cell and glandular papilloma & -- & 1 \\
Bronchioloalveolar Carcinoma & -- & 1 & Organizing pneumonia & 1 & 1 \\
Bronchogenic cyst & -- & 1 & Paragonimiasis & -- & 1 \\
Chronic Hypersensitivity Pneumonitis & -- & 1 & PEComa (clear cell "sugar" tumour) & 1 & -- \\
Congenital cystic adenomatoid malformation & -- & 1 & Pleuropulmonary blastoma & -- & 1 \\
Cryptococcosis & 2 & -- & Pneumocystis pneumonia & 1 & 1 \\
Cytomegalovirus pneumonia & -- & 1 & Pulmonary blastoma & -- & 1 \\
Desquamative Interstitial Pneumonia & -- & 1 & Pulmonary hamartoma & 1 & 1 \\
Diffuse large B-cell lymphoma & 1 & -- & Pulmonary myxoid sarcoma with EWSR1-CREB1 fusion & -- & 1 \\
Diffuse Panbronchiolitis & -- & 1 & Sarcoidosis & 1 & -- \\
Extranodal marginal zone lymphoma of mucosa-associated lymphoid tissue & 1 & -- & Sclerosing pneumocytoma & 1 & 1 \\
Glandular papilloma & -- & 1 & Silicosis & 1 & 1 \\
Histoplasmosis & 1 & 1 & Small cell carcinoma & 1 & 1 \\
Inflammatory myofibroblastic tumour & 1 & -- & Solitary fibrous tumour & 1 & -- \\
Inflammatory pseudotumour & 1 & -- & Squamous cell carcinoma & 1 & -- \\
Invasive mucinous adenocarcinoma (lung) & 1 & 1 & Tuberculosis & 2 & -- \\
Large cell carcinoma & 1 & 1 & Typical carcinoid / neuroendocrine tumour, grade 1 & -- & 1 \\
Large cell neuroendocrine carcinoma & 1 & 1 & Usual Interstitial Pneumonia & -- & 1 \\
Lobar pneumonia & -- & 1 &  &  &  \\[4pt]
\multicolumn{6}{@{}l}{\textbf{Heart, mediastinum \& pleura} (Teaching 25 cases; Atlas 23)}\\*[2pt]
Atherosclerosis & -- & 1 & Neuroendocrine tumour & 1 & -- \\*
Branchial cleft cyst & -- & 1 & Papillary fibroelastoma & -- & 1 \\
Cardiac myxoma & 1 & 1 & Primary mediastinal large B-cell lymphoma & 1 & 1 \\
Castleman disease & 1 & -- & Reactive mesothelial hyperplasia & -- & 1 \\
Cystic tumour of atrioventricular node & -- & 1 & Rheumatic myocarditis & -- & 1 \\
Embryonal carcinoma & 1 & -- & Seminoma / dysgerminoma / germinoma & 1 & -- \\
Extranodal marginal zone lymphoma of mucosa-associated lymphoid tissue & 1 & -- & T-lymphoblastic leukaemia/lymphoma & 1 & -- \\
Haemangioma & 1 & -- & Thymic basaloid carcinoma & 1 & 1 \\
Hodgkin lymphoma & 1 & -- & Thymic cyst & 1 & -- \\
Immature teratoma & 1 & -- & Thymic lymphoepithelial carcinoma & 1 & 1 \\
Intimal sarcoma & -- & 1 & Thymic mucoepidermoid carcinoma & 1 & 1 \\
Keshan disease & -- & 1 & Thymic squamous cell carcinoma & 1 & 1 \\
Lymphangioma (lymphatic malformation) & 1 & -- & Thymoma, type A & 1 & 1 \\
Mature (cystic) teratoma & 1 & -- & Thymoma, type AB & 1 & 1 \\
Mesothelioma (malignant) & 1 & 1 & Thymoma, type B1 & 1 & 1 \\
Metaplastic thymoma & -- & 1 & Thymoma, type B2 & 1 & 1 \\
Micronodular thymoma with lymphoid stroma & -- & 1 & Thymoma, type B3 & 1 & 1 \\
Myocardial infarction & -- & 1 & Yolk sac tumour & 1 & -- \\[4pt]
\multicolumn{6}{@{}l}{\textbf{Liver, biliary \& pancreas} (Teaching 31 cases; Atlas 45)}\\*[2pt]
Acute cholecystitis & 1 & -- & Gallbladder adenoma & 2 & 1 \\*
Acute Liver Rejection & -- & 1 & Glucagonoma & -- & 1 \\
Acute pancreatitis & -- & 1 & Hepatic schistosomiasis (egg deposition) & 1 & 1 \\
Acute viral hepatitis & 1 & 1 & Hepatic steatosis & 1 & 1 \\
Adenosarcoma & -- & 1 & Hepatoblastoma & 1 & 1 \\
Alcoholic (steato)hepatitis & 1 & 1 & Hepatocellular adenoma & 1 & 1 \\
Amoebic liver abscess & 1 & 1 & Hepatocellular carcinoma & 2 & 1 \\
Ampullary carcinoma & -- & 1 & Insulinoma & -- & 1 \\
Bile duct adenoma & -- & 1 & Intraductal papillary mucinous neoplasm & 1 & 1 \\
Cholesterol polyp (cholesterolosis) & -- & 1 & Intraductal papillary mucinous neoplasm with associated invasive carcinoma & -- & 1 \\
Chronic cholecystitis & 1 & 1 & Intrahepatic cholangiocarcinoma & 1 & 1 \\
Chronic hepatic congestion & -- & 1 & Mucinous cystadenocarcinoma & -- & 1 \\
Chronic hepatitis & 1 & 1 & Mucinous cystic neoplasm (pancreas) & 1 & -- \\
Chronic pancreatitis & 1 & 1 & Mucinous cystic neoplasm with high-grade intraepithelial neoplasia & -- & 1 \\
Cirrhosis & 1 & 1 & Neuroendocrine tumour & 2 & -- \\
Clonorchiasis & -- & 1 & Non-functioning pancreatic neuroendocrine tumour & -- & 1 \\
Combined hepatocellular-cholangiocarcinoma & 1 & 1 & Pancreatic ductal adenocarcinoma & 1 & 1 \\
Congenital hepatic fibrosis & -- & 1 & Pancreatic pseudocyst & 1 & 1 \\
Embryonal sarcoma & -- & 1 & Pancreatoblastoma & -- & 1 \\
Extrahepatic cholangiocarcinoma & 1 & 1 & Pyogenic liver abscess & 1 & 1 \\
Fibrolamellar carcinoma & 1 & 1 & Serous cystadenocarcinoma & -- & 1 \\
Focal nodular hyperplasia & 1 & 1 & Serous cystadenoma & -- & 1 \\
Fulminant hepatitis (massive hepatic necrosis) & -- & 1 & Serous cystadenoma (pancreas) & 1 & 1 \\
Gallbladder adenocarcinoma & 1 & 1 & Solid pseudopapillary neoplasm & 1 & 1 \\[4pt]
\multicolumn{6}{@{}l}{\textbf{Peritoneum} (Teaching 8 cases; Atlas 7)}\\*[2pt]
Atypical lipomatous tumour / well-differentiated liposarcoma & 1 & -- & Myelolipoma & 1 & -- \\*
Desmoid-type fibromatosis & 1 & -- & Neuroblastoma & 1 & -- \\
Desmoplastic small round cell tumour & 1 & -- & Paraganglioma & 1 & -- \\
Ganglioneuroblastoma & 1 & -- & PEComa (clear cell "sugar" tumour) & -- & 1 \\
Ganglioneuroma & 1 & -- & Peritoneal inclusion cysts & -- & 1 \\
Idiopathic Retroperitoneal Fibrosis & -- & 1 & Pseudomyxoma peritonei & -- & 1 \\
Leiomyomatosis, peritonealis disseminata & -- & 1 & Well-differentiated papillary mesothelial tumour & -- & 1 \\
Multicystic mesothelioma (peritoneal inclusion cyst) & -- & 1 &  &  &  \\[4pt]
\multicolumn{6}{@{}l}{\textbf{Endocrine} (Teaching 18 cases; Atlas 28)}\\*[2pt]
Adrenal cortical adenoma & 2 & 1 & Papillary cystadenoma & -- & 1 \\*
Adrenal cyst & 1 & 1 & Papillary thyroid carcinoma & 1 & 1 \\
Adrenocortical carcinoma & -- & 1 & Paraganglioma & -- & 1 \\
Chordoma & 1 & -- & Parathyroid adenoma & 1 & 1 \\
Columnar cell variant of PTC & -- & 1 & Parathyroid carcinoma & -- & 1 \\
Encapsulated variant of PTC & -- & 1 & Parathyroid hyperplasia & 1 & -- \\
Follicular thyroid adenoma & 1 & 1 & Phaeochromocytoma & 2 & 1 \\
Follicular thyroid carcinoma & 1 & 1 & Pituitary adenoma / PitNET & -- & 1 \\
Follicular variant of PTC & -- & 1 & Poorly Differentiated Thyroid Carcinoma & -- & 1 \\
Hashimoto (chronic lymphocytic) thyroiditis & 1 & 1 & Pretibial myxedema & -- & 1 \\
Medullary thyroid carcinoma & 2 & 1 & Solid papillary carcinoma with invasion & -- & 1 \\
Myelolipoma & -- & 1 & Subacute granulomatous thyroiditis & 1 & 1 \\
Neuroendocrine carcinoma (poorly differentiated neuroendocrine neoplasm) & -- & 1 & Thyroglossal duct cyst & -- & 1 \\
Nodular goitre (adenomatoid nodule) & 2 & 1 & Toxic goitre (Graves disease) & -- & 1 \\
Oncocytic (Hürthle cell) adenoma of thyroid & 1 & 1 & True hermaphroditism & -- & 1 \\[4pt]
\multicolumn{6}{@{}l}{\textbf{Urinary} (Teaching 19 cases; Atlas 44)}\\*[2pt]
Acute Allergic Interstitial Nephritis & -- & 1 & Mesoblastic nephroma & -- & 1 \\*
Acute Pyelonephritis & -- & 1 & Metanephric adenofibroma & -- & 1 \\
ALK-rearranged renal cell carcinomas & -- & 1 & Metanephric adenoma & -- & 1 \\
Angiomyolipoma & 1 & -- & Metanephric stromal tumour & -- & 1 \\
Angiomyolipoma, epithelioid & -- & 1 & Mixed epithelial and stromal tumour & -- & 1 \\
Chromophobe renal cell carcinoma & 1 & 1 & Multilocular cystic renal neoplasm of low malignant potential & 1 & 1 \\
Chronic Pyelonephritis & -- & 1 & Nephroblastoma (Wilms tumour) & 1 & 1 \\
Chronic sclerosing glomerulonephritis & -- & 1 & Non-invasive papillary urothelial carcinoma, high grade & 1 & 1 \\
Clear cell papillary renal cell tumour & 1 & -- & Non-invasive papillary urothelial carcinoma, low grade & 2 & 1 \\
Clear cell renal cell carcinoma & 1 & 1 & Ossifying renal tumour of infancy & -- & 1 \\
Clear cell sarcoma of kidney & -- & 1 & Papillary renal cell carcinoma & 1 & 1 \\
Collecting duct carcinoma & 1 & 1 & Papillary urothelial neoplasm of low malignant potential & -- & 1 \\
Crescentic glomerulonephritis & -- & 1 & Renal oncocytoma & 2 & 1 \\
Cystic partially differentiated nephroblastoma & -- & 1 & Succinate dehydrogenase-deficient renal cell carcinoma & -- & 1 \\
Cystitis glandularis & 1 & 1 & Teratoma with somatic-type malignancy & -- & 1 \\
Diabetic nephropathy & -- & 1 & TFE3-rearranged renal cell carcinoma (Xp11.2 translocation) & 1 & 1 \\
Endocapillary Proliferative Glomerulonephritis & -- & 1 & TFEB-altered renal cell carcinoma (t(6;11)) & 1 & -- \\
Fibroepithelial polyp (acrochordon) & -- & 1 & TFEB-altered renal cell carcinomas & -- & 1 \\
Genital rhabdomyoma & -- & 1 & Urachal carcinoma & -- & 1 \\
IgA nephropathy & -- & 1 & Urothelial carcinoma in situ & 1 & 1 \\
Invasive urothelial carcinoma & 1 & 1 & Urothelial carcinoma involving prostate & -- & 1 \\
Inverted urothelial papilloma & 1 & 1 & Urothelial papilloma & -- & 1 \\
Juxtaglomerular tumour & -- & 1 & Xanthogranulomatous pyelonephritis & -- & 1 \\
Malignant rhabdoid tumour of the kidney & -- & 1 &  &  &  \\[4pt]
\multicolumn{6}{@{}l}{\textbf{Male genital} (Teaching 22 cases; Atlas 28)}\\*[2pt]
Acute prostatitis & 1 & 1 & Large cell calcifying Sertoli cell tumour & -- & 1 \\*
Adenocarcinoma of the epididymis & -- & 1 & Leydig cell tumour & -- & 1 \\
Adenoid basal carcinoma & -- & 1 & Low-grade squamous intraepithelial lesion (LSIL) & 1 & -- \\
Adenomatoid tumour & 1 & -- & Malignant Sertoli cell tumour & -- & 1 \\
Androgen Insensitivity Syndrome & -- & 1 & Mixed germ cell tumour & -- & 1 \\
Basal cell carcinoma of prostate & 1 & 1 & Multicystic mesothelioma (peritoneal inclusion cyst) & 1 & -- \\
Basal cell hyperplasia (prostate) & 1 & 1 & Prostatic acinar adenocarcinoma & 3 & 1 \\
Benign prostatic hyperplasia & 1 & 1 & Prostatic atrophy & 1 & 1 \\
Carcinoma with medullary pattern & -- & 1 & Prostatic ductal adenocarcinoma & 1 & 1 \\
Chronic prostatitis & 1 & 1 & Reactive mesothelial hyperplasia & 1 & -- \\
Condyloma acuminatum & 1 & 1 & Sarcomatoid acinar adenocarcinoma & -- & 1 \\
Erythroplasia of Queyrat (penile SCC in situ) & -- & 1 & Seminoma / dysgerminoma / germinoma & -- & 1 \\
Germ cell neoplasia in situ & -- & 1 & Sertoli cell tumour & -- & 1 \\
Gonadoblastoma & -- & 1 & Sertoli-Leydig cell tumour & -- & 1 \\
Granulomatous prostatitis & 1 & 1 & Small cell neuroendocrine carcinoma of prostate & 1 & 1 \\
High-grade prostatic intraepithelial neoplasia & 1 & 1 & Stromal tumour of uncertain malignant potential & -- & 1 \\
High-grade squamous intraepithelial lesion (CIN 2) & 1 & -- & Urothelial carcinoma involving prostate & 1 & -- \\
High-grade squamous intraepithelial lesion (CIN 3) & 1 & -- & Well-differentiated papillary mesothelial tumour & 1 & -- \\[4pt]
\multicolumn{6}{@{}l}{\textbf{Female genital} (Teaching 56 cases; Atlas 78)}\\*[2pt]
Actinomycosis & -- & 1 & High-grade squamous intraepithelial lesion (CIN 2) & -- & 1 \\*
Acute salpingitis & 1 & 1 & High-grade squamous intraepithelial lesion (CIN 3) & -- & 1 \\
Adenocarcinoma in situ of cervix & 1 & 1 & HSIL with endocervical gland involvement & 1 & 1 \\
Adenomatoid tumour & -- & 1 & Hydatidiform mole & 1 & -- \\
Adenomyoma & 1 & 1 & Immature teratoma & -- & 1 \\
Adenomyosis & -- & 1 & Invasive hydatidiform mole & -- & 1 \\
Adenosarcoma & 1 & -- & Late secretory endometrium & 1 & -- \\
Adult granulosa cell tumour & 1 & 1 & Leiomyosarcoma & 1 & -- \\
Angiomyofibroblastoma & 1 & 1 & Lipoleiomyoma & 1 & -- \\
Atrophic Endometrium & -- & 1 & Low-grade appendiceal mucinous neoplasm & -- & 1 \\
Atypical hyperplasia / endometrioid intraepithelial neoplasia & 2 & 1 & Low-grade endometrial stromal sarcoma & 1 & 1 \\
Atypical polypoid adenomyoma & -- & 1 & Low-grade serous carcinoma & -- & 1 \\
Bartholin gland cyst & -- & 1 & Low-grade squamous intraepithelial lesion (LSIL) & -- & 1 \\
Behcet's disease & -- & 1 & Mature (cystic) teratoma & 1 & 1 \\
Brenner tumour NOS & -- & 1 & Mesonephric remnants & 1 & -- \\
Brenner tumour, borderline malignancy & -- & 1 & Metastatic tumour to ovary (Krukenberg tumour) & 1 & 1 \\
Brenner tumour, malignant & -- & 1 & Microinvasive squamous cell carcinoma & -- & 1 \\
Carcinosarcoma & 1 & 1 & Mid-to-late secretory endometrium & 1 & -- \\
Choriocarcinoma & 1 & 1 & Mucinous borderline tumour & 1 & -- \\
Chronic (Autoimmune) Oophoritis & -- & 1 & Mucinous carcinoma (ovary) & 1 & 1 \\
Chronic Endometritis & -- & 1 & Mucinous cystadenoma & 1 & -- \\
Chronic salpingitis & 1 & -- & Mucosal melanoma & -- & 1 \\
Clear cell borderline tumour & -- & 1 & Myxoid leiomyoma & 1 & -- \\
Clear cell carcinoma (female genital tract) & 1 & 1 & Neuroendocrine tumour & 1 & -- \\
Complete hydatidiform mole & -- & 1 & Ovarian endometrioid tumour & 1 & -- \\
Corpus luteum cyst & 1 & 1 & Ovarian fibroma & 1 & 1 \\
Cortical inclusion cyst (ovary) & 2 & -- & Paget disease (mammary / extramammary) & 1 & -- \\
Cystic lymphangioma & -- & 1 & Partial hydatidiform mole & -- & 1 \\
Dedifferentiated carcinoma & -- & 1 & Phyllodes tumour, benign & -- & 1 \\
Deep (aggressive) angiomyxoma & 1 & -- & Phyllodes tumour, borderline & -- & 1 \\
Ductal carcinoma in situ, low nuclear grade & -- & 1 & Placental site trophoblastic tumour & -- & 1 \\
Dysgerminoma & -- & 1 & Postoperative spindle cell nodule & -- & 1 \\
Early secretory endometrium & 1 & -- & Pregnancy luteoma & -- & 1 \\
Ectopic pregnancy in the fallopian tube & -- & 1 & Proliferative endometrium (early-mid) & 1 & -- \\
Embryonal carcinoma & -- & 1 & Sclerosing adenosis & -- & 1 \\
Embryonal rhabdomyosarcoma & 1 & -- & Sclerosing stromal tumour & -- & 1 \\
Endocervical polyp & 1 & 1 & Secondary Syphilis & -- & 1 \\
Endometrial hyperplasia without atypia & 1 & 1 & Seromucinous borderline tumour & -- & 1 \\
Endometrial polyp & 1 & 1 & Serous adenofibroma NOS & -- & 1 \\
Endometrial stromal nodule & 1 & 1 & Serous borderline tumour & 1 & 1 \\
Endometrioid carcinoma (endometrium) & 1 & 1 & Serous cystadenoma & 1 & -- \\
Endometrioid cystadenoma NOS & -- & 1 & Sertoli-Leydig cell tumour & 1 & -- \\
Endometrioid tumour, borderline & -- & 1 & Small cell carcinoma, hypercalcaemic type & -- & 1 \\
Endometriosis & 1 & 1 & Squamous cell carcinoma & 1 & -- \\
Epithelioid leiomyoma & 1 & -- & Squamous metaplasia (cervix) & 1 & -- \\
Epithelioid trophoblastic tumour & -- & 1 & Steroid cell tumour NOS & -- & 1 \\
Fibroepithelial polyp (acrochordon) & 1 & -- & Struma ovarii & 1 & 1 \\
Fumarate hydratase-deficient leiomyoma & 1 & 1 & Thecoma & 1 & 1 \\
Galactocele & -- & 1 & Thecoma, luteinized & -- & 1 \\
Granulosa cell tumour, juvenile & -- & 1 & Tunnel clusters (cervix) & 1 & -- \\
High-grade endometrial stromal sarcoma & 1 & 1 & Vaginal adenosis & -- & 1 \\
High-grade serous carcinoma / endometrial serous carcinoma & 1 & 1 & Yolk sac tumour & -- & 1 \\[4pt]
\multicolumn{6}{@{}l}{\textbf{Breast} (Teaching 35 cases; Atlas 42)}\\*[2pt]
Adenoma of nipple & -- & 1 & Lipid-rich carcinoma & -- & 1 \\*
Adenomyoepithelioma NOS & -- & 1 & Lobular carcinoma in situ & 2 & 1 \\
Apocrine adenocarcinoma & -- & 1 & Lobular carcinoma in situ, pleomorphic & -- & 1 \\
Atypical ductal hyperplasia & 1 & 1 & Metaplastic carcinoma NOS & -- & 1 \\
Atypical lobular hyperplasia & 1 & 1 & Microglandular adenosis & 1 & 1 \\
Carcinoma with medullary pattern & 1 & -- & Microinvasive carcinoma (breast) & 1 & -- \\
Cellular fibroadenoma & 1 & 1 & Mucinous carcinoma (breast) & 1 & 1 \\
Columnar cell change / hyperplasia & 1 & -- & Mucinous tubular and spindle cell carcinoma & -- & 1 \\
Complex fibroadenoma & 1 & 1 & Myeloid sarcoma & -- & 1 \\
Cribriform carcinoma NOS & -- & 1 & Myoid hamartoma & -- & 1 \\
Duct adenoma NOS & -- & 1 & Oncocytic carcinoma & -- & 1 \\
Duct ectasia & 1 & 1 & Phyllodes tumour, benign & 1 & -- \\
Ductal carcinoma in situ, high nuclear grade & 1 & 1 & Phyllodes tumour, borderline & 1 & -- \\
Ductal carcinoma in situ, low nuclear grade & 1 & -- & Phyllodes tumour, malignant & 1 & -- \\
Fat necrosis & 1 & 1 & Pleomorphic carcinoma & -- & 1 \\
Fibroadenoma & 1 & -- & Polymorphous adenocarcinoma & -- & 1 \\
Fibrocystic change & 1 & 1 & Pseudoangiomatous stromal hyperplasia & 1 & 1 \\
Glycogen-rich carcinoma & -- & 1 & Radial scar / complex sclerosing lesion & 1 & 1 \\
Granulomatous lobular mastitis & -- & 1 & Sclerosing adenosis & 1 & -- \\
Intraductal (encapsulated) papillary carcinoma & 2 & 1 & Secretory carcinoma (mammary analogue) & -- & 1 \\
Intraductal papilloma & 3 & 1 & Solid papillary carcinoma & 1 & 1 \\
Invasive breast carcinoma of no special type & 1 & 1 & Solid papillary carcinoma in situ & -- & 1 \\
Invasive lobular carcinoma & 1 & 1 & Tall cell carcinoma with reversed polarity & -- & 1 \\
Invasive micropapillary carcinoma & 1 & 1 & Tubular carcinoma & 1 & 1 \\
Juvenile fibroadenoma & 1 & 1 & Usual ductal hyperplasia & 1 & 1 \\
Lactating adenoma & -- & 1 &  &  &  \\[4pt]
\multicolumn{6}{@{}l}{\textbf{Soft tissue} (Teaching 28 cases; Atlas 121)}\\*[2pt]
Acral fibromyxoma & -- & 1 & Intramuscular lipoma & -- & 1 \\*
Adenomyoepithelioma with carcinoma & -- & 1 & Intravenous leiomyomatosis & -- & 1 \\
Adult cellular rhabdomyoma & -- & 1 & Ischaemic fasciitis & -- & 1 \\
Adult fibrosarcoma & 1 & 1 & Juvenile hyaline fibromatosis & -- & 1 \\
Alveolar rhabdomyosarcoma & 1 & 1 & Kaposi sarcoma & 1 & -- \\
Alveolar soft part sarcoma & -- & 1 & Leiomyoma & -- & 1 \\
Angiofibroma NOS & -- & 1 & Leiomyoma with bizarre nuclei & -- & 1 \\
Angiolipoma & 1 & 1 & Leiomyosarcoma & -- & 1 \\
Angiomatoid fibrous histiocytoma & -- & 1 & Lipoblastoma & 1 & 1 \\
Angiomatosis & -- & 1 & Lipofibromatosis & -- & 1 \\
Angiomyolipoma & -- & 1 & Lipoma & 1 & 1 \\
Angiosarcoma & 1 & 1 & Lipoma-like liposarcoma & -- & 1 \\
Atypical lipomatous tumour / well-differentiated liposarcoma & -- & 1 & Low-grade fibromyxoid sarcoma & -- & 1 \\
Atypical spindle cell / pleomorphic lipomatous tumour & -- & 1 & Malignant glomus tumour & -- & 1 \\
Benign fibrous histiocytoma (dermatofibroma) & 1 & 1 & Malignant granular cell tumour & -- & 1 \\
Calcifying aponeurotic fibroma & -- & 1 & Malignant PEComa & -- & 1 \\
Calcifying fibrous tumour & -- & 1 & Malignant peripheral nerve sheath tumour & 1 & 1 \\
Capillary haemangioma & 1 & -- & Mesenchymal chondrosarcoma & -- & 1 \\
Cavernous haemangioma & 1 & 1 & Metastasizing leiomyoma & -- & 1 \\
Cellular angiofibroma & -- & 1 & Mixed tumour NOS & -- & 1 \\
Cellular leiomyoma & -- & 1 & Myoepithelioma & -- & 1 \\
Chondroid lipoma & 1 & 1 & Myolipoma of soft tissue & 1 & 1 \\
Chondrolipoma & -- & 1 & Myxofibrosarcoma & -- & 1 \\
Clear cell chondrosarcoma & -- & 1 & Myxoid liposarcoma & 1 & 1 \\
Clear cell sarcoma NOS & -- & 1 & Myxoinflammatory fibroblastic sarcoma & -- & 1 \\
Dedifferentiated chordoma & -- & 1 & Neurofibroma & 1 & -- \\
Dedifferentiated liposarcoma & -- & 1 & Nodular fasciitis & 1 & 1 \\
Deep (aggressive) angiomyxoma & -- & 1 & Nuchal fibroma & -- & 1 \\
Deep benign fibrous histiocytoma & -- & 1 & Ossifying fibromyxoid tumour NOS & -- & 1 \\
Dermatofibrosarcoma protuberans & 1 & -- & Osteosarcoma, extraskeletal & -- & 1 \\
Desmoid-type fibromatosis & -- & 1 & Palmar/plantar-type fibromatosis & -- & 1 \\
Desmoplastic fibroma & -- & 1 & Papillary intralymphatic angioendothelioma & -- & 1 \\
Desmoplastic small round cell tumour & -- & 1 & Phosphaturic mesenchymal tumour, NOS & -- & 1 \\
Elastofibroma & -- & 1 & Phyllodes tumour, malignant & -- & 1 \\
Eosinophilic fasciitis & -- & 1 & Pleomorphic hyalinizing angiectatic tumour & -- & 1 \\
Epithelioid angiosarcoma & -- & 1 & Pleomorphic liposarcoma & 1 & 1 \\
Epithelioid haemangioendothelioma & 2 & 1 & Pleomorphic rhabdomyosarcoma & 1 & 1 \\
Epithelioid haemangioma & -- & 1 & Pleomorphic sarcoma & -- & 1 \\
Epithelioid leiomyoma & -- & 1 & Plexiform fibrohistiocytic tumour & -- & 1 \\
Epithelioid liposarcoma & -- & 1 & Postradiation angiosarcoma & -- & 1 \\
Epithelioid malignant peripheral nerve sheath tumour & -- & 1 & Proliferative fasciitis & -- & 1 \\
Epithelioid sarcoma & -- & 1 & Proliferative myositis & -- & 1 \\
Epithelioid schwannoma & -- & 1 & Pseudomyogenic (epithelioid sarcoma-like) haemangioendothelioma & -- & 1 \\
Ewing sarcoma & -- & 1 & Pyogenic granuloma (lobular capillary haemangioma) & 1 & -- \\
Extrarenal rhabdoid tumour & -- & 1 & Rapidly involuting congenital haemangioma & -- & 1 \\
Fetal rhabdomyoma & -- & 1 & Schwannoma & 1 & -- \\
Fibro-osseous pseudotumour of digits & -- & 1 & Sclerosing epithelioid fibrosarcoma & -- & 1 \\
Fibroma of tendon sheath & -- & 1 & Signet ring stromal tumour & -- & 1 \\
Fibrous hamartoma of infancy & -- & 1 & Smooth muscle hamartoma & -- & 1 \\
Gardner fibroma & -- & 1 & Smooth muscle tumour of uncertain malignant potential & -- & 1 \\
Giant cell fibroblastoma & -- & 1 & Solitary fibrous tumour & -- & 1 \\
Giant cell tumour of soft parts NOS & -- & 1 & Solitary fibrous tumour, malignant & -- & 1 \\
Glomangiomatosis & -- & 1 & Spindle cell lipoma & 1 & 1 \\
Glomus tumour & 1 & 1 & Spindle cell rhabdomyosarcoma & 1 & 1 \\
Glomus tumour of uncertain malignant potential & -- & 1 & Superficial angiomyxoma & -- & 1 \\
Granular cell tumor & -- & 1 & Superficial CD34-positive fibroblastic tumour & -- & 1 \\
Granulomatous hemangioma & -- & 1 & Superficial leiomyosarcoma & -- & 1 \\
Haemangioma & -- & 1 & Synovial sarcoma & 1 & 1 \\
Haemosiderotic fibrolipomatous tumour & -- & 1 & Synovial sarcoma, biphasic & -- & 1 \\
Hibernoma & 1 & 1 & Synovial sarcoma, monophasic & -- & 1 \\
Histiocytic sarcoma & -- & 1 & Synovial sarcoma, poorly differentiated & -- & 1 \\
Infantile fibrosarcoma & -- & 1 & Tenosynovial giant cell tumour NOS & -- & 1 \\
Inflammatory myofibroblastic tumour & -- & 1 & Thromboangiitis obliterans & -- & 1 \\
Intramuscular haemangioma & -- & 1 &  &  &  \\[4pt]
\multicolumn{6}{@{}l}{\textbf{Lymph node, spleen \& marrow} (Teaching 23 cases; Atlas 56)}\\*[2pt]
ALK-positive large B-cell lymphoma & -- & 1 & Kikuchi-Fujimoto disease & 1 & 1 \\*
Anaplastic large cell lymphoma & 1 & 1 & Kimura disease & -- & 1 \\
Angioimmunoblastic T-cell lymphoma & 1 & 1 & Langerhans cell histiocytosis & -- & 1 \\
B-lymphoblastic leukaemia/lymphoma & 1 & 1 & Large cell lymphoma involving spleen (B- and T-cell types) & 1 & -- \\
Blastic plasmacytoid dendritic cell neoplasm & -- & 1 & Leukocytoclastic vasculitis & -- & 1 \\
Burkitt lymphoma & 1 & 1 & Littoral cell angioma (spleen) & 1 & -- \\
Castleman disease & -- & 1 & Lymphangioma (lymphatic malformation) & -- & 1 \\
Cat-scratch disease & -- & 1 & Lymphocyte-rich classic Hodgkin lymphoma & -- & 1 \\
Chronic lymphocytic leukaemia / small lymphocytic lymphoma & 2 & 1 & Lymphoepithelial carcinoma & -- & 1 \\
Chronic myeloid leukaemia, BCR-ABL1-positive & -- & 1 & Lymphoplasmacytic lymphoma (Waldenström macroglobulinaemia) & -- & 1 \\
Classic Hodgkin lymphoma, lymphocyte depletion & -- & 1 & Mantle cell lymphoma & 1 & 1 \\
Classic Hodgkin lymphoma, NOS & -- & 1 & Mixed cellularity classic Hodgkin lymphoma & -- & 1 \\
Dermatopathic lymphadenopathy & 1 & -- & Multiple myeloma & -- & 1 \\
Diffuse large B-cell lymphoma & -- & 1 & Nodal marginal zone lymphoma & 1 & 1 \\
Diffuse large B-cell lymphoma associated with chronic inflammation & -- & 1 & Nodular sclerosis classic Hodgkin lymphoma & -- & 1 \\
Enteropathy-associated T-cell lymphoma & 1 & -- & Peripheral T-cell lymphoma, NOS & 1 & 1 \\
Erdheim-Chester disease & -- & 1 & Plasmablastic lymphoma & -- & 1 \\
Essential thrombocythaemia & -- & 1 & Polycythaemia vera & -- & 1 \\
Extranodal marginal zone lymphoma of mucosa-associated lymphoid tissue & -- & 1 & Primary myelofibrosis, overt fibrotic stage & -- & 1 \\
Extranodal NK/T-cell lymphoma & 1 & 1 & Rosai-Dorfman disease & 1 & 1 \\
Follicular dendritic cell sarcoma & -- & 1 & Sarcoidosis & 1 & -- \\
Follicular lymphoma & 1 & 1 & Sclerosing Hemangioma-like Nodular Transformation of the Spleen & -- & 1 \\
Fungal lymphadenitis & 1 & 1 & Splenic diffuse red pulp small B-cell lymphoma & -- & 1 \\
Hairy cell leukaemia & -- & 1 & Splenic marginal zone lymphoma & 1 & 1 \\
Hepatosplenic T-cell lymphoma & 1 & 1 & Syphilitic lymphadenitis & -- & 1 \\
Hodgkin lymphoma & 1 & 1 & Systemic Lupus Erythematosus & -- & 1 \\
Hodgkin lymphoma, nodular lymphocyte-predominant type & -- & 1 & T-cell/histiocyte-rich large B-cell lymphoma & -- & 1 \\
Infectious mononucleosis & -- & 1 & T-lymphoblastic leukaemia/lymphoma & -- & 1 \\
Interdigitating dendritic cell sarcoma & -- & 1 & Toxoplasma lymphadenitis & 1 & 1 \\
Intravascular large B-cell lymphoma & -- & 1 & Tuberculosis & -- & 1 \\
Kaposi sarcoma & -- & 1 &  &  &  \\[4pt]
\multicolumn{6}{@{}l}{\textbf{Bone \& joint} (Teaching 13 cases; Atlas 29)}\\*[2pt]
Aneurysmal bone cyst & -- & 1 & Langerhans cell histiocytosis & 1 & -- \\*
Aseptic necrosis & -- & 1 & Low-grade central osteosarcoma & -- & 1 \\
Calcium pyrophosphate dihydrate crystal deposition (CPPD) & -- & 1 & Non-ossifying fibroma & -- & 1 \\
Chondroblastoma & 1 & 1 & Ossifying fibroma & 1 & 1 \\
Chondromyxoid fibroma & 1 & 1 & Osteoblastoma & 1 & 1 \\
Chondrosarcoma & 1 & 1 & Osteochondroma & 1 & 1 \\
Chronic osteomyelitis & -- & 1 & Osteoid osteoma NOS & -- & 1 \\
Dedifferentiated chondrosarcoma & -- & 1 & Osteoma & -- & 1 \\
Enchondroma & 1 & 1 & Osteosarcoma & 1 & 1 \\
Fibrous dysplasia & 1 & 1 & Parosteal osteosarcoma & -- & 1 \\
Giant cell arteritis & -- & 1 & Periosteal osteosarcoma & -- & 1 \\
Giant cell tumour of bone & 1 & 1 & Primary intraosseous carcinoma, NOS & -- & 1 \\
Gout (tophus) & -- & 1 & Rheumatoid arthritis (synovitis) & 1 & 1 \\
High-grade surface osteosarcoma & -- & 1 & Simple Bone Cyst & -- & 1 \\
Juxta-articular myxoma & -- & 1 & Synovial chondromatosis & 1 & 1 \\[4pt]
\multicolumn{6}{@{}l}{\textbf{Nervous system} (Teaching 12 cases; Atlas 49)}\\*[2pt]
Anaplastic astrocytoma (CNS WHO grade 3) & 1 & 1 & Lipomatosis of nerve & -- & 1 \\*
Anaplastic oligodendroglioma (CNS WHO grade 3) & 1 & 1 & Malignant melanotic nerve sheath tumour & -- & 1 \\
Angiocentric glioma & -- & 1 & Medulloblastoma & 1 & 1 \\
Astroblastoma, MN1-altered & -- & 1 & Melanotic neuroectodermal tumour of infancy & -- & 1 \\
Atherosclerosis & 1 & -- & Meningioma & -- & 1 \\
Atypical carcinoid / neuroendocrine tumour, grade 2 & -- & 1 & Microcystic/reticular schwannoma & -- & 1 \\
Atypical choroid plexus papilloma & -- & 1 & Nerve sheath myxoma & -- & 1 \\
Atypical teratoid/rhabdoid tumour & -- & 1 & Neuroblastoma & -- & 1 \\
Cellular schwannoma & -- & 1 & Neuroendocrine tumour & -- & 1 \\
Chordoma & -- & 1 & Neurofibroma & -- & 1 \\
Choroid plexus carcinoma & -- & 1 & Oligodendroglioma (CNS WHO grade 2) & 1 & 1 \\
Choroid plexus tumour (papilloma) & 1 & -- & Papillary tumour of the pineal region & -- & 1 \\
Craniopharyngioma & -- & 1 & Perineurioma & -- & 1 \\
Diffuse astrocytoma (CNS WHO grade 2) & 1 & 1 & Pilocytic astrocytoma & 1 & 1 \\
Endolymphatic sac tumour & -- & 1 & Pineoblastoma & -- & 1 \\
Ependymoblastoma & -- & 1 & Pleomorphic xanthoastrocytoma & 1 & 1 \\
Ependymoma & 1 & 1 & Plexiform neurofibroma & -- & 1 \\
Epidemic cerebrospinal meningitis & -- & 1 & Primary diffuse large B-cell lymphoma of the CNS & -- & 1 \\
Ganglioglioma & -- & 1 & Primitive neuroectodermal tumour NOS & -- & 1 \\
Ganglioneuroblastoma & -- & 1 & Schwannoma & -- & 1 \\
Ganglioneuroma & -- & 1 & Subependymal giant cell astrocytoma & -- & 1 \\
Glioblastoma & 1 & 1 & Subependymoma & -- & 1 \\
Haemangioblastoma & 1 & 1 & Traumatic neuroma & -- & 1 \\
Hybrid nerve sheath tumour & -- & 1 & Uveal melanoma & -- & 1 \\
Japanese Encephalitis & -- & 1 & Vagal paraganglioma & -- & 1 \\
Jugulotympanic paraganglioma & -- & 1 &  &  &  \\[4pt]
\multicolumn{6}{@{}l}{\textbf{Skin} (Teaching 21 cases; Atlas 91)}\\*[2pt]
Acanthosis nigricans & -- & 1 & Fixed Drug Eruption & -- & 1 \\*
Acquired tufted haemangioma & -- & 1 & Foliaceous pemphigus & -- & 1 \\
Acute eczema & -- & 1 & Gout (tophus) & 1 & -- \\
Acute Febrile Neutrophilic Dermatosis & -- & 1 & Herpes Simplex & -- & 1 \\
Acute generalized exanthematous pustulosis & -- & 1 & Herpetiform dermatitis & -- & 1 \\
Acute Graft-versus-Host Disease & -- & 1 & Hydroa vacciniforme-like lymphoproliferative disorder & -- & 1 \\
Alopecia areata & -- & 1 & Intradermal melanocytic naevus & 1 & 1 \\
Amyloidosis & -- & 1 & Junctional melanocytic naevus & 1 & 1 \\
Annular granuloma & -- & 1 & Juvenile xanthogranuloma & -- & 1 \\
Apocrine adenoma & -- & 1 & Keloid & 1 & 1 \\
Atypical apocrine adenosis & -- & 1 & Keratoacanthoma & 1 & 1 \\
Atypical fibroxanthoma & -- & 1 & Langerhans cell sarcoma & -- & 1 \\
Atypical melanocytic naevus of genital type & -- & 1 & Lentigo simplex & 1 & -- \\
Atypical vascular lesion & -- & 1 & Lepromatous leprosy & -- & 1 \\
Bacillary angiomatosis & -- & 1 & Lichen planus & -- & 1 \\
Basal cell carcinoma & 1 & 1 & Lichen sclerosus & 1 & 1 \\
Basaloid squamous cell carcinoma & -- & 1 & Lupus vulgaris & -- & 1 \\
Blue naevus & 1 & 1 & Lymphadenoma & -- & 1 \\
Bowen disease (squamous cell carcinoma in situ of skin) & 1 & 1 & Lymphomatoid papulosis & -- & 1 \\
Bullous Congenital Ichthyosiform Erythroderma & -- & 1 & Melanoma & -- & 1 \\
Bullous Pemphigoid & -- & 1 & Merkel cell carcinoma & -- & 1 \\
Capillary haemangioma & -- & 1 & Molluscum contagiosum & 1 & 1 \\
Cellular blue naevus & -- & 1 & Multicentric Reticulohistiocytosis & -- & 1 \\
Chondroid syringoma NOS & -- & 1 & Nodular melanoma & -- & 1 \\
Chromoblastomycosis & -- & 1 & Non-involuting congenital haemangioma & -- & 1 \\
Chronic actinic dermatitis & -- & 1 & Paget disease (mammary / extramammary) & 1 & -- \\
Chronic discoid lupus erythematosus & -- & 1 & Papillary hidradenoma & -- & 1 \\
Chronic eczema & -- & 1 & Pemphigus vulgaris & -- & 1 \\
Clear cell hidradenoma & -- & 1 & Pilomatricoma & 1 & 1 \\
Compound naevus & -- & 1 & Pityriasis rubra pilaris & -- & 1 \\
Congenital melanocytic naevus NOS & -- & 1 & Polyarteritis nodosa & -- & 1 \\
Cutaneous horn & -- & 1 & Porokeratosis & -- & 1 \\
Cutaneous leiomyoma & -- & 1 & Primary cutaneous DLBCL, leg type & -- & 1 \\
Cutaneous mastocytosis & -- & 1 & Primary cutaneous follicle centre lymphoma & -- & 1 \\
Cylindroma & -- & 1 & Psoriasis & -- & 1 \\
Darier disease (keratosis follicularis) & -- & 1 & Pustular Psoriasis & -- & 1 \\
Dermatofibrosarcoma protuberans & -- & 1 & Pyogenic granuloma (lobular capillary haemangioma) & -- & 1 \\
Dermatomyositis & -- & 1 & Sebaceous adenoma & -- & 1 \\
Dysplastic naevus & -- & 1 & Sebaceous carcinoma & -- & 1 \\
Eccrine Poroma & -- & 1 & Seborrheic keratosis & 1 & 1 \\
Eosinophilic Pustular Folliculitis & -- & 1 & Sporotrichosis & -- & 1 \\
Ephelis (freckle) & 1 & 1 & Subungual exostosis & -- & 1 \\
Epidermoid (epidermal inclusion) cyst & 1 & 1 & Suppurative hidradenitis & -- & 1 \\
Erythema induratum & -- & 1 & Trichoepithelioma (trichoblastoma) & 1 & 1 \\
Erythema multiforme & -- & 1 & Tuberculosis & 1 & -- \\
Erythema Nodosum & -- & 1 & Verruca vulgaris & 1 & 1 \\
Erythroplasia of Queyrat (penile SCC in situ) & 1 & -- & Xanthoma & -- & 1 \\
Fibroepithelial polyp (acrochordon) & 1 & -- & Xeroderma Pigmentosum & -- & 1 \\
Fibrosarcomatous dermatofibrosarcoma protuberans & -- & 1 &  &  &  \\[4pt]
\multicolumn{6}{@{}l}{\textbf{Ear \& eye} (Teaching 8 cases; Atlas 18)}\\*[2pt]
Angioleiomyoma & -- & 1 & Middle ear neuroendocrine tumour (middle ear adenoma) & 1 & 1 \\*
Ceruminous adenocarcinoma & 1 & 1 & Preauricular sinus & 1 & 1 \\
Ceruminous adenoma & -- & 1 & Pterygium & -- & 1 \\
Cholesterol granuloma & 1 & 1 & Retinoblastoma & 1 & 1 \\
Coats disease & -- & 1 & Sarcoidosis & -- & 1 \\
Composite haemangioendothelioma & -- & 1 & Squamous cell carcinoma, keratinizing, NOS & -- & 1 \\
Dermoid cyst & -- & 1 & Squamous papilloma & -- & 1 \\
Endolymphatic sac tumour & 1 & -- & Sympathetic ophthalmia & -- & 1 \\
Inflammatory pseudotumour & -- & 1 & Uveal melanoma & 1 & -- \\
Middle ear cholesteatoma & 1 & 1 & Venous haemangioma & -- & 1 \\[4pt]
\midrule
\multicolumn{3}{@{}l}{\textbf{Total: 843 entities (347 in both sources)}} & \multicolumn{3}{r@{}}{\textbf{Teaching \detailCases\ cases; Atlas \detailAtlCases}}\\
\end{longtable}
\endgroup

\begin{center}
\includegraphics[width=0.9\textwidth]{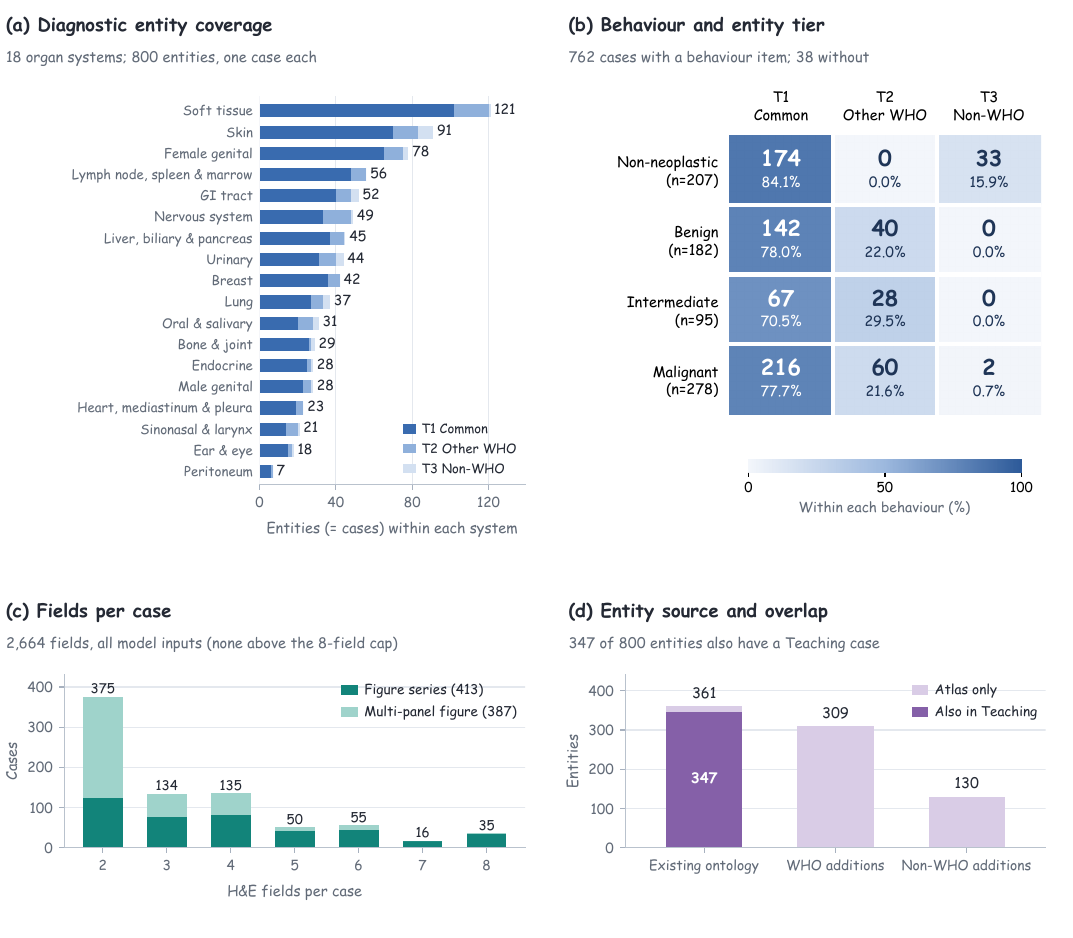}
\captionof{figure}{\textbf{Detailed composition of \bench-Atlas}, the counterpart of \cref{fig:dataset-detail}. Of its 800 cases, 413 show one lesion and 387 are figure series of one disease, possibly from several patients (Sec.~3.6\ofmain). \bench-Atlas has one case per entity, and its H\&E-sufficiency prior is weak---for \detailAtlHeFallback\ of the \detailAtlAdded\ entities added to the ontology for Atlas the two LLMs that assigned it disagreed and it defaults to P---so panels (b) and (d) differ.
(a) Entities, equal to cases, within each organ system by tier: T1, common diagnosis (in three or more of the source books or a core residency diagnosis; \detailAtlTone); T2, other WHO entity (\detailAtlTtwo); T3, other non-WHO diagnosis (\detailAtlTthree).
(b) Case counts and within-behaviour percentages by tier for the \detailAtlWithB\ cases with a behaviour item; the other \detailAtlWithoutB\ entities have low-confidence behaviour codes and no B item (\cref{supp:ontology}).
(c) Fields per case by unit: a series of figures of one case (\detailAtlSeries\ cases) or one multi-panel figure (\detailAtlMultiPanel); all \detailAtlFields\ fields are model inputs.
(d) Source of the entities: the existing ontology (\detailAtlOntology), WHO additions (\detailAtlWho) and non-WHO additions (\detailAtlNew); \detailAtlShared\ entities also have a Teaching case.}
\label{fig:dataset-detail-atlas}
\end{center}

\let\suppOnecolumn\onecolumn\let\onecolumn\relax
\onecolumn
\raggedbottom
\section{Prompts}
\label{supp:prompts}
All prompts are reproduced from the evaluation code and the released task files. Models receive one user turn and no system prompt: each field is preceded by its label (\texttt{Image $i$:}) and followed, after the last field, by the question, the options and the answer instruction (the instruction follows the options, as in MMBench). Decoding is greedy (temperature 0) with up to 1,024 new tokens, 4,096 for reasoning models. \{\textit{...}\} marks a field filled per item, [\,\dots] a sentence present only when it applies (the specimen site is omitted for the cases filed under several organ chapters).
\subsection*{1--5. Task prompts}
\par\medskip\noindent\textbf{1. A1, open-ended diagnosis.}\par\smallskip
\par\noindent\fbox{\parbox{\dimexpr\linewidth-2\fboxsep-2\fboxrule}{\raggedright\footnotesize\ttfamily
The following \{\textit{k}\} H\&E-stained microscopy images are fields from the same lesion of one case, ordered from low to high magnification. [Specimen site: \{\textit{site}\}.] Give the most likely pathological diagnosis.\\
Integrate all images. Output only this JSON: \{"diagnosis": "most likely pathological diagnosis", "differential": ["differential 1", "differential 2"], "confidence": integer 0-100\}}}\par\smallskip

\par\medskip\noindent\textbf{2. A2, ten-way diagnosis (asked once per rotation of the options).}\par\smallskip
\par\noindent\fbox{\parbox{\dimexpr\linewidth-2\fboxsep-2\fboxrule}{\raggedright\footnotesize\ttfamily
The following \{\textit{k}\} H\&E-stained microscopy images are fields from the same lesion of one case, ordered from low to high magnification. [Specimen site: \{\textit{site}\}.] What is the most likely pathological diagnosis?\\
Options:\\
A. \{\textit{option A}\}\\
...\\
J. \{\textit{option J}\}\\
Integrate all images and answer with the option's letter from the given choices directly.}}\par\smallskip

\par\medskip\noindent\textbf{3. B, behaviour triage (the four options are the same for every case; asked in four rotations).}\par\smallskip
\par\noindent\fbox{\parbox{\dimexpr\linewidth-2\fboxsep-2\fboxrule}{\raggedright\footnotesize\ttfamily
The following \{\textit{k}\} H\&E-stained microscopy images are fields from the same lesion of one case, ordered from low to high magnification. [Specimen site: \{\textit{site}\}.] Which category best describes this lesion?\\
Options:\\
A. Non-neoplastic lesion (inflammatory, infectious, reactive, cystic, developmental, etc.)\\
B. Benign neoplasm\\
C. Borderline / uncertain malignant potential, in situ or precursor lesion\\
D. Malignant neoplasm\\
Integrate all images and answer with the option's letter from the given choices directly.}}\par\smallskip

\par\medskip\noindent\textbf{4. C, ordinal grading (all levels of the case's group, two to four; asked once per rotation).}\par\smallskip
\par\noindent\fbox{\parbox{\dimexpr\linewidth-2\fboxsep-2\fboxrule}{\raggedright\footnotesize\ttfamily
The following \{\textit{k}\} H\&E-stained microscopy images are fields from the same lesion of one case, ordered from low to high magnification. [Specimen site: \{\textit{site}\}.] Determine the grade of this lesion.\\
Options:\\
A. \{\textit{level 1}\}\\
...\\
\{\textit{level n}\}\\
Integrate all images and answer with the option's letter from the given choices directly.}}\par\smallskip

\par\medskip\noindent\textbf{5. R1, finding verification (one item per case and level; for the high-power item the question reads \texttt{high-power cytology} and every statement starts with \texttt{At high power,}).}\par\smallskip
\par\noindent\fbox{\parbox{\dimexpr\linewidth-2\fboxsep-2\fboxrule}{\raggedright\footnotesize\ttfamily
The following \{\textit{k}\} H\&E-stained microscopy images are fields from the same lesion of one case, ordered from low to high magnification. [Specimen site: \{\textit{site}\}.] Judge whether each of the following four statements about the low-power architecture of this case is true or false:\\
1. At low power, \{\textit{statement 1}\}\\
...\\
4. At low power, \{\textit{statement 4}\}\\
Integrate all images. Output only JSON in this format, with four judgements in statement order, each true (correct) or false (incorrect): \{"answers": [true/false, true/false, true/false, true/false]\}}}\par\smallskip

\subsection*{6. A complete A2 item (case C0001, first rotation; the seven fields precede the text)}
\par\noindent\fbox{\parbox{\dimexpr\linewidth-2\fboxsep-2\fboxrule}{\raggedright\footnotesize\ttfamily
The following 7 H\&E-stained microscopy images are fields from the same lesion of one case, ordered from low to high magnification. Specimen site: skin. What is the most likely pathological diagnosis?\\
Options:\\
A. Solitary fibrous tumour\\
B. Junctional melanocytic naevus\\
C. Melanoma\\
D. Ephelis (freckle)\\
E. Blue naevus\\
F. Lentigo simplex\\
G. Schwannoma\\
H. Keloid\\
I. Intradermal melanocytic naevus\\
J. Seborrheic keratosis\\
Integrate all images and answer with the option's letter from the given choices directly.}}\par\smallskip

\subsection*{7. Evidence ablations}
The ablations change only the evidence and the sentences that describe it; options and instructions are unchanged unless stated. \emph{Highest-power field only} (A2; only the last field is given):
\par\noindent\fbox{\parbox{\dimexpr\linewidth-2\fboxsep-2\fboxrule}{\raggedright\footnotesize\ttfamily
The following H\&E-stained microscopy image is a field from the lesion of one case. Specimen site: skin. What is the most likely pathological diagnosis?}}\par\smallskip

\emph{Without specimen site} (A2; the site sentence is removed):
\par\noindent\fbox{\parbox{\dimexpr\linewidth-2\fboxsep-2\fboxrule}{\raggedright\footnotesize\ttfamily
The following 7 H\&E-stained microscopy images are fields from the same lesion of one case, ordered from low to high magnification. What is the most likely pathological diagnosis?}}\par\smallskip

\emph{Text alone} (R1; no fields are given, and the first sentence of the instruction is replaced):
\par\noindent\fbox{\parbox{\dimexpr\linewidth-2\fboxsep-2\fboxrule}{\raggedright\footnotesize\ttfamily
The following question concerns H\&E-stained sections of one lesion of one case (no images are provided). Specimen site: breast. Judge whether each of the following four statements about the low-power architecture of this case is true or false:\\
Answer from the text. Output only JSON in this format, with four judgements in statement order, each true (correct) or false (incorrect): \{"answers": [true/false, true/false, true/false, true/false]\}}}\par\smallskip

\emph{Single-image models} (LLaVA-Med, Quilt-LLaVA, PathGen-LLaVA) receive the fields tiled into one image, numbered in reading order, and the first sentence reads:
\par\noindent\fbox{\parbox{\dimexpr\linewidth-2\fboxsep-2\fboxrule}{\raggedright\footnotesize\ttfamily
The image is a mosaic of 7 numbered H\&E fields (1-7, from low to high magnification) from the same lesion of one case.}}\par\smallskip

\par\noindent\begin{minipage}{\linewidth}
\subsection*{8. A1 judge (Qwen2.5-32B-Instruct, AWQ; temperature 0, at most 160 tokens)}
Fields come from the ontology entry of the reference (\textit{ref\_en} and \textit{ref\_zh}: the reference diagnosis in English and Chinese; \textit{syn}: synonyms and equivalent terms; \textit{alt}: accepted alternates; \textit{fam}: the other entities of its family with their behaviour) and from the parsed answer. The item score is the higher of the judge grade (0, 0.5, 1 for \texttt{primary} = 0, 1, 2) and the deterministic ontology match on the parsed answer (Sec.~4\ofmain).
\par\noindent\fbox{\parbox{\dimexpr\linewidth-2\fboxsep-2\fboxrule}{\raggedright\footnotesize\ttfamily
You are an experienced surgical pathologist grading a trainee's diagnosis against the reference diagnosis of a case.\\
\mbox{}\\
Specimen site: \{\textit{site}\}\\
Reference diagnosis: \{\textit{ref\_en}\} (\{\textit{ref\_zh}\})\\
Accepted synonyms and equivalent terms: \{\textit{syn}\}\\
Also accepted (alternate reference): \{\textit{alt}\}\\
Reference behaviour: \{\textit{beh}\}\\
Other entities in the same diagnostic family (with behaviour): \{\textit{fam}\}\\
\mbox{}\\
Trainee's primary diagnosis: \{\textit{pred}\}\\
Trainee's differential diagnoses: \{\textit{ddx}\}\\
\mbox{}\\
Grading rules (apply in order):\\
- primary = 2 if the primary diagnosis names the reference entity: any synonym, translation, outdated equivalent term, an\\
\hspace*{1.0em}accepted alternate, or a MORE SPECIFIC subtype/variant of the reference (e.g. "invasive pituitary adenoma" for pituitary\\
\hspace*{1.0em}adenoma, "keratinizing squamous cell carcinoma" for squamous cell carcinoma).\\
- primary = 1 if (a) it names a correct BROADER category that contains the reference (e.g. "breast carcinoma" for invasive\\
\hspace*{1.0em}ductal carcinoma NST, "lymphoma" for follicular lymphoma), or (b) it names a DIFFERENT entity of the same diagnostic family\\
\hspace*{1.0em}that has the SAME behaviour class as the reference (e.g. adenoid cystic carcinoma for epithelial-myoepithelial carcinoma).\\
- primary = 0 otherwise. In particular, any answer whose behaviour class differs from the reference (benign/reactive vs\\
\hspace*{1.0em}malignant, non-neoplastic vs neoplastic, in situ vs invasive) scores 0 even within the same family, as does no answer.\\
- top3 = true if the reference (score-2 level) appears in the primary or any differential diagnosis.\\
- error = "correct" if primary is 2; "within\_family" if primary is 1; otherwise one of "wrong\_family\_same\_lineage",\\
\hspace*{1.0em}"wrong\_lineage", "wrong\_behaviour", "no\_answer".\\
\mbox{}\\
Reply with JSON only: \{"primary": 0|1|2, "top3": true|false, "error": "...", "reason": "<= 25 words"\}}}\par\smallskip

\end{minipage}\par\medskip
\subsection*{9. Answer extraction for multiple-choice responses the parser cannot map (same model; temperature 0, at most 4 tokens)}
\texttt{Z} counts as no answer. The extracted letter is used only when the deterministic parser finds no option.
\par\noindent\fbox{\parbox{\dimexpr\linewidth-2\fboxsep-2\fboxrule}{\raggedright\footnotesize\ttfamily
Below are the answer options of a multiple-choice question and a model's response to it.\\
Which single option does the response choose as its final answer? Reply with only the option letter.\\
If the response does not clearly commit to exactly one option (e.g. it refuses, names several, or answers something else), reply Z.\\
\mbox{}\\
Options:\\
\{\textit{opts}\}\\
\mbox{}\\
Response:\\
\{\textit{resp}\}\\
\mbox{}\\
Letter:}}\par\smallskip

\par\noindent\begin{minipage}{\linewidth}
\subsection*{10. R1: authoring specification and checks}
An LLM (DeepSeek-V4.1-Flash, release 1017) first generated the R1 routes and statements for each case according to the specification below. A pathologist then reviewed the generated content. All released items also passed the automated checks described below.
\par\noindent\fbox{\parbox{\dimexpr\linewidth-2\fboxsep-2\fboxrule}{\raggedright\footnotesize\ttfamily
Input per case: the reference entity (expert label) and its WHO criteria; the specimen site; the entities offered as distractors in the case's A2 item, including its classic look-alikes; for graded entities, the levels of its group.\\
Cases: every case whose entity is diagnosable from H\&E alone, plus the case of Fig. 1 (195 cases).\\
Write a route: low-power architecture -> high-power cytology -> differential (two look-alikes, each with the finding that separates it) -> conclusion.\\
For each level (low power, high power) write four statements, exactly two true and two false:\\
- true: an essential criterion of the reference entity that is seen at this level;\\
- false: a defining finding of one of the case's look-alikes or distractors (or of another level of the same grading group) that the reference excludes at this level; record the supporting diagnosis or grading level and its relation to the reference;\\
- one finding per statement, phrased as what is seen ("At low power, ..."), in English and Chinese;\\
- never name the reference entity, its synonyms, any option of the case's A2 item, or the site;\\
- no true/false pair that mirrors one another; true and false statements alike in length and in the use of negation.}}\par\smallskip

\end{minipage}\par\medskip
Automated checks: four statements per level with exactly two true; every false statement has a basis that names an ontology entity or grading level; differential entities exist in the ontology; no statement contains the name or a synonym (four characters or longer) of the reference or of any A2 option of the case; statement order is shuffled per item with a fixed seed. Answer-free heuristics then score at chance on the 1,560 statements (Sec.~4\ofmain: two shortest statements true, 50\%; negated statements true, 51\%). A released item (case of Fig.~1, low power), with the hidden truth values and supporting diagnoses written out in full:
\par\noindent\fbox{\parbox{\dimexpr\linewidth-2\fboxsep-2\fboxrule}{\raggedright\footnotesize\ttfamily
1. At low power, the expanded stroma projects into cleft-like spaces as broad leaf-like fronds, and the border of the lesion is infiltrative.  [true; basis: reference diagnostic criterion]\\
2. At low power, single-file cords of small discohesive cells infiltrate the stroma, without a biphasic epithelial-stromal pattern.  [false; basis: distractor: Invasive lobular carcinoma]\\
3. At low power, benign epithelium-lined clefts and ducts are set in a markedly expanded, hypercellular stroma.  [true; basis: reference diagnostic criterion]\\
4. At low power, the lesion is sharply circumscribed, with an even intracanalicular or pericanalicular pattern and no stromal overgrowth.  [false; basis: distractor: Cellular fibroadenoma]}}\par\smallskip

\twocolumn
\flushbottom

\let\onecolumn\suppOnecolumn

\section{Worked examples with model answers}
\label{supp:examples}
\begin{figure*}[p]
\centering
\includegraphics[width=\textwidth]{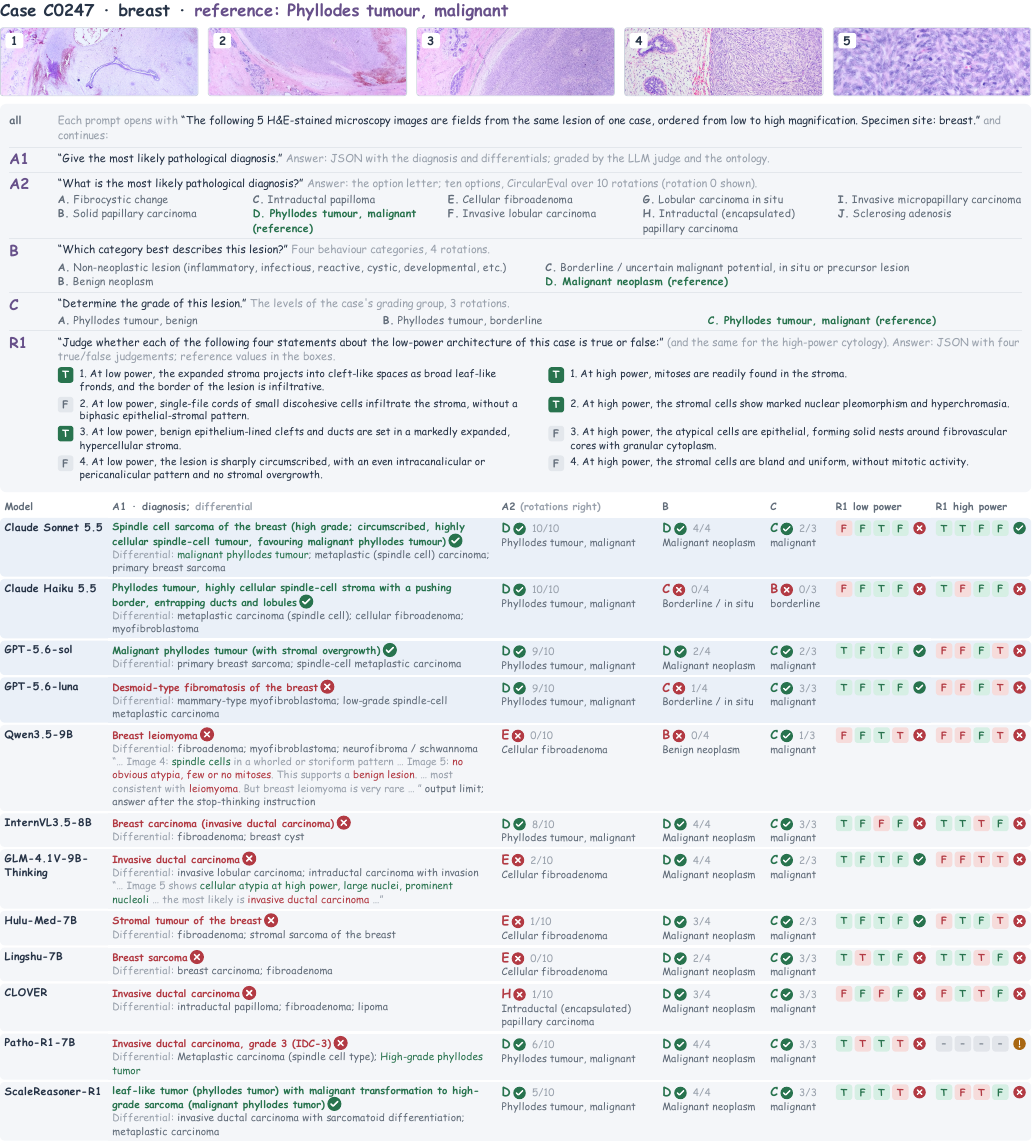}
\caption{\textbf{One teaching case, five tasks: the prompts and the answers of twelve models.} Case C0247 (malignant phyllodes tumour, breast). Top: the five fields and, per task, the prompt (its shared opening once), the options with the reference in green and, for R1, the eight statements with their reference truth values. Bottom: each model's answers at rotation 0 (blue rows: proprietary models). A1: the diagnosis (green/red: reference entity or synonym/other), the differential the model returned (reference mentions in green) and, for the two reasoning models, an excerpt of the reasoning (green/red: findings consistent/inconsistent with the reference; ellipses mark omissions). A2, B, C: the option chosen, the number of rotations answered correctly and the option's text. R1: the four judgements per level (green: matches the reference) and whether all four were right; Patho-R1-7B returned no parseable high-power answer. Qwen3.5-9B reached the output limit while still reasoning on A1, C and low-power R1 and answered after the stop-thinking instruction (Sec.~\ref{supp:parsing}).}
\label{fig:ex-case}\label{fig:ex-r1}
\end{figure*}

\begin{figure*}[t]
\centering
\includegraphics[width=\textwidth]{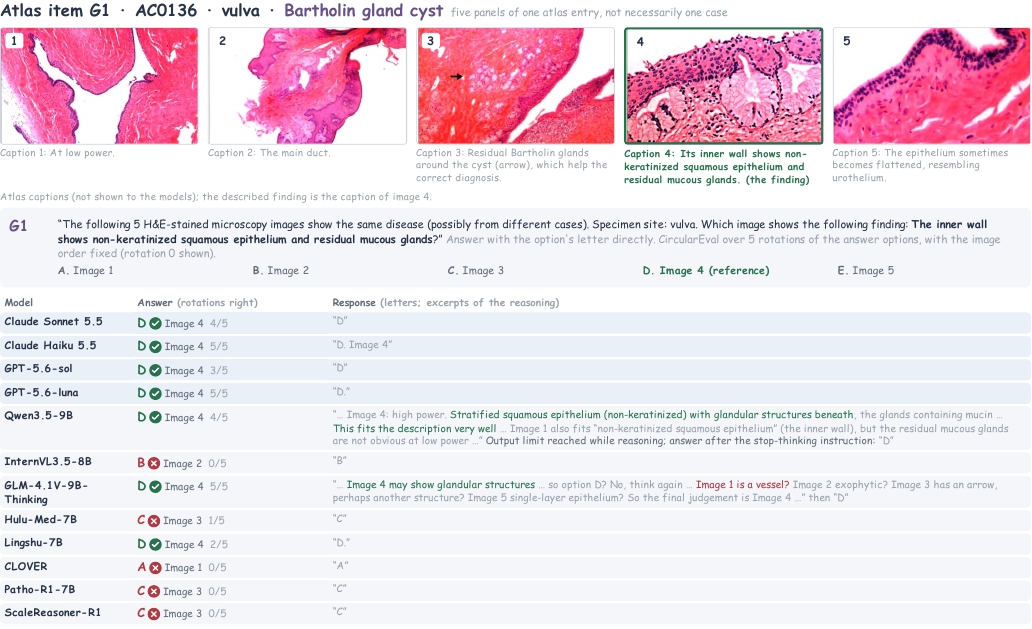}
\caption{\textbf{G1 on an atlas item.} Five panels of a Bartholin gland cyst (vulva) from one atlas entry, each with its own atlas caption (the captions are not shown to the models); the described finding is the caption of image 4, the reference (green frame). Middle: the prompt and its options. Bottom: each model's answer at option rotation 0, the image it names, the number of option rotations answered correctly (the options rotate, the images keep their order) and the response: the letter alone or, for the two reasoning models, excerpts of the reasoning (green/red: statements consistent/inconsistent with the panels and their captions). Qwen3.5-9B reached the output limit while still reasoning and answered after the stop-thinking instruction (Sec.~\ref{supp:parsing}).}
\label{fig:ex-g1}
\end{figure*}

\Cref{fig:ex-case,fig:ex-g1} show one complete item of every task together with the answers of the four proprietary and eight open models of Table~3\ofmain.
The teaching case C0247 (malignant phyllodes tumour of the breast, the case of Fig.~1\ofmain) is an item of all five teaching tasks; the atlas item is a Bartholin gland cyst.
Answers are those of option rotation 0; the number of rotations answered correctly shows how far an answer survives circular evaluation.
The marks of the A1 diagnoses follow the A1 grade of Sec.~4\ofmain, the higher of the LLM judge and the ontology match.

\section{Distractor policy and differential pairs}
\label{supp:distractors}
Answer and distractor may never come from the same exclusion group. Groups cover synonyms and \emph{in situ} equivalents (\eg Bowen disease, erythroplasia of Queyrat, CIN3, squamous carcinoma \emph{in situ}), parent--variant relations (lipoma and its variants, leiomyoma and its variants, fibroadenoma subtypes, gastric adenocarcinoma and its special types), site variants of the same morphology (adenocarcinomas of different organs, clear-cell carcinomas of kidney and female genital tract), and H\&E-inseparable pairs (ADH vs.\ low-grade DCIS, ALH vs.\ LCIS, parathyroid hyperplasia vs.\ adenoma, T- vs.\ B-lymphoblastic lymphoma, TFE3- vs.\ TFEB-altered RCC).

\section{Label quality control and human baseline protocols}
\label{supp:review}
\textbf{(1) Automated pre-review.} A multimodal LLM agent (DeepSeek-V4.1-Flash, release 1017) reviewed the contact sheet, label, site and the names under which the library filed it for every case, and the options of every A2 item, and listed its doubts with the evidence for each; it changed nothing. It flagged 64 cases: doubts about the diagnosis or its evidence (23 cases), naming or classification (16), a possible mixture of cases or unclear provenance (12), the stated site (9), teaching annotations on an image (7) and a possibly non-H\&E image (1); four cases carry two flags. It also flagged 7 A2 items, naming for each one distractor that might also be acceptable.
\textbf{(2) Pathologist confirmation.} The confirmation review was performed by a pathologist of attending rank or above, with completed specialist training, experience in routine H\&E diagnosis across organ systems, and familiarity with WHO diagnostic criteria. For each of the 64 flagged cases, the review form presented the contact sheet and the pre-review's note to the reviewer, who recorded whether the concern was valid, invalid or uncertain and selected a disposition: retain the case unchanged; correct the diagnosis or site; remove or crop an image; or split or remove the case. For each of the 7 flagged A2 items, the pathologist assessed whether the flagged distractor was also acceptable. The other 372 cases keep the diagnosis under which the teaching pathologists filed them, harmonised to the ontology; the pathologist's check of the field order (Sec.~\ref{supp:pipeline}) covers all 436 cases. Teaching annotations (arrows, circles) on 9 fields were previously confirmed by visual inspection. Review took approximately 2--3 hours for the flagged cases and 10 minutes for the A2 items.
\textbf{(3) Judge validation.} The A1 reviewer had the qualifications described above. The validation protocol presented \judgeN\ A1 responses without revealing the model identity or automatic grade and required the pathologist to assign each grade without LLM assistance. The grading rubric assigns 2 for the reference diagnosis, an accepted synonym or a more specific subtype; 1 for a correct broader category or another entity in the same family and behaviour class; and 0 otherwise. This task took approximately 1.5--2 hours. Each returned grade is compared with the automatic grade of the exact response shown on the sheet; archived runs are used when a model has subsequently been re-run. The available returned sheet yields exact agreement of \judgeExactFinal\% with the final A1 grade and quadratic-weighted $\kappa$ of \judgeKappaFinal\ (\judgeKappaJudge\ for the LLM judge alone); mean grades are \judgeMeanFinal\ for the final A1 grader and \judgeMeanRater\ for the returned sheet.
\textbf{Reader study.} The reader study included four experience groups: residents in years 1, 2 and 3, and attending pathologists. Table~\ref{tab:human-reader-accuracy} reports individual A2 scores for two readers in each group, together with mean A1 and A2 accuracies. Each reader received the same set of 120 cases stratified across 18 organ systems. The offline browser interface recorded an anonymous participant identifier and training level. For each case, participants first entered an open diagnosis and only then saw the ten A2 options; the time spent on each step was recorded separately. The instructions prohibited consulting reference materials or discussing answers with others. Pathologists exposed to the reference answers through quality control or judge validation were ineligible for the reader study.
A1 is reported as exact diagnosis accuracy; A2 is single-pass accuracy with one answer per case. Each reader's A2 accuracy is the number of correct answers divided by 120, expressed as a percentage; group A2 means are computed from unrounded reader accuracies, and all displayed values are rounded to one decimal place. The lower rows of Table~\ref{tab:human-reader-accuracy} give the A2 single-pass accuracy of the proprietary models, Qwen3.5-9B and Patho-R1-7B on the same \readerNCases\ items, with the same fields and option order (rotation 0), and their exact A1 accuracy on the same cases. The analysis script computes participant-level A2 accuracy and response times, and group accuracy with 95\% confidence intervals from 2,000 case-bootstrap resamples; open diagnoses are exported separately for A1 grading.
\par\smallskip
\noindent\begin{minipage}{\linewidth}
\centering
\color{black}
\footnotesize
\captionof{table}{\textbf{Reader and model accuracy on the \readerNCases\ reader cases} (\%). A2: single-pass multiple-choice accuracy; readers: two per group and their mean; models: one pass over the same items in the same option order (rotation 0). A1: exact-diagnosis accuracy on the same cases (readers: mean of the two; models: graded as in Table~\ref{tab:a1}).}
\label{tab:human-reader-accuracy}
\setlength{\tabcolsep}{3pt}
\renewcommand{\arraystretch}{1.08}
\begin{tabular}{@{}lccc@{}}
\toprule
Readers / model & \makecell{A2 readers\\1 / 2} & A2 & \makecell{A1\\exact} \\
\midrule
Residency year 1 & 56.7 / 63.3 & 60.0 & 40.0 \\
Residency year 2 & 66.7 / 70.8 & 68.8 & 52.5 \\
Residency year 3 & 75.8 / 80.0 & 77.9 & 65.0 \\
Attendings & 84.2 / 88.3 & 86.3 & 77.5 \\
\midrule
Claude Sonnet 5.5 & -- & \readerAtwoClaudeSonnet & \readerAoneClaudeSonnet \\
Claude Haiku 5.5 & -- & \readerAtwoClaudeHaiku & \readerAoneClaudeHaiku \\
GPT-5.6-sol & -- & \readerAtwoGptSol & \readerAoneGptSol \\
GPT-5.6-luna & -- & \readerAtwoGptLuna & \readerAoneGptLuna \\
Qwen3.5-9B & -- & \readerAtwoQwenThreeFiveNine & \readerAoneQwenThreeFiveNine \\
Patho-R1-7B & -- & \readerAtwoPathoRone & \readerAonePathoRone \\
\bottomrule
\end{tabular}
\end{minipage}
\par\smallskip

\section{Inference, answer parsing and forced answers}
\label{supp:parsing}
Every model may generate up to 1,024 new tokens, and up to 4,096 if it reasons before answering. The budgets are fixed in advance and equal within each group, so that all models answer under the same conditions and the cost of the evaluation stays bounded; a cap is needed in any case because some outputs do not end but repeat a phrase until the limit.
Without reasoning, 1,024 tokens rarely bind: apart from MedGemma-1.5-4B, \budgetOtherHits\ of \budgetOtherN\ responses reach them, whereas the long and often repetitive explanations of MedGemma-1.5-4B reach the limit in \budgetMedgemmaPct\% of its responses (\budgetMedgemmaAtlasAtwoPct\% on \bench-Atlas A2); such responses are parsed as they are.
The single-image LLaVA-Med, Quilt-LLaVA and PathGen-LLaVA run with the LLaVA code and budgets of 512 new tokens for A1, 128 for R1 and 64 for the multiple-choice tasks; \budgetLlavaAoneHits\ of their A1 answers reach 512.
Among the reasoning models, mainly two reach the budget of 4,096 tokens, as described below.
HuatuoGPT-Vision runs with its input limit of 4,096 tokens raised, so that no case is truncated.
A deterministic parser maps each multiple-choice response to an option: an option letter at the start of the response (bare, bracketed or in markdown emphasis), a boxed letter (\texttt{\textbackslash boxed\{\}} or a model's own box tokens), or, in a long response, only a committed conclusion (the last explicit answer statement, or a final line that is an option), so that options mentioned along a chain of reasoning are not read as the answer. On G1, whose options are the fields themselves, a response that names a field (``Image 4'') is mapped to the option that is this field, since the options rotate while the fields keep their numbers; in the other tasks a mention of an image is not an answer.
Responses it cannot map go to the judge LLM (Qwen2.5-32B-Instruct, AWQ), which sees the options and the end of the final answer and returns the option the response commits to, or none; answers mapped to no option count as wrong.
Reasoning models sometimes reach their budget of new tokens inside their reasoning, before they answer.
As in the thinking budget of Qwen3~\cite{yang2025qwen3}, the unfinished response is then closed with the stop-thinking instruction (``Considering the limited time by the user, I have to give the solution based on the thinking directly now.'' and the end-of-reasoning tag) and the model answers from there, with the same prompt, images and greedy decoding and at most 512 further tokens; the forced answer is parsed like any other.
This concerns \forcedN\ responses of the default runs of \forcedModels\ models on both sources, most of them from \forcedFirstName\ (\forcedFirst\% of its responses) and \forcedSecondName\ (\forcedSecond\%), at most \forcedRestMax\% for each of the others; \forcedLeft\ still unfinished after forcing and scored as giving no answer.
The field-set ablations and the text-only R1 runs are treated in the same way.

\section{Additional results}
\paragraph{Contamination search.}
We searched for near-duplicates of every field of both sources---the 2,590 images of the teaching library (its 2,585 pictures and the five fields cut from composites) and the 2,664 atlas fields---among 228,899 images of public pathology image--text data: PathCap (223,169 PubMed figure panels)~\cite{sun2024pathasst}, PathMMU (3,887)~\cite{sun2024pathmmu}, and the PathVQA (858)~\cite{he2020pathvqa} and Quilt-VQA (985)~\cite{seyfioglu2024quiltllava} test sets.
For every benchmark image we retrieve the five nearest corpus images in QuiltNet space (cosine $\geq$0.93) and confirm candidates with perceptual hashes and pixel correlation, using the rule of our de-duplication without its aHash branch (across corpora a 64-bit aHash only encodes the coarse stain and layout of an H\&E field).
No near-duplicate was found in either source; the 49 teaching and 71 atlas fields flagged by looser rules were inspected visually against their candidates and all show different tissue.
The search compares whole images and would miss a benchmark field that appears only as a crop of a larger published figure.

\begin{figure}[t]
\centering
\includegraphics[width=\linewidth]{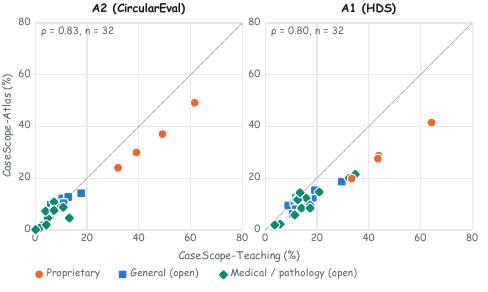}
\caption{\textbf{Teaching vs.\ Atlas.} Each point is a model: A2 circular accuracy (left) and A1 hierarchical score (right) on \bench-Teaching and on \bench-Atlas; dashed: equal scores; $\rho$: Spearman correlation over the models shown.}
\label{fig:replication}
\end{figure}

\begin{figure}[t]
\centering
\includegraphics[width=\linewidth]{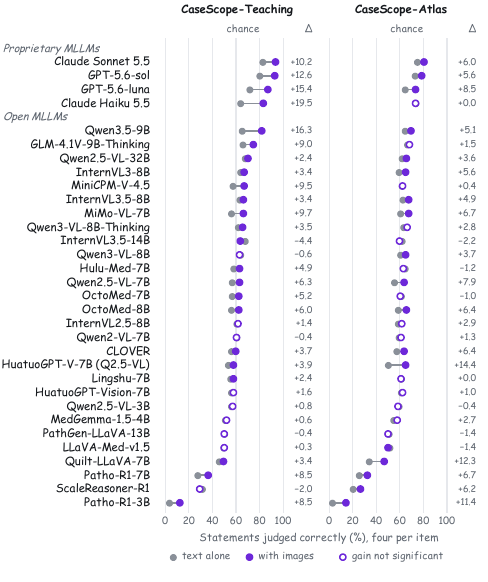}
\caption{\textbf{R1 with and without the images on \bench-Teaching (left) and \bench-Atlas (right)} (the proprietary, then the open models, each block in the order of Teaching accuracy with images). Share of finding statements judged correctly from the text alone (grey) and with the fields (violet; open: gain not significant, paired case bootstrap); $\Delta$: with minus without the images, in points.}
\label{fig:r1}
\end{figure}

\label{supp:results}
\Cref{tab:main-detail,tab:atlas-detail} give the complete results on \bench-Teaching and \bench-Atlas, respectively, including all evaluated models and per-model A1 and A2 confidence intervals. Table~3\ofmain shows a selection from both sources.
\Cref{fig:replication} shows each model's scores on the two sources, \cref{fig:r1} the R1 statement accuracy with and without the images on both sources, \cref{fig:diagnostic-reliability}, per model and for both sources, how often A2 returns the same diagnosis under all ten option rotations and how often a wrong answer is a classic look-alike, and \cref{fig:per_system} A2 accuracy per organ system on both sources.
\Cref{tab:multiview,tab:bias,tab:subsets,tab:a1} give, for both sources, the field-set ablations, the consistency under option rotation, the accuracy by case subset and the types of A1 errors.
\begin{table*}[t]
\centering
\footnotesize
\setlength{\tabcolsep}{4pt}
\caption{\textbf{Complete CaseScope-Teaching results with intervals} (\%). Columns and marks as in Table~3\ofmain; $\pm$: half-width of the 95\% case-bootstrap interval.}
\label{tab:main-detail}
\begin{tabular}{lccccccccc}
\toprule
 & A1 & A2 & A2\textsubscript{v} & B & Miss$\downarrow$ & C & C\textsubscript{$\kappa$} & R1 & $\Delta_{\text{img}}$ \\
\midrule
Random choice & -- & 0.0 & 10.1 & 0.5 & 49.7 & 11.4 & 0.08 & -- & -- \\
\midrule
\multicolumn{10}{l}{\textit{Proprietary MLLMs}} \\
Claude Sonnet 5.5 & \textbf{64.2}\,{\scriptsize$\pm$4.5} & \textbf{61.7}\,{\scriptsize$\pm$4.4} & \textbf{78.3} & \textbf{75.5} & \underline{11.4} & \textbf{79.5} & \textbf{0.81} & \textbf{93.2} & +10.2$^\ast$ \\
Claude Haiku 5.5 & \underline{43.9}\,{\scriptsize$\pm$4.6} & 39.0\,{\scriptsize$\pm$4.6} & 61.0 & \underline{57.4} & \textbf{10.3} & 54.5 & 0.55 & 83.2 & +19.5$^\ast$ \\
GPT-5.6-sol & 43.5\,{\scriptsize$\pm$4.5} & \underline{49.3}\,{\scriptsize$\pm$4.8} & \underline{67.4} & 55.5 & 19.4 & \underline{61.4} & \underline{0.64} & \underline{92.9} & +12.6$^\ast$ \\
GPT-5.6-luna & 33.5\,{\scriptsize$\pm$4.1} & 31.9\,{\scriptsize$\pm$4.2} & 57.3 & 51.1 & 33.6 & 56.8 & 0.61 & 87.1 & +15.4$^\ast$ \\
\midrule
\multicolumn{10}{l}{\textit{General MLLMs (open)}} \\
Qwen2.5-VL-3B~\cite{bai2025qwen25vl} & 8.9\,{\scriptsize$\pm$2.4} & 7.3\,{\scriptsize$\pm$2.4} & 25.4 & 9.7 & 26.7 & 22.7 & 0.10 & 57.1 & +0.8 \\
Qwen2.5-VL-7B~\cite{bai2025qwen25vl} & 11.7\,{\scriptsize$\pm$2.6} & 6.0\,{\scriptsize$\pm$2.2} & 28.1 & 18.7 & 15.8 & 9.1 & 0.06 & 63.0 & +6.3$^\ast$ \\
Qwen2.5-VL-32B~\cite{bai2025qwen25vl} & 16.9\,{\scriptsize$\pm$3.2} & 7.6\,{\scriptsize$\pm$2.5} & 30.6 & 21.8 & 23.1 & 25.0 & 0.02 & 70.1 & +2.4$^\ast$ \\
Qwen2-VL-7B~\cite{bai2025qwen25vl} & 11.2\,{\scriptsize$\pm$2.6} & 7.8\,{\scriptsize$\pm$2.4} & 25.3 & 25.5 & \textbf{3.3} & 13.6 & \textminus{}0.02 & 60.5 & \textminus{}0.4 \\
InternVL2.5-8B~\cite{chen2024internvl25} & 11.8\,{\scriptsize$\pm$2.6} & 8.5\,{\scriptsize$\pm$2.6} & 27.0 & 10.5 & 41.0 & 22.7 & 0.09 & 61.8 & +1.4 \\
InternVL3-8B~\cite{zhu2025internvl3} & 14.8\,{\scriptsize$\pm$3.0} & 10.8\,{\scriptsize$\pm$2.9} & 29.7 & 29.7 & 22.7 & 25.0 & 0.13 & 67.1 & +3.4$^\ast$ \\
InternVL3.5-8B~\cite{wang2025internvl35} & 18.3\,{\scriptsize$\pm$3.3} & 11.0\,{\scriptsize$\pm$2.9} & 34.7 & 30.5 & 15.0 & 27.3 & 0.16 & 66.5 & +3.4$^\ast$ \\
Qwen3-VL-8B~\cite{bai2025qwen3vl} & 11.9\,{\scriptsize$\pm$2.8} & 10.6\,{\scriptsize$\pm$2.9} & 30.2 & 30.5 & 6.7 & 27.3 & 0.02 & 63.3 & \textminus{}0.6 \\
Qwen3.5-9B~\cite{qwen2026qwen35} & 29.7\,{\scriptsize$\pm$4.0} & \textbf{17.9}\,{\scriptsize$\pm$3.7} & \textbf{50.6} & 31.8 & 45.1 & \textbf{36.4} & \textbf{0.42} & \textbf{81.5} & +16.3$^\ast$ \\
MiMo-VL-7B~\cite{xiaomi2025mimovl} & 15.9\,{\scriptsize$\pm$3.0} & 7.8\,{\scriptsize$\pm$2.5} & 29.9 & 26.8 & 22.5 & 29.5 & 0.13 & 66.1 & +9.7$^\ast$ \\
Qwen3-VL-8B-Thinking~\cite{bai2025qwen3vl} & 13.9\,{\scriptsize$\pm$3.0} & 10.3\,{\scriptsize$\pm$2.9} & 29.1 & 25.5 & 13.0 & \textbf{36.4} & 0.06 & 65.6 & +3.5$^\ast$ \\
GLM-4.1V-9B-Thinking~\cite{glmv2025glm41v} & 19.4\,{\scriptsize$\pm$3.4} & 12.6\,{\scriptsize$\pm$3.1} & 36.3 & 31.1 & 24.7 & 29.5 & 0.15 & \underline{74.7} & +9.0$^\ast$ \\
MiniCPM-V-4.5~\cite{yu2026minicpmv45} & 14.2\,{\scriptsize$\pm$3.0} & 11.0\,{\scriptsize$\pm$2.9} & 34.6 & 30.3 & 20.6 & 27.3 & 0.21 & 66.8 & +9.5$^\ast$ \\
InternVL3.5-14B~\cite{wang2025internvl35} & 10.7\,{\scriptsize$\pm$2.5} & 5.5\,{\scriptsize$\pm$2.1} & 25.7 & 30.0 & 11.9 & 13.6 & \textminus{}0.05 & 63.7 & \textminus{}4.4$^\ast$ \\
\midrule
\multicolumn{10}{l}{\textit{Medical MLLMs}} \\
HuatuoGPT-Vision-7B~\cite{chen2024huatuogptvision} & 16.9\,{\scriptsize$\pm$2.9} & 3.0\,{\scriptsize$\pm$1.6} & 27.1 & 25.3 & 15.6 & 18.2 & 0.13 & 57.7 & +1.6 \\
HuatuoGPT-Vision-7B (Q2.5-VL)~\cite{chen2024huatuogptvision} & 12.0\,{\scriptsize$\pm$2.6} & 8.0\,{\scriptsize$\pm$2.5} & 27.3 & 30.8 & 6.1 & 13.6 & 0.05 & 57.8 & +3.9$^\ast$ \\
Lingshu-7B~\cite{lasa2025lingshu} & 21.0\,{\scriptsize$\pm$3.4} & 5.7\,{\scriptsize$\pm$2.2} & 33.6 & \underline{34.2} & 35.4 & \textbf{36.4} & 0.32 & 57.8 & +2.4$^\ast$ \\
OctoMed-7B~\cite{ossowski2026octomed} & 14.1\,{\scriptsize$\pm$2.7} & 4.8\,{\scriptsize$\pm$2.1} & 32.1 & 12.9 & 21.8 & 20.5 & 0.02 & 62.2 & +5.2$^\ast$ \\
OctoMed-8B~\cite{ossowski2026octomed} & 15.9\,{\scriptsize$\pm$2.9} & 7.3\,{\scriptsize$\pm$2.4} & 27.0 & 30.0 & \underline{5.7} & \underline{31.8} & 0.15 & 62.2 & +6.0$^\ast$ \\
MedGemma-1.5-4B~\cite{sellergren2025medgemma} & 11.5\,{\scriptsize$\pm$2.5} & 2.5\,{\scriptsize$\pm$1.5} & 21.5 & 18.9 & 15.2 & 20.5 & 0.15 & 51.7 & +0.6 \\
Hulu-Med-7B~\cite{jiang2025hulumed} & 17.3\,{\scriptsize$\pm$3.2} & \underline{13.3}\,{\scriptsize$\pm$3.2} & 41.2 & 33.4 & 19.7 & 13.6 & 0.28 & 63.3 & +4.9$^\ast$ \\
LLaVA-Med-v1.5$^\dagger$~\cite{li2023llavamed} & 5.7\,{\scriptsize$\pm$1.7} & 0.0\,{\scriptsize$\pm$0.0} & 13.1 & 0.8 & 13.5 & 2.3 & 0.01 & 49.7 & +0.3 \\
\midrule
\multicolumn{10}{l}{\textit{Pathology MLLMs}} \\
Quilt-LLaVA-7B$^\dagger$~\cite{seyfioglu2024quiltllava} & 3.8\,{\scriptsize$\pm$1.6} & 1.6\,{\scriptsize$\pm$1.3} & 17.7 & 1.1 & 40.0 & 2.3 & \textminus{}0.08 & 49.4 & +3.4$^\ast$ \\
PathGen-LLaVA-13B$^\dagger$~\cite{sun2025pathgen} & 17.4\,{\scriptsize$\pm$3.1} & 4.4\,{\scriptsize$\pm$1.9} & 20.7 & 13.9 & 52.4 & 4.5 & 0.06 & 50.1 & \textminus{}0.4 \\
CLOVER~\cite{chen2024clover} & 12.5\,{\scriptsize$\pm$2.9} & 4.1\,{\scriptsize$\pm$1.9} & 25.1 & 20.5 & 22.9 & \textbf{36.4} & 0.18 & 59.9 & +3.7$^\ast$ \\
Patho-R1-3B~\cite{zhang2025pathor1} & 13.4\,{\scriptsize$\pm$3.0} & 0.2\,{\scriptsize$\pm$0.3} & 9.8 & 2.4 & 36.6 & 6.8 & \textbf{0.42} & 12.1 & +8.5$^\ast$ \\
Patho-R1-7B~\cite{zhang2025pathor1} & \textbf{34.7}\,{\scriptsize$\pm$4.4} & 7.1\,{\scriptsize$\pm$2.4} & 36.2 & 33.2 & 11.1 & 29.5 & 0.28 & 36.1 & +8.5$^\ast$ \\
ScaleReasoner-R1~\cite{phan2026scalevqa} & \underline{32.5}\,{\scriptsize$\pm$4.2} & 10.8\,{\scriptsize$\pm$2.8} & \underline{41.8} & \textbf{34.5} & 6.1 & 25.0 & \underline{0.33} & 29.2 & \textminus{}2.0 \\
\bottomrule
\end{tabular}
\end{table*}

\begin{table*}[t]
\centering
\footnotesize
\setlength{\tabcolsep}{4pt}
\caption{\textbf{Complete CaseScope-Atlas results with intervals} (\%). Columns and marks as in Table~3\ofmain; $\pm$ as in Table~\ref{tab:main-detail}, also for G1. G1 rotates the options while the images keep their order: naming one image, whichever it is, in every rotation is right on \atlGoneChanceVan\% of the items on average (always image \atlGoneFixedBestImage: \atlGoneFixedBest\%), independent guesses on \atlGoneChance\%.}
\label{tab:atlas-detail}
\begin{tabular}{lcccccccc}
\toprule
 & A1 & A2 & A2\textsubscript{v} & B & Miss$\downarrow$ & R1 & $\Delta_{\text{img}}$ & G1 \\
\midrule
Random choice & -- & 0.0 & 9.9 & 0.9 & 50.9 & -- & -- & 9.2 \\
\midrule
\multicolumn{9}{l}{\textit{Proprietary MLLMs}} \\
Claude Sonnet 5.5 & \textbf{41.6}\,{\scriptsize$\pm$3.3} & \textbf{49.1}\,{\scriptsize$\pm$3.4} & \textbf{68.5} & \textbf{61.5} & \textbf{21.6} & \textbf{80.8} & +6.0$^\ast$ & \textbf{61.5}\,{\scriptsize$\pm$5.0} \\
Claude Haiku 5.5 & \underline{28.8}\,{\scriptsize$\pm$3.1} & 29.9\,{\scriptsize$\pm$3.2} & 52.9 & \underline{47.0} & \underline{25.4} & 73.6 & +0.0 & \underline{48.2}\,{\scriptsize$\pm$4.7} \\
GPT-5.6-sol & 27.6\,{\scriptsize$\pm$3.1} & \underline{37.1}\,{\scriptsize$\pm$3.4} & \underline{57.1} & 41.9 & 35.4 & \underline{78.6} & +5.6$^\ast$ & 45.6\,{\scriptsize$\pm$4.9} \\
GPT-5.6-luna & 19.8\,{\scriptsize$\pm$2.8} & 24.1\,{\scriptsize$\pm$3.0} & 46.6 & 32.7 & 46.4 & 73.1 & +8.5$^\ast$ & 39.9\,{\scriptsize$\pm$4.6} \\
\midrule
\multicolumn{9}{l}{\textit{General MLLMs (open)}} \\
Qwen2.5-VL-3B~\cite{bai2025qwen25vl} & 9.6\,{\scriptsize$\pm$1.9} & 10.0\,{\scriptsize$\pm$2.1} & 25.8 & 10.9 & 34.8 & 58.7 & \textminus{}0.4 & 31.4\,{\scriptsize$\pm$4.1} \\
Qwen2.5-VL-7B~\cite{bai2025qwen25vl} & 10.9\,{\scriptsize$\pm$2.0} & 9.8\,{\scriptsize$\pm$2.1} & 28.7 & 17.2 & 19.8 & 63.6 & +7.9$^\ast$ & \underline{39.4}\,{\scriptsize$\pm$4.6} \\
Qwen2.5-VL-32B~\cite{bai2025qwen25vl} & 12.8\,{\scriptsize$\pm$2.2} & 11.0\,{\scriptsize$\pm$2.1} & 29.0 & 17.5 & 29.3 & 65.4 & +3.6$^\ast$ & 36.7\,{\scriptsize$\pm$4.6} \\
Qwen2-VL-7B~\cite{bai2025qwen25vl} & 10.7\,{\scriptsize$\pm$1.8} & 8.2\,{\scriptsize$\pm$1.9} & 25.7 & 20.1 & 13.1 & 60.9 & +1.3 & 31.4\,{\scriptsize$\pm$4.3} \\
InternVL2.5-8B~\cite{chen2024internvl25} & 8.1\,{\scriptsize$\pm$1.6} & 8.8\,{\scriptsize$\pm$1.9} & 26.4 & 13.1 & 39.1 & 61.4 & +2.9 & 20.6\,{\scriptsize$\pm$3.9} \\
InternVL3-8B~\cite{zhu2025internvl3} & 10.6\,{\scriptsize$\pm$1.9} & 9.6\,{\scriptsize$\pm$2.1} & 28.7 & 23.9 & 27.9 & 65.1 & +5.6$^\ast$ & 26.4\,{\scriptsize$\pm$3.9} \\
InternVL3.5-8B~\cite{wang2025internvl35} & 12.3\,{\scriptsize$\pm$2.1} & 9.4\,{\scriptsize$\pm$2.0} & 31.4 & 19.9 & 22.9 & 67.6 & +4.9$^\ast$ & 36.7\,{\scriptsize$\pm$4.3} \\
Qwen3-VL-8B~\cite{bai2025qwen3vl} & 12.3\,{\scriptsize$\pm$2.1} & 12.1\,{\scriptsize$\pm$2.2} & 30.3 & 17.1 & 21.7 & 64.6 & +3.7$^\ast$ & 32.8\,{\scriptsize$\pm$4.3} \\
Qwen3.5-9B~\cite{qwen2026qwen35} & 18.6\,{\scriptsize$\pm$2.6} & \textbf{14.2}\,{\scriptsize$\pm$2.5} & \textbf{41.7} & 28.1 & 44.0 & \textbf{69.5} & +5.1$^\ast$ & \textbf{42.7}\,{\scriptsize$\pm$4.8} \\
MiMo-VL-7B~\cite{xiaomi2025mimovl} & 12.2\,{\scriptsize$\pm$2.2} & 8.9\,{\scriptsize$\pm$1.9} & 28.3 & 23.6 & 47.9 & 67.3 & +6.7$^\ast$ & 33.5\,{\scriptsize$\pm$4.6} \\
Qwen3-VL-8B-Thinking~\cite{bai2025qwen3vl} & 12.8\,{\scriptsize$\pm$2.2} & 12.1\,{\scriptsize$\pm$2.2} & 29.4 & 17.1 & 35.5 & 66.2 & +2.8 & 26.1\,{\scriptsize$\pm$4.2} \\
GLM-4.1V-9B-Thinking~\cite{glmv2025glm41v} & 15.4\,{\scriptsize$\pm$2.3} & \underline{12.6}\,{\scriptsize$\pm$2.2} & \underline{33.6} & \underline{31.6} & 42.6 & \underline{68.3} & +1.5 & 36.2\,{\scriptsize$\pm$4.3} \\
MiniCPM-V-4.5~\cite{yu2026minicpmv45} & 12.6\,{\scriptsize$\pm$2.1} & 10.4\,{\scriptsize$\pm$2.0} & 31.3 & 22.7 & 29.9 & 62.4 & +0.4 & 32.3\,{\scriptsize$\pm$4.4} \\
InternVL3.5-14B~\cite{wang2025internvl35} & 6.3\,{\scriptsize$\pm$1.4} & 5.5\,{\scriptsize$\pm$1.6} & 22.1 & 28.3 & 12.2 & 59.9 & \textminus{}2.2 & 11.5\,{\scriptsize$\pm$2.9} \\
\midrule
\multicolumn{9}{l}{\textit{Medical MLLMs}} \\
HuatuoGPT-Vision-7B~\cite{chen2024huatuogptvision} & 9.2\,{\scriptsize$\pm$1.7} & 1.5\,{\scriptsize$\pm$0.9} & 22.2 & 21.5 & 15.1 & 62.1 & +1.0 & 23.9\,{\scriptsize$\pm$3.8} \\
HuatuoGPT-Vision-7B (Q2.5-VL)~\cite{chen2024huatuogptvision} & 11.8\,{\scriptsize$\pm$2.0} & 9.4\,{\scriptsize$\pm$2.1} & 29.2 & 24.7 & 13.9 & 64.9 & +14.4$^\ast$ & 31.7\,{\scriptsize$\pm$4.2} \\
Lingshu-7B~\cite{lasa2025lingshu} & 14.8\,{\scriptsize$\pm$2.2} & 7.0\,{\scriptsize$\pm$1.9} & 29.1 & \textbf{32.3} & 30.8 & 61.3 & +0.0 & \underline{39.4}\,{\scriptsize$\pm$4.5} \\
OctoMed-7B~\cite{ossowski2026octomed} & 8.4\,{\scriptsize$\pm$1.7} & 4.8\,{\scriptsize$\pm$1.4} & 26.4 & 15.1 & 40.6 & 60.4 & \textminus{}1.0 & 22.5\,{\scriptsize$\pm$4.0} \\
OctoMed-8B~\cite{ossowski2026octomed} & 12.2\,{\scriptsize$\pm$2.0} & 10.8\,{\scriptsize$\pm$2.1} & 29.3 & 18.2 & 22.7 & 65.3 & +6.4$^\ast$ & 22.2\,{\scriptsize$\pm$3.9} \\
MedGemma-1.5-4B~\cite{sellergren2025medgemma} & 5.9\,{\scriptsize$\pm$1.3} & 1.8\,{\scriptsize$\pm$0.9} & 16.4 & 17.7 & 19.6 & 57.8 & +2.7 & 11.0\,{\scriptsize$\pm$2.8} \\
Hulu-Med-7B~\cite{jiang2025hulumed} & 9.6\,{\scriptsize$\pm$1.7} & 4.8\,{\scriptsize$\pm$1.4} & 29.5 & 20.7 & 29.3 & 63.2 & \textminus{}1.2 & 21.6\,{\scriptsize$\pm$3.6} \\
LLaVA-Med-v1.5$^\dagger$~\cite{li2023llavamed} & 2.4\,{\scriptsize$\pm$0.8} & 0.0\,{\scriptsize$\pm$0.0} & 10.2 & 0.0 & \textbf{7.4} & 50.0 & \textminus{}1.4$^\ast$ & 0.7\,{\scriptsize$\pm$0.8} \\
\midrule
\multicolumn{9}{l}{\textit{Pathology MLLMs}} \\
Quilt-LLaVA-7B$^\dagger$~\cite{seyfioglu2024quiltllava} & 2.0\,{\scriptsize$\pm$0.8} & 0.6\,{\scriptsize$\pm$0.6} & 14.5 & 10.6 & 62.9 & 46.4 & +12.3$^\ast$ & 15.1\,{\scriptsize$\pm$3.0} \\
PathGen-LLaVA-13B$^\dagger$~\cite{sun2025pathgen} & 8.5\,{\scriptsize$\pm$1.7} & 1.9\,{\scriptsize$\pm$1.0} & 18.7 & 22.8 & 72.5 & 49.7 & \textminus{}1.4 & 16.1\,{\scriptsize$\pm$3.2} \\
CLOVER~\cite{chen2024clover} & 11.6\,{\scriptsize$\pm$2.0} & 7.2\,{\scriptsize$\pm$1.8} & 25.7 & 14.2 & 42.4 & 63.6 & +6.4$^\ast$ & 16.1\,{\scriptsize$\pm$3.6} \\
Patho-R1-3B~\cite{zhang2025pathor1} & 14.3\,{\scriptsize$\pm$2.2} & 0.1\,{\scriptsize$\pm$0.2} & 11.5 & 3.1 & 40.8 & 14.4 & +11.4$^\ast$ & 2.1\,{\scriptsize$\pm$1.3} \\
Patho-R1-7B~\cite{zhang2025pathor1} & \textbf{21.5}\,{\scriptsize$\pm$2.7} & 7.6\,{\scriptsize$\pm$1.8} & 28.9 & 19.4 & 18.7 & 32.2 & +6.7$^\ast$ & 20.6\,{\scriptsize$\pm$3.7} \\
ScaleReasoner-R1~\cite{phan2026scalevqa} & \underline{20.1}\,{\scriptsize$\pm$2.6} & 8.6\,{\scriptsize$\pm$1.9} & 32.5 & 28.7 & \underline{9.3} & 26.3 & +6.2$^\ast$ & 18.8\,{\scriptsize$\pm$3.4} \\
\bottomrule
\end{tabular}
\end{table*}

\begin{table*}[t]
\centering
\footnotesize
\setlength{\tabcolsep}{4pt}
\caption{\textbf{Field-set ablation on \bench-Teaching and \bench-Atlas} (A2, CircularEval, \%). Same cases and options, different evidence: all fields with the specimen site (All), only the highest-power field, all fields without the specimen site; in brackets the change from All in points.}
\label{tab:multiview}\label{tab:atlas-multiview}
\begin{tabular}{l|ccc|ccc}
\toprule
 & \multicolumn{3}{c|}{\bench-Teaching} & \multicolumn{3}{c}{\bench-Atlas} \\
Model & All & High-power only & No site & All & High-power only & No site \\
\midrule
Qwen2.5-VL-3B & 7.3 & 7.3 {\scriptsize(+0.0)} & 2.3 {\scriptsize(\textminus{}5.0)} & 10.0 & 6.4 {\scriptsize(\textminus{}3.6)} & 7.8 {\scriptsize(\textminus{}2.3)} \\
Qwen2.5-VL-7B & 6.0 & 8.0 {\scriptsize(+2.1)} & 1.4 {\scriptsize(\textminus{}4.6)} & 9.8 & 8.0 {\scriptsize(\textminus{}1.8)} & 6.1 {\scriptsize(\textminus{}3.6)} \\
Qwen2.5-VL-32B & 7.6 & 5.5 {\scriptsize(\textminus{}2.1)} & 1.8 {\scriptsize(\textminus{}5.7)} & 11.0 & 7.0 {\scriptsize(\textminus{}4.0)} & 6.4 {\scriptsize(\textminus{}4.6)} \\
Qwen2-VL-7B & 7.8 & 7.3 {\scriptsize(\textminus{}0.5)} & 2.3 {\scriptsize(\textminus{}5.5)} & 8.2 & 7.1 {\scriptsize(\textminus{}1.1)} & 6.6 {\scriptsize(\textminus{}1.6)} \\
InternVL2.5-8B & 8.5 & 8.7 {\scriptsize(+0.2)} & 2.1 {\scriptsize(\textminus{}6.4)} & 8.8 & 7.5 {\scriptsize(\textminus{}1.2)} & 5.4 {\scriptsize(\textminus{}3.4)} \\
InternVL3-8B & 10.8 & 11.5 {\scriptsize(+0.7)} & 3.4 {\scriptsize(\textminus{}7.3)} & 9.6 & 10.4 {\scriptsize(+0.7)} & 5.2 {\scriptsize(\textminus{}4.4)} \\
InternVL3.5-8B & 11.0 & 10.3 {\scriptsize(\textminus{}0.7)} & 3.9 {\scriptsize(\textminus{}7.1)} & 9.4 & 8.5 {\scriptsize(\textminus{}0.9)} & 4.4 {\scriptsize(\textminus{}5.0)} \\
Qwen3-VL-8B & 10.6 & 10.1 {\scriptsize(\textminus{}0.5)} & 3.9 {\scriptsize(\textminus{}6.7)} & 12.1 & 10.1 {\scriptsize(\textminus{}2.0)} & 8.9 {\scriptsize(\textminus{}3.2)} \\
Qwen3.5-9B & 17.9 & 10.8 {\scriptsize(\textminus{}7.1)} & 11.0 {\scriptsize(\textminus{}6.9)} & 14.2 & 11.8 {\scriptsize(\textminus{}2.5)} & 10.4 {\scriptsize(\textminus{}3.9)} \\
MiMo-VL-7B & 7.8 & 8.5 {\scriptsize(+0.7)} & 3.9 {\scriptsize(\textminus{}3.9)} & 8.9 & 7.0 {\scriptsize(\textminus{}1.9)} & 6.8 {\scriptsize(\textminus{}2.1)} \\
Qwen3-VL-8B-Thinking & 10.3 & 7.6 {\scriptsize(\textminus{}2.8)} & 0.9 {\scriptsize(\textminus{}9.4)} & 12.1 & 10.1 {\scriptsize(\textminus{}2.0)} & 6.6 {\scriptsize(\textminus{}5.5)} \\
GLM-4.1V-9B-Thinking & 12.6 & 9.6 {\scriptsize(\textminus{}3.0)} & 3.2 {\scriptsize(\textminus{}9.4)} & 12.6 & 10.2 {\scriptsize(\textminus{}2.4)} & 8.9 {\scriptsize(\textminus{}3.8)} \\
MiniCPM-V-4.5 & 11.0 & 8.7 {\scriptsize(\textminus{}2.3)} & 2.3 {\scriptsize(\textminus{}8.7)} & 10.4 & 8.8 {\scriptsize(\textminus{}1.6)} & 6.8 {\scriptsize(\textminus{}3.6)} \\
InternVL3.5-14B & 5.5 & 7.6 {\scriptsize(+2.1)} & 0.9 {\scriptsize(\textminus{}4.6)} & 5.5 & 7.1 {\scriptsize(+1.6)} & 0.9 {\scriptsize(\textminus{}4.6)} \\
HuatuoGPT-Vision-7B & 3.0 & 3.7 {\scriptsize(+0.7)} & 1.1 {\scriptsize(\textminus{}1.8)} & 1.5 & 1.8 {\scriptsize(+0.3)} & 0.5 {\scriptsize(\textminus{}1.0)} \\
HuatuoGPT-Vision-7B (Q2.5-VL) & 8.0 & 6.7 {\scriptsize(\textminus{}1.4)} & 2.1 {\scriptsize(\textminus{}6.0)} & 9.4 & 6.6 {\scriptsize(\textminus{}2.7)} & 6.5 {\scriptsize(\textminus{}2.9)} \\
Lingshu-7B & 5.7 & 2.3 {\scriptsize(\textminus{}3.4)} & 2.1 {\scriptsize(\textminus{}3.7)} & 7.0 & 5.4 {\scriptsize(\textminus{}1.6)} & 5.2 {\scriptsize(\textminus{}1.8)} \\
OctoMed-7B & 4.8 & 5.0 {\scriptsize(+0.2)} & 2.5 {\scriptsize(\textminus{}2.3)} & 4.8 & 3.8 {\scriptsize(\textminus{}1.0)} & 1.2 {\scriptsize(\textminus{}3.5)} \\
OctoMed-8B & 7.3 & 6.2 {\scriptsize(\textminus{}1.1)} & 1.4 {\scriptsize(\textminus{}6.0)} & 10.8 & 8.2 {\scriptsize(\textminus{}2.5)} & 6.5 {\scriptsize(\textminus{}4.2)} \\
MedGemma-1.5-4B & 2.5 & 3.4 {\scriptsize(+0.9)} & 0.7 {\scriptsize(\textminus{}1.8)} & 1.8 & 2.1 {\scriptsize(+0.4)} & 0.2 {\scriptsize(\textminus{}1.5)} \\
Hulu-Med-7B & 13.3 & 7.8 {\scriptsize(\textminus{}5.5)} & 6.7 {\scriptsize(\textminus{}6.7)} & 4.8 & 4.6 {\scriptsize(\textminus{}0.1)} & 2.1 {\scriptsize(\textminus{}2.6)} \\
LLaVA-Med-v1.5$^\dagger$ & 0.0 & 0.0 {\scriptsize(+0.0)} & 0.0 {\scriptsize(+0.0)} & 0.0 & 0.2 {\scriptsize(+0.2)} & 0.0 {\scriptsize(+0.0)} \\
Quilt-LLaVA-7B$^\dagger$ & 1.6 & 0.7 {\scriptsize(\textminus{}0.9)} & 0.2 {\scriptsize(\textminus{}1.4)} & 0.6 & 0.9 {\scriptsize(+0.3)} & 0.2 {\scriptsize(\textminus{}0.4)} \\
PathGen-LLaVA-13B$^\dagger$ & 4.4 & 5.7 {\scriptsize(+1.4)} & 1.6 {\scriptsize(\textminus{}2.8)} & 1.9 & 2.1 {\scriptsize(+0.3)} & 0.5 {\scriptsize(\textminus{}1.4)} \\
CLOVER & 4.1 & 2.3 {\scriptsize(\textminus{}1.8)} & 1.4 {\scriptsize(\textminus{}2.8)} & 7.2 & 5.2 {\scriptsize(\textminus{}2.0)} & 5.6 {\scriptsize(\textminus{}1.6)} \\
Patho-R1-3B & 0.2 & 0.0 {\scriptsize(\textminus{}0.2)} & 0.2 {\scriptsize(+0.0)} & 0.1 & 0.4 {\scriptsize(+0.2)} & 0.1 {\scriptsize(+0.0)} \\
Patho-R1-7B & 7.1 & 5.5 {\scriptsize(\textminus{}1.6)} & 4.8 {\scriptsize(\textminus{}2.3)} & 7.6 & 5.0 {\scriptsize(\textminus{}2.6)} & 6.9 {\scriptsize(\textminus{}0.7)} \\
ScaleReasoner-R1 & 10.8 & 8.0 {\scriptsize(\textminus{}2.8)} & 6.0 {\scriptsize(\textminus{}4.8)} & 8.6 & 5.4 {\scriptsize(\textminus{}3.2)} & 7.4 {\scriptsize(\textminus{}1.2)} \\
\bottomrule
\end{tabular}
\end{table*}

\begin{table*}[t]
\centering
\footnotesize
\setlength{\tabcolsep}{3.5pt}
\caption{\textbf{Consistency under option rotation} (A2, \%) on \bench-Teaching and \bench-Atlas. Vanilla: mean accuracy over the ten rotations; circular: correct in all ten. Content-consistent: the same option text is chosen in all ten rotations, right or wrong. Top letter: the most frequently chosen position and its share of answers (uniform = 10\%). $^\dagger$Single-image model given a numbered mosaic of the fields. \textbf{Bold}/\underline{underlined}: best/second-best model in the vanilla and circular columns of each source, ranked separately among the proprietary and among the open models.}
\label{tab:bias}
\begin{tabular}{lcccc@{\hspace{10pt}}cccc}
\toprule
 & \multicolumn{4}{c}{\bench-Teaching} & \multicolumn{4}{c}{\bench-Atlas} \\
\cmidrule(lr){2-5}\cmidrule(lr){6-9}
Model & Vanilla & Circular & \makecell{Content-\\consistent} & \makecell{Top letter\\(share)} & Vanilla & Circular & \makecell{Content-\\consistent} & \makecell{Top letter\\(share)} \\
\midrule
\multicolumn{9}{l}{\textit{Proprietary MLLMs}} \\
Claude Sonnet 5.5 & \textbf{78.3} & \textbf{61.7} & 67.2 & D (10) & \textbf{68.5} & \textbf{49.1} & 57.1 & C (11) \\
Claude Haiku 5.5 & 61.0 & 39.0 & 46.8 & B (12) & 52.9 & 29.9 & 36.9 & B (11) \\
GPT-5.6-sol & \underline{67.4} & \underline{49.3} & 58.7 & C (11) & \underline{57.1} & \underline{37.1} & 49.2 & D (11) \\
GPT-5.6-luna & 57.3 & 31.9 & 40.4 & D (11) & 46.6 & 24.1 & 32.4 & D (11) \\
\midrule
\multicolumn{9}{l}{\textit{Open MLLMs}} \\
Qwen2.5-VL-3B & 25.4 & 7.3 & 19.3 & E (15) & 25.8 & 10.0 & 18.6 & E (15) \\
Qwen2.5-VL-7B & 28.1 & 6.0 & 11.0 & E (16) & 28.7 & 9.8 & 13.5 & D (15) \\
Qwen2.5-VL-32B & 30.6 & 7.6 & 14.0 & C (15) & 29.0 & 11.0 & 17.2 & C (14) \\
Qwen2-VL-7B & 25.3 & 7.8 & 17.4 & C (19) & 25.7 & 8.2 & 18.0 & C (20) \\
InternVL2.5-8B & 27.0 & 8.5 & 13.5 & B (21) & 26.4 & 8.8 & 14.5 & B (22) \\
InternVL3-8B & 29.7 & 10.8 & 18.6 & B (16) & 28.7 & 9.6 & 14.4 & A (16) \\
InternVL3.5-8B & 34.7 & 11.0 & 15.8 & B (23) & 31.4 & 9.4 & 13.0 & B (26) \\
Qwen3-VL-8B & 30.2 & 10.6 & 21.8 & B (15) & 30.3 & 12.1 & 21.4 & B (16) \\
Qwen3.5-9B & \textbf{50.6} & \textbf{17.9} & 21.6 & B (16) & \textbf{41.7} & \textbf{14.2} & 16.0 & C (16) \\
MiMo-VL-7B & 29.9 & 7.8 & 14.7 & C (13) & 28.3 & 8.9 & 15.2 & C (15) \\
Qwen3-VL-8B-Thinking & 29.1 & 10.3 & 20.2 & C (15) & 29.4 & 12.1 & 19.4 & C (16) \\
GLM-4.1V-9B-Thinking & 36.3 & 12.6 & 20.0 & B (14) & \underline{33.6} & \underline{12.6} & 19.2 & B (16) \\
MiniCPM-V-4.5 & 34.6 & 11.0 & 14.0 & C (15) & 31.3 & 10.4 & 14.1 & C (16) \\
InternVL3.5-14B & 25.7 & 5.5 & 14.4 & B (18) & 22.1 & 5.5 & 12.2 & B (17) \\
HuatuoGPT-V-7B & 27.1 & 3.0 & 3.4 & B (19) & 22.2 & 1.5 & 1.5 & B (20) \\
HuatuoGPT-V-7B (Q2.5-VL) & 27.3 & 8.0 & 12.8 & C (15) & 29.2 & 9.4 & 13.5 & B (17) \\
Lingshu-7B & 33.6 & 5.7 & 6.4 & B (26) & 29.2 & 7.0 & 7.6 & B (29) \\
OctoMed-7B & 32.1 & 4.8 & 7.3 & C (14) & 26.4 & 4.8 & 6.2 & C (15) \\
OctoMed-8B & 27.0 & 7.3 & 12.2 & A (26) & 29.3 & 10.8 & 14.9 & A (27) \\
MedGemma-1.5-4B & 21.5 & 2.5 & 5.0 & A (33) & 16.4 & 1.8 & 2.8 & A (27) \\
Hulu-Med-7B & 41.2 & \underline{13.3} & 15.6 & B (22) & 29.5 & 4.8 & 7.0 & B (30) \\
LLaVA-Med-v1.5$^\dagger$ & 13.1 & 0.0 & 0.7 & A (54) & 10.2 & 0.0 & 0.1 & A (58) \\
Quilt-LLaVA-7B$^\dagger$ & 17.7 & 1.6 & 5.3 & D (28) & 14.5 & 0.6 & 1.8 & D (29) \\
PathGen-LLaVA-13B$^\dagger$ & 20.7 & 4.4 & 7.6 & C (30) & 18.7 & 1.9 & 4.4 & C (30) \\
CLOVER & 25.1 & 4.1 & 6.4 & D (23) & 25.7 & 7.2 & 10.6 & D (23) \\
Patho-R1-3B & 9.8 & 0.2 & 0.2 & J (27) & 11.5 & 0.1 & 0.1 & D (31) \\
Patho-R1-7B & 36.2 & 7.1 & 11.7 & G (14) & 28.9 & 7.6 & 14.8 & G (15) \\
ScaleReasoner-R1 & \underline{41.8} & 10.8 & 13.8 & G (14) & 32.5 & 8.6 & 13.5 & G (14) \\
\bottomrule
\end{tabular}
\end{table*}

\begin{table*}[t]
\centering
\scriptsize
\setlength{\tabcolsep}{3pt}
\renewcommand{\arraystretch}{0.88}
\caption{\textbf{A2 single-pass accuracy by case subset on \bench-Teaching (a) and \bench-Atlas (b)} (\%, mean over the ten rotations). H\&E sufficiency of the entity: Y diagnosable from H\&E alone, P suggestive, N ancillary tests needed. Behaviour: NN non-neoplastic, B benign, I intermediate, M malignant. $n$: cases. $^\dagger$Single-image model given a mosaic of the fields. \textbf{Bold}/\underline{underlined}: best/second-best model per column, ranked separately among the proprietary and among the open models. On \bench-Atlas the H\&E-sufficiency prior of \detailAtlHeFallback\ of the \detailAtlAdded\ entities added for it is a default P (\cref{fig:dataset-detail-atlas}).}
\label{tab:subsets}\label{tab:subsets-atlas}
{\footnotesize\textbf{(a) \bench-Teaching}}\\[2pt]
\begin{tabular}{l|ccc|ccc|cccc}
\toprule
 & \multicolumn{3}{c}{H\&E sufficiency} & \multicolumn{3}{c}{Fields} & \multicolumn{4}{c}{Behaviour} \\
Model & Y (194) & P (221) & N (21) & 2 (57) & 3-5 (184) & 6+ (195) & NN (114) & B (96) & I (61) & M (165) \\
\midrule
\multicolumn{11}{l}{\textit{Proprietary MLLMs}} \\
Claude Sonnet 5.5 & \textbf{83.9} & \textbf{74.6} & \textbf{66.2} & \textbf{75.6} & \textbf{79.6} & \textbf{77.9} & \textbf{83.9} & \textbf{76.1} & \textbf{73.4} & \textbf{77.5} \\
Claude Haiku 5.5 & 66.1 & 57.9 & 45.2 & \underline{62.6} & 61.4 & 60.1 & 64.6 & 54.6 & 57.7 & 63.3 \\
GPT-5.6-sol & \underline{73.9} & \underline{62.4} & \underline{60.5} & 62.1 & \underline{70.0} & \underline{66.5} & \underline{70.4} & \underline{73.8} & \underline{62.1} & \underline{63.6} \\
GPT-5.6-luna & 64.0 & 52.4 & 46.2 & 59.3 & 60.4 & 53.7 & 60.0 & 66.1 & 51.0 & 52.5 \\
\midrule
\multicolumn{11}{l}{\textit{Open MLLMs}} \\
Qwen2.5-VL-3B & 26.8 & 24.0 & 27.6 & 28.4 & 29.9 & 20.2 & 26.8 & 12.6 & 41.6 & 25.9 \\
Qwen2.5-VL-7B & 33.6 & 23.2 & 28.6 & 36.3 & 30.9 & 23.0 & 32.5 & 22.1 & 29.5 & 27.9 \\
Qwen2.5-VL-32B & 37.9 & 24.7 & 25.7 & 34.9 & 32.1 & 27.9 & 34.6 & 23.5 & 34.1 & 30.7 \\
Qwen2-VL-7B & 29.9 & 21.0 & 27.6 & 26.0 & 26.1 & 24.3 & 25.4 & 15.9 & 36.6 & 26.4 \\
InternVL2.5-8B & 32.5 & 22.2 & 26.2 & 32.6 & 31.2 & 21.3 & 29.6 & 20.1 & 31.3 & 27.6 \\
InternVL3-8B & 33.5 & 27.0 & 23.8 & 33.3 & 34.5 & 24.3 & 31.3 & 26.5 & 35.6 & 28.4 \\
InternVL3.5-8B & 37.7 & 33.3 & 21.0 & 41.9 & 36.3 & 31.0 & 33.6 & 31.8 & 37.2 & 36.1 \\
Qwen3-VL-8B & 32.3 & 28.5 & 28.1 & 34.9 & 32.2 & 26.8 & 32.2 & 24.5 & 36.1 & 29.9 \\
Qwen3.5-9B & \textbf{58.4} & \textbf{45.0} & \underline{37.1} & \textbf{54.7} & \textbf{48.9} & \textbf{51.0} & \textbf{54.8} & \textbf{62.2} & \underline{44.3} & \textbf{43.2} \\
MiMo-VL-7B & 38.2 & 24.2 & 12.4 & 32.1 & 30.6 & 28.6 & 35.1 & 28.5 & 35.2 & 25.1 \\
Qwen3-VL-8B-Thinking & 34.9 & 25.3 & 14.8 & 29.6 & 30.8 & 27.2 & 30.1 & 23.9 & 33.4 & 29.8 \\
GLM-4.1V-9B-Thinking & 42.7 & 30.6 & 36.2 & 33.0 & 37.7 & 35.8 & 39.0 & 25.1 & 40.3 & 39.3 \\
MiniCPM-V-4.5 & 40.6 & 30.4 & 23.8 & 35.4 & 36.0 & 33.1 & 31.0 & 40.4 & 41.3 & 31.3 \\
InternVL3.5-14B & 30.3 & 21.5 & 27.1 & 31.6 & 27.3 & 22.5 & 27.8 & 17.8 & 28.7 & 27.7 \\
HuatuoGPT-Vision-7B & 31.3 & 23.5 & 26.2 & 30.9 & 28.6 & 24.7 & 28.7 & 21.0 & 26.1 & 30.0 \\
HuatuoGPT-Vision-7B (Q2.5-VL) & 30.3 & 24.6 & 28.1 & 34.4 & 31.3 & 21.4 & 28.9 & 17.6 & 32.8 & 29.8 \\
Lingshu-7B & 36.9 & 30.7 & 33.3 & 38.8 & 38.2 & 27.7 & 39.0 & 31.8 & 36.9 & 29.6 \\
OctoMed-7B & 37.6 & 28.6 & 19.0 & 34.9 & 30.9 & 32.5 & 32.2 & 30.3 & 35.9 & 31.8 \\
OctoMed-8B & 29.1 & 25.0 & 28.1 & 27.7 & 29.8 & 24.2 & 25.7 & 22.4 & 29.8 & 29.5 \\
MedGemma-1.5-4B & 24.9 & 18.6 & 21.0 & 27.0 & 22.7 & 18.7 & 24.0 & 17.8 & 26.1 & 20.2 \\
Hulu-Med-7B & 46.5 & 36.8 & \textbf{39.0} & 46.5 & \underline{45.7} & 35.5 & \underline{47.1} & \underline{42.4} & 41.6 & 36.4 \\
LLaVA-Med-v1.5$^\dagger$ & 12.8 & 12.5 & 22.9 & 13.3 & 13.6 & 12.7 & 12.1 & 11.1 & 15.4 & 14.2 \\
Quilt-LLaVA-7B$^\dagger$ & 18.0 & 16.7 & 25.7 & 24.6 & 18.2 & 15.2 & 17.4 & 14.7 & 22.6 & 17.8 \\
PathGen-LLaVA-13B$^\dagger$ & 20.3 & 20.5 & 27.1 & 25.6 & 22.9 & 17.1 & 21.8 & 17.4 & 31.5 & 17.9 \\
CLOVER & 27.3 & 22.8 & 29.5 & 29.1 & 28.5 & 20.7 & 27.0 & 19.5 & 36.6 & 22.8 \\
Patho-R1-3B & 10.1 & 9.7 & 7.1 & 15.1 & 10.5 & 7.5 & 8.0 & 8.5 & 17.5 & 8.8 \\
Patho-R1-7B & 40.5 & 33.5 & 25.2 & 37.5 & 38.5 & 33.6 & 35.5 & 37.0 & \textbf{47.5} & 32.1 \\
ScaleReasoner-R1 & \underline{48.2} & \underline{37.2} & 31.9 & \underline{49.8} & 41.9 & \underline{39.4} & 45.2 & 37.9 & 43.8 & \underline{41.1} \\
\bottomrule
\end{tabular}
\\[6pt]
{\footnotesize\textbf{(b) \bench-Atlas}}\\[2pt]
\begin{tabular}{l|ccc|ccc|cccc}
\toprule
 & \multicolumn{3}{c}{H\&E sufficiency} & \multicolumn{3}{c}{Fields} & \multicolumn{4}{c}{Behaviour} \\
Model & Y (176) & P (601) & N (23) & 2 (375) & 3-5 (319) & 6+ (106) & NN (210) & B (190) & I (107) & M (293) \\
\midrule
\multicolumn{11}{l}{\textit{Proprietary MLLMs}} \\
Claude Sonnet 5.5 & \textbf{74.0} & \textbf{67.1} & \textbf{62.2} & \textbf{63.4} & \textbf{72.6} & \textbf{73.9} & \textbf{73.4} & \textbf{68.8} & \textbf{66.0} & \textbf{65.5} \\
Claude Haiku 5.5 & 56.3 & 52.2 & \underline{45.7} & 51.9 & 53.0 & 55.9 & 56.7 & 52.5 & 50.2 & 51.4 \\
GPT-5.6-sol & \underline{57.0} & \underline{57.7} & 43.5 & \underline{54.0} & \underline{59.3} & \underline{61.7} & \underline{62.5} & \underline{59.3} & \underline{50.5} & \underline{54.4} \\
GPT-5.6-luna & 48.4 & 46.3 & 42.2 & 46.5 & 45.9 & 49.2 & 52.5 & 48.4 & 41.1 & 43.2 \\
\midrule
\multicolumn{11}{l}{\textit{Open MLLMs}} \\
Qwen2.5-VL-3B & 31.2 & 24.3 & 23.0 & 25.4 & 23.9 & 32.8 & 28.9 & 20.4 & 28.0 & 26.2 \\
Qwen2.5-VL-7B & 38.1 & 26.1 & 26.5 & 27.6 & 27.7 & 35.7 & 34.0 & 21.8 & 25.2 & 30.7 \\
Qwen2.5-VL-32B & 41.9 & 25.5 & 22.6 & 25.3 & 29.5 & \underline{40.7} & 35.2 & 21.2 & 27.9 & 30.0 \\
Qwen2-VL-7B & 35.5 & 22.7 & 27.0 & 25.7 & 23.1 & 32.9 & 29.2 & 22.1 & 27.1 & 24.9 \\
InternVL2.5-8B & 33.2 & 24.4 & 28.3 & 23.8 & 26.4 & 35.7 & 30.0 & 20.2 & 22.3 & 29.4 \\
InternVL3-8B & 36.1 & 26.4 & 33.5 & 28.3 & 27.6 & 33.4 & 33.3 & 23.9 & 23.7 & 30.3 \\
InternVL3.5-8B & 39.0 & 29.4 & 24.3 & 30.0 & 31.4 & 36.1 & 33.9 & 26.8 & 28.3 & 33.7 \\
Qwen3-VL-8B & 40.7 & 27.3 & 30.0 & 28.2 & 30.9 & 36.0 & 35.6 & 21.9 & 26.5 & 33.4 \\
Qwen3.5-9B & \textbf{50.7} & \textbf{39.5} & 32.2 & \textbf{41.7} & \textbf{41.7} & \textbf{42.3} & \textbf{49.5} & \textbf{41.7} & \textbf{35.0} & \textbf{38.6} \\
MiMo-VL-7B & 41.3 & 24.6 & 25.7 & 26.0 & 28.6 & 35.8 & 36.8 & 21.9 & 25.6 & 27.4 \\
Qwen3-VL-8B-Thinking & 43.6 & 25.4 & 24.3 & 27.4 & 28.8 & 38.2 & 35.1 & 24.4 & 31.2 & 27.8 \\
GLM-4.1V-9B-Thinking & \underline{45.1} & 30.5 & 27.4 & 31.6 & \underline{34.0} & 39.3 & \underline{39.4} & 24.9 & 29.7 & \underline{36.5} \\
MiniCPM-V-4.5 & 38.4 & 29.5 & 24.8 & 31.0 & 30.1 & 36.3 & 35.2 & 25.6 & 28.9 & 33.2 \\
InternVL3.5-14B & 29.5 & 19.9 & 23.0 & 21.1 & 21.9 & 25.8 & 22.9 & 16.9 & 18.8 & 26.0 \\
HuatuoGPT-Vision-7B & 26.9 & 20.8 & 21.7 & 22.8 & 20.9 & 24.1 & 24.1 & 17.7 & 20.5 & 24.4 \\
HuatuoGPT-Vision-7B (Q2.5-VL) & 35.8 & 27.3 & 29.1 & 28.7 & 28.2 & 34.3 & 33.3 & 21.6 & 26.6 & 32.2 \\
Lingshu-7B & 31.5 & 28.2 & 36.5 & 28.7 & 28.3 & 33.4 & 33.8 & 25.1 & 23.7 & 30.4 \\
OctoMed-7B & 37.0 & 23.4 & 23.0 & 25.9 & 26.8 & 27.0 & 32.1 & 20.9 & 25.0 & 26.3 \\
OctoMed-8B & 35.8 & 27.6 & 26.5 & 29.3 & 27.6 & 34.5 & 34.7 & 23.0 & 26.4 & 30.7 \\
MedGemma-1.5-4B & 21.6 & 15.0 & 14.3 & 13.8 & 18.2 & 20.3 & 20.5 & 12.5 & 16.1 & 16.2 \\
Hulu-Med-7B & 35.3 & 27.7 & 31.3 & 29.5 & 29.7 & 28.6 & 35.0 & 26.3 & 26.7 & 28.6 \\
LLaVA-Med-v1.5$^\dagger$ & 11.7 & 9.8 & 7.8 & 8.7 & 11.1 & 12.4 & 9.0 & 7.8 & 12.9 & 11.6 \\
Quilt-LLaVA-7B$^\dagger$ & 17.0 & 13.9 & 11.7 & 12.9 & 15.4 & 17.5 & 15.9 & 13.6 & 13.2 & 14.5 \\
PathGen-LLaVA-13B$^\dagger$ & 18.9 & 18.7 & 14.8 & 18.4 & 18.7 & 19.7 & 20.7 & 15.6 & 21.4 & 18.2 \\
CLOVER & 28.2 & 24.9 & 26.5 & 25.8 & 23.5 & 31.8 & 28.8 & 20.7 & 23.6 & 27.4 \\
Patho-R1-3B & 10.7 & 11.6 & 14.8 & 13.0 & 10.8 & 8.1 & 10.9 & 11.0 & 12.4 & 11.9 \\
Patho-R1-7B & 29.0 & 28.5 & \textbf{38.7} & 29.3 & 28.0 & 30.3 & 32.8 & 26.4 & 31.9 & 26.6 \\
ScaleReasoner-R1 & 37.2 & \underline{31.0} & \underline{38.3} & \underline{32.6} & 32.3 & 33.4 & 36.4 & \underline{28.9} & \underline{32.3} & 32.3 \\
\bottomrule
\end{tabular}
\end{table*}

\begin{table*}[t]
\centering
\footnotesize
\setlength{\tabcolsep}{2.6pt}
\caption{\textbf{Open-ended diagnosis (A1) on \bench-Teaching and \bench-Atlas} (\%), graded by the higher of the LLM judge and the ontology match (Sec.~4\ofmain). HDS: hierarchical diagnostic score (exact entity 1, same family 0.5). Errors, as a share of all answers: F within the family, L another family of the same lineage, W wrong lineage, B wrong behaviour, N no answer. \textbf{Bold}/\underline{underlined}: best/second-best model in the HDS, exact and top-3 columns of each source, ranked separately among the proprietary and among the open models.}
\label{tab:a1}\label{tab:a1-atlas}
\begin{tabular}{lccc|ccccc||ccc|ccccc}
\toprule
 & \multicolumn{8}{c||}{\bench-Teaching} & \multicolumn{8}{c}{\bench-Atlas} \\
Model & HDS & Exact & Top-3 & F & L & W & B & N & HDS & Exact & Top-3 & F & L & W & B & N \\
\midrule
\multicolumn{17}{l}{\textit{Proprietary MLLMs}} \\
Claude Sonnet 5.5 & \textbf{64.2} & \textbf{60.6} & \textbf{80.7} & 7.3 & 15.8 & 2.1 & 14.2 & 0.0 & \textbf{41.6} & \textbf{37.8} & \textbf{55.1} & 7.8 & 35.2 & 4.6 & 14.6 & 0.0 \\
Claude Haiku 5.5 & \underline{43.9} & 39.4 & \underline{65.1} & 8.9 & 22.9 & 2.5 & 26.1 & 0.0 & \underline{28.8} & \underline{25.4} & \underline{38.9} & 6.9 & 38.0 & 7.9 & 21.9 & 0.0 \\
GPT-5.6-sol & 43.5 & \underline{39.9} & 57.1 & 7.1 & 31.7 & 6.2 & 15.1 & 0.0 & 27.6 & 25.2 & 33.9 & 4.8 & 38.4 & 15.6 & 16.0 & 0.0 \\
GPT-5.6-luna & 33.5 & 30.3 & 51.6 & 6.4 & 35.1 & 6.0 & 22.2 & 0.0 & 19.8 & 17.8 & 25.8 & 4.1 & 43.8 & 13.4 & 21.0 & 0.0 \\
\midrule
\multicolumn{17}{l}{\textit{Open MLLMs}} \\
Qwen2.5-VL-3B & 8.9 & 5.0 & 11.5 & 7.8 & 32.8 & 11.0 & 43.1 & 0.2 & 9.6 & 6.8 & 9.8 & 5.8 & 30.8 & 10.9 & 45.5 & 0.4 \\
Qwen2.5-VL-7B & 11.7 & 6.4 & 16.3 & 10.6 & 31.4 & 11.7 & 39.9 & 0.0 & 10.9 & 6.9 & 11.5 & 8.1 & 34.0 & 17.5 & 33.5 & 0.0 \\
Qwen2.5-VL-32B & 16.9 & 11.9 & 20.0 & 9.9 & 23.4 & 12.8 & 42.0 & 0.0 & 12.8 & 9.2 & 14.5 & 7.0 & 30.8 & 20.4 & 32.6 & 0.0 \\
Qwen2-VL-7B & 11.2 & 6.4 & 14.9 & 9.6 & 20.9 & 13.8 & 49.3 & 0.0 & 10.7 & 7.1 & 11.6 & 7.1 & 26.9 & 16.5 & 42.4 & 0.0 \\
InternVL2.5-8B & 11.8 & 6.7 & 17.0 & 10.3 & 23.6 & 14.9 & 44.5 & 0.0 & 8.1 & 4.8 & 9.9 & 6.6 & 26.5 & 21.5 & 40.6 & 0.0 \\
InternVL3-8B & 14.8 & 9.4 & 19.0 & 10.8 & 22.9 & 14.2 & 42.7 & 0.0 & 10.6 & 7.0 & 11.2 & 7.2 & 28.6 & 20.5 & 36.6 & 0.0 \\
InternVL3.5-8B & 18.3 & 12.6 & 22.9 & 11.5 & 25.7 & 9.6 & 40.6 & 0.0 & 12.3 & 8.1 & 13.8 & 8.4 & 31.1 & 18.9 & 33.5 & 0.0 \\
Qwen3-VL-8B & 11.9 & 6.9 & 17.7 & 10.1 & 26.8 & 10.8 & 45.4 & 0.0 & 12.3 & 9.5 & 14.6 & 5.6 & 30.6 & 15.8 & 38.5 & 0.0 \\
Qwen3.5-9B & 29.7 & 25.0 & \textbf{47.7} & 9.4 & 33.3 & 10.8 & 21.6 & 0.0 & 18.6 & 14.5 & \textbf{24.6} & 8.2 & 35.4 & 19.2 & 22.6 & 0.0 \\
MiMo-VL-7B & 15.9 & 10.6 & 20.0 & 10.8 & 33.3 & 11.9 & 33.5 & 0.0 & 12.2 & 9.5 & 14.6 & 5.5 & 36.2 & 19.5 & 29.2 & 0.0 \\
Qwen3-VL-8B-Thinking & 13.9 & 8.9 & 20.2 & 9.9 & 24.8 & 9.2 & 47.2 & 0.0 & 12.8 & 9.6 & 14.8 & 6.2 & 27.0 & 16.4 & 40.8 & 0.0 \\
GLM-4.1V-9B-Thinking & 19.4 & 14.4 & 26.8 & 9.9 & 23.6 & 5.7 & 46.3 & 0.0 & 15.4 & 12.0 & 19.0 & 6.9 & 33.6 & 13.6 & 33.9 & 0.0 \\
MiniCPM-V-4.5 & 14.2 & 8.5 & 18.1 & 11.5 & 28.0 & 12.4 & 39.7 & 0.0 & 12.6 & 8.5 & 13.5 & 8.2 & 34.4 & 17.6 & 31.2 & 0.0 \\
InternVL3.5-14B & 10.7 & 6.2 & 13.5 & 8.9 & 18.3 & 17.0 & 49.5 & 0.0 & 6.3 & 3.6 & 7.4 & 5.4 & 23.8 & 23.4 & 43.9 & 0.0 \\
HuatuoGPT-Vision-7B & 16.9 & 9.4 & 18.6 & 14.9 & 22.7 & 8.9 & 44.0 & 0.0 & 9.2 & 4.6 & 9.5 & 9.1 & 29.2 & 18.1 & 38.9 & 0.0 \\
HuatuoGPT-Vision-7B (Q2.5-VL) & 12.0 & 7.6 & 18.6 & 8.9 & 31.4 & 12.2 & 39.9 & 0.0 & 11.8 & 8.1 & 13.2 & 7.4 & 33.6 & 17.8 & 33.1 & 0.0 \\
Lingshu-7B & 21.0 & 13.8 & 25.5 & 14.4 & 32.8 & 11.5 & 27.5 & 0.0 & 14.8 & 9.8 & 16.0 & 10.1 & 35.5 & 15.4 & 29.2 & 0.0 \\
OctoMed-7B & 14.1 & 7.3 & 17.4 & 13.5 & 29.1 & 11.5 & 38.5 & 0.0 & 8.4 & 4.9 & 9.2 & 7.1 & 37.5 & 18.4 & 32.1 & 0.0 \\
OctoMed-8B & 15.9 & 8.3 & 18.3 & 15.4 & 26.6 & 10.1 & 39.7 & 0.0 & 12.2 & 7.5 & 14.2 & 9.4 & 31.5 & 16.8 & 34.9 & 0.0 \\
MedGemma-1.5-4B & 11.5 & 5.5 & 12.2 & 11.9 & 25.0 & 16.3 & 41.3 & 0.0 & 5.9 & 2.8 & 6.0 & 6.2 & 29.5 & 21.9 & 39.6 & 0.0 \\
Hulu-Med-7B & 17.3 & 12.4 & 22.5 & 9.9 & 31.0 & 8.5 & 38.3 & 0.0 & 9.6 & 4.9 & 10.4 & 9.5 & 37.5 & 13.0 & 35.1 & 0.0 \\
LLaVA-Med-v1.5$^\dagger$ & 5.7 & 2.1 & 5.0 & 7.3 & 28.0 & 8.3 & 54.1 & 0.2 & 2.4 & 0.6 & 2.5 & 3.5 & 38.9 & 10.8 & 46.1 & 0.1 \\
Quilt-LLaVA-7B$^\dagger$ & 3.8 & 2.1 & 5.3 & 3.4 & 33.3 & 22.7 & 35.1 & 3.4 & 2.0 & 0.4 & 2.9 & 3.2 & 36.1 & 25.0 & 26.8 & 8.5 \\
PathGen-LLaVA-13B$^\dagger$ & 17.4 & 11.7 & 15.8 & 11.5 & 33.5 & 6.0 & 37.4 & 0.0 & 8.5 & 4.2 & 9.4 & 8.5 & 40.9 & 13.5 & 32.9 & 0.0 \\
CLOVER & 12.5 & 7.6 & 16.1 & 9.9 & 32.8 & 11.9 & 37.8 & 0.0 & 11.6 & 7.5 & 12.4 & 8.2 & 33.5 & 17.6 & 33.1 & 0.0 \\
Patho-R1-3B & 13.4 & 10.3 & 13.1 & 6.2 & 14.0 & 1.6 & 26.4 & 41.5 & 14.3 & 10.2 & 13.8 & 8.1 & 20.6 & 0.2 & 48.0 & 12.8 \\
Patho-R1-7B & \textbf{34.7} & \textbf{30.0} & \underline{40.8} & 9.4 & 27.3 & 8.3 & 24.5 & 0.5 & \textbf{21.5} & \textbf{16.8} & \underline{23.9} & 9.5 & 40.8 & 7.2 & 25.2 & 0.5 \\
ScaleReasoner-R1 & \underline{32.5} & \underline{28.7} & 37.6 & 7.6 & 28.2 & 8.7 & 26.4 & 0.5 & \underline{20.1} & \underline{16.1} & 23.2 & 7.9 & 41.0 & 7.8 & 26.6 & 0.6 \\
\bottomrule
\end{tabular}
\end{table*}

\newsavebox{\diagnosticreliabilitybox}
\begin{figure*}[t]
\begin{lrbox}{\diagnosticreliabilitybox}
\begin{minipage}{\textwidth}
\centering
\includegraphics[width=\textwidth]{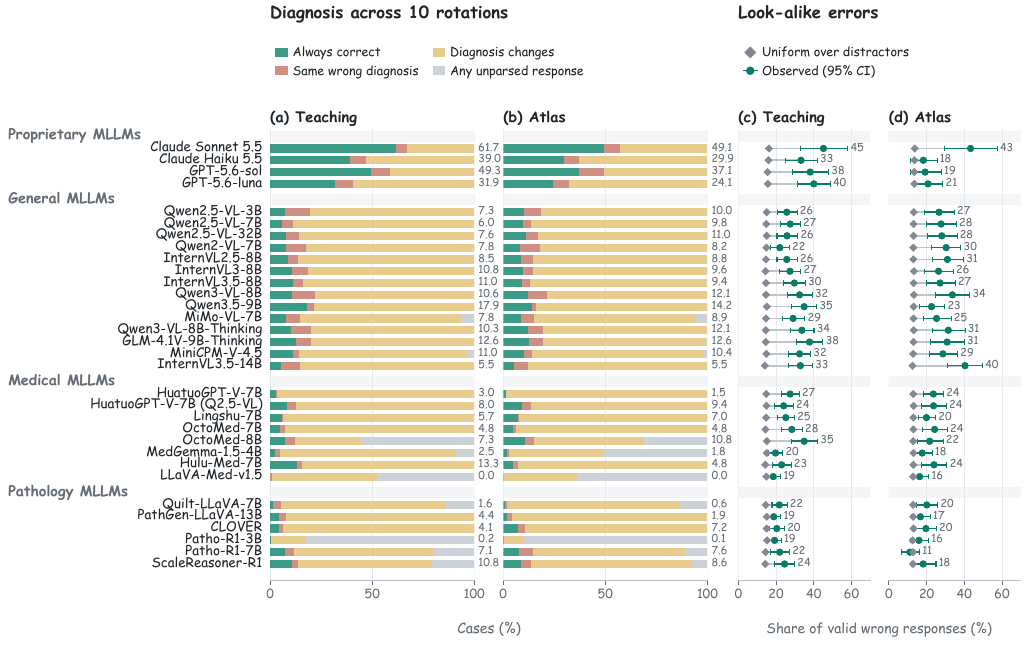}
\caption{\textbf{A2 diagnostic stability and look-alike errors on both sources} (four proprietary and \diagNModels\ open models).
(a, b) Cases across ten rotations; grey denotes any unparsed response.
(c, d) Valid wrong responses on the cases with a classic look-alike among the nine distractors (\diagNLookCases\ Teaching, \diagAtlNLookCases\ Atlas): observed shares versus uniform allocation over the distractors. Teaching look-alikes are the curated textbook differentials and the pairs the library sets side by side; Atlas look-alikes are the curated textbook differentials only.
Bars: 95\% CIs from 10,000 case-bootstrap resamples, preserving each case's rotations. Numbers: cases always correct (a, b) and observed look-alike share (c, d), \%.}
\label{fig:diagnostic-reliability}
\end{minipage}
\end{lrbox}
\typeout{RELIABILITY-HEIGHT: \the\dimexpr\ht\diagnosticreliabilitybox+\dp\diagnosticreliabilitybox\relax}
\typeout{RELIABILITY-WITH-GAP: \the\dimexpr\ht\diagnosticreliabilitybox+\dp\diagnosticreliabilitybox+\dbltextfloatsep\relax}
\ifdim\dimexpr\ht\diagnosticreliabilitybox+\dp\diagnosticreliabilitybox+\dbltextfloatsep\relax>\dimexpr\textheight*2/3\relax
  \PackageError{CaseScope figure}{Diagnostic reliability figure exceeds two thirds of a text page}{Reduce the figure height or caption.}
\fi
\centering\usebox{\diagnosticreliabilitybox}
\end{figure*}

\begin{figure*}[t]
\centering
\includegraphics[width=\linewidth]{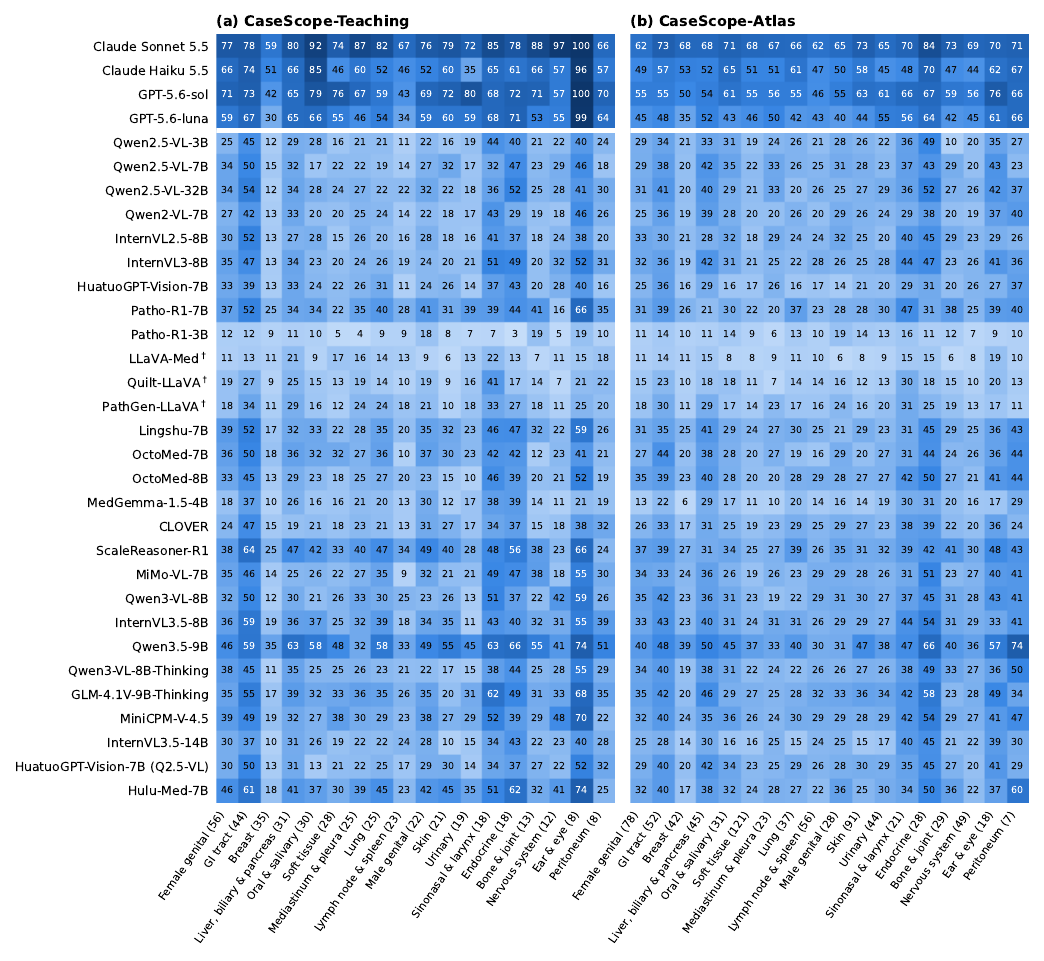}
\caption{\textbf{A2 single-pass accuracy (\%) per organ system} on (a) \bench-Teaching and (b) \bench-Atlas, for the same models (proprietary models in the top four rows) and the same organ systems (number of cases in brackets; chance: 10\%).}
\label{fig:per_system}
\end{figure*}

\end{document}